\pdfoutput=1
\PassOptionsToPackage{dvipsnames,table}{xcolor}
\documentclass[]{style}

\titlecardtrue

\title{\textit{Carve VFM}: Carving Specialists out of Generalists, Automatic Task-Specific Compression of Vision Foundation Models}

\toplogo[trim=12.23bp 12.38bp 18.98bp 12.23bp, clip]{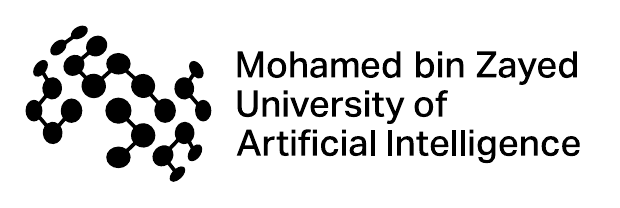}

\author[1]{Lotfi Abdelkrim Mecharbat}
\author[1]{Abdelrahman Elsayed}
\author[1]{Ibrahim Almakky}
\author[1]{Martin Tak\'a\v{c}}
\author[1]{Mohammad Yaqub}

\affiliation[1]{Mohamed bin Zayed University of Artificial Intelligence}

\date{Oct 12, 2026}

\usepackage{amsmath}
\usepackage{amssymb}
\usepackage{array}
\usepackage{colortbl}
\usepackage{makecell}
\usepackage{hhline}
\usepackage{adjustbox}
\usepackage{url}
\usepackage{pdflscape}
\usepackage{float}
\usepackage{xspace}
\usepackage{tikz}

\let\cite\citep
\providecommand{\std}[1]{\,{\scriptsize$\pm$#1}}


\abstract{
Vision foundation models (VFMs) provide transferable visual representations, but their inference cost limits deployment in many settings. Fine-tuning specializes a VFM without shrinking it; distillation shrinks it, but through expensive teacher-guided training. Structured pruning offers a more direct route by extracting a compact subnetwork from the VFM. However, as compression gets aggressive, existing methods fail because they treat subnetwork selection and recovery as separate axes. Across layers, capacity must follow task sensitivity rather than spread uniformly. Within a layer, a removal cannot be judged by importance alone; its impact also depends on how well the lost contribution can be recovered from what remains. Selection and recovery must therefore be decided jointly. We introduce \emph{Carve VFM}, which resolves this interdependence. Given a VFM, a calibration set, and a parameter budget, it (i) searches for task-specific capacity allocations across layers; (ii) guides the search via an adaptive recovery-aware fitness criterion; and (iii) reconstructs removed behavior through iterative closed-form low-rank corrections merged into the surviving weights. The result is a dense encoder produced in minutes on a single GPU without backbone retraining. Across twelve benchmarks and four models from DINOv3 and MetaCLIP-2, \emph{Carve VFM} achieves the highest reported mean accuracy in seven of eight backbone--budget configurations among the evaluated dense structured-pruning methods. Further fine-tuning surpasses the strongest evaluated edge model (\textbf{88.1\%} vs.\ 85.8\%), and carved models cut compute and memory while delivering \textbf{1.3--2.0$\times$} higher GPU throughput than the uncompressed one. These results show that the compact model you need is inside the VFM; it only has to be \emph{carved} out.
}

\begin{document}

\maketitle

\section{Introduction}

Vision foundation models (VFMs) derive their value from generality: their representations transfer well across tasks and domains \cite{vemulapalli2023knowledge}. In deployment, however, this generality can become a practical liability \cite{aduvuku2026bridging}. A model trained to represent a broad visual world may ultimately serve a narrow task such as distinguishing cars or tissues.  We call resolving this mismatch \emph{VFM specialization}: given a pretrained VFM, a target task, and a parameter budget, derive the most accurate task-specific model that fits within that budget.

Pretraining is done once and amortized over many uses, whereas inference is paid for on every input for the model’s lifetime. In production, inference indeed accounts for a large share of ML energy use~\cite{patterson2022carbon}.

Fine-tuning techniques fit the VFM to the target task but preserve its computational cost. Parameter-efficient updates reduce training cost but still execute the full model at inference ~\cite{hu2022lora}. On the other side, knowledge distillation can yield compact student models that run faster, but it requires a costly teacher-guided training process ~\cite {hinton2015distilling, beyer2022knowledge}.

\begin{figure}[!t]
    \centering
    \includegraphics[width=.9\linewidth]{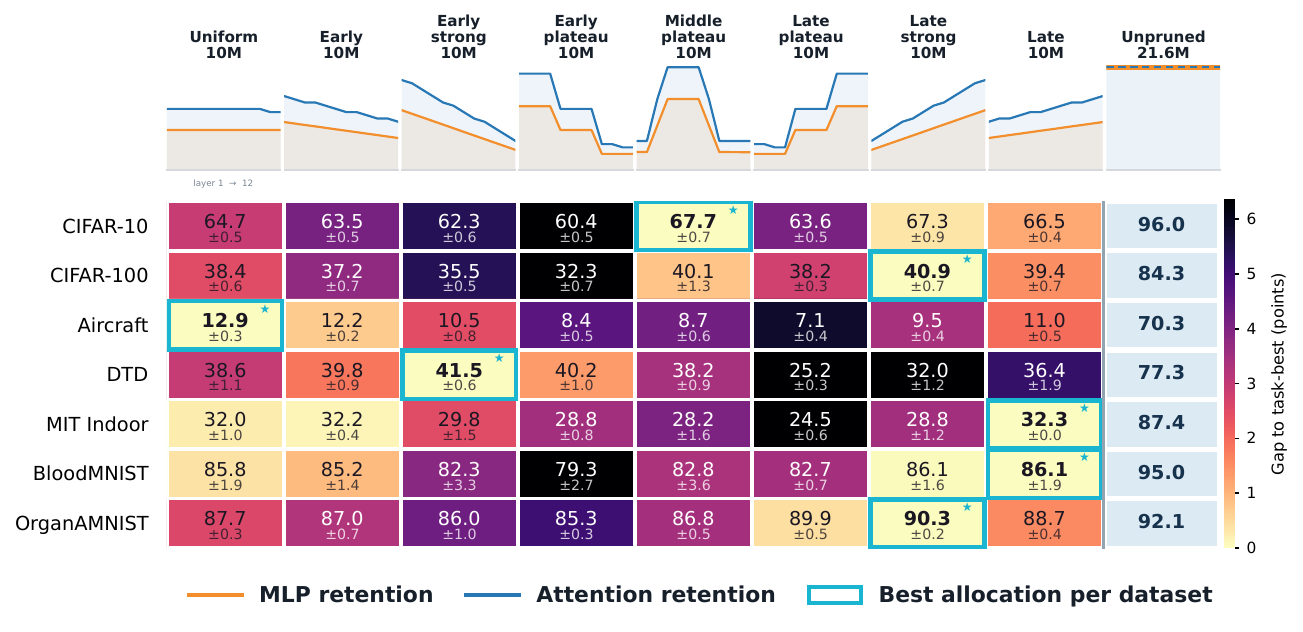}
    \caption{Task accuracy after pruning ViT-S from 21.6M to 10M parameters under different layer-wise allocations. Results are mean$\pm$std over 3 seeds. The unpruned model is included for reference.}
    \label{fig:allocation-motivation}
\end{figure}

Structured pruning offers a more direct route by removing entire components, such as MLP channels or attention heads, to produce a smaller dense model~\cite {ahsaei2026pruning}. Beyond efficiency, pruning can also improve transfer learning by eliminating pretrained components that are detrimental to a target task~\cite{liu2021transtailor}. These observations suggest that a VFM contains task-specific subnetworks hidden within its general-purpose architecture, a hypothesis supported by recent work that successfully extracts such subnetworks~\cite{zheng2022savit,shi2023upop,simoncini2025elastic}. Local importance scores can rank structures within a layer, yet they provide little guidance on how model capacity should be distributed across layers or which pruning operator is suitable for each layer. As Figure~\ref{fig:allocation-motivation} illustrates, under the same 10M parameter budget, the optimal allocation varies across tasks, while poorly allocated removals can degrade accuracy. A first challenge is therefore \emph{allocation}; specialization requires discovering a task-specific allocation to find a good subnetwork within the VFM.

At mild compression, pruning can concentrate on redundant or detrimental structures, often preserving and occasionally improving accuracy. As the budget tightens, however, useful structures can also be removed. At this point, a removal depends not only on the importance of its contribution, but also on how well that contribution can be reconstructed from the remaining structures. Recovery methods exploit this redundancy by compensating output bias or fitting low-rank corrections to the compression residual~\cite{an2024flap,liu2024eora}, yet they are typically applied only after subnetwork selection. As a result, two equal-budget subnetworks may incur similar degradation after pruning, yet differ substantially in how faithfully their lost behavior can be reconstructed. The second gap is therefore \emph{recoverability}: subnetwork selection should favor structures whose lost contributions can be recovered from the surviving model, rather than simply the least important ones.

Ideally, the search would rank subnetworks by their recovered task accuracy. In practice, this would require repeatedly extracting, repairing, and evaluating thousands of candidate subnetworks, making exhaustive evaluation infeasible. Cheap local criteria enable efficient exploration but cannot capture cross-layer interactions or the preservation of the target task's decision boundaries. Conversely, feature reconstruction alone is not enough, as faithfully reproducing pretrained representations does not guarantee task discrimination~\cite{liu2020discrimination}. The third gap is therefore \emph{evaluation}: the search requires a proxy that is both inexpensive for exploration and predictive of the recovered task accuracy that ultimately determines deployment performance.

We address these challenges with \emph{Carve VFM}, which jointly searches capacity allocations and attention removal operators under a parameter budget, using recoverability-aware local-global evaluation. The selected subnetwork is carved and repaired into a dense task-specific model. Our main contributions are:

\noindent- \textbf{Task-specific allocation.} We formulate VFM specialization as allocation under a parameter budget and introduce a unified search space over layer-wise MLP and attention capacity together with attention removal operators.

\noindent- \textbf{Adaptive local-to-global selection.} We introduce a
    recoverability-aware fitness in which cached conditional-recovery estimates
    enable inexpensive exploration, while a held-out task audit detects when
    the local ordering is unreliable and adaptively guides the search.

\noindent- \textbf{Mergeable iterative repair.} We develop a closed-form low-rank recovery that iteratively transfers useful behavior from removed structures into the surviving weights and leaves no additional inference branch.

\section{Related Work}
\label{sec:related-work}

\textbf{Compression of pretrained transformers.}
Compression reduces deployment cost while trying to preserve the pretrained knowledge. Unstructured pruning~\cite{frantar2022obc,sun2024wanda} and quantization~\cite{yuan2022ptq4vit} compress weights or numerical precision, but typically preserve tensor shapes and therefore rely on sparse or low-precision hardware for efficient execution. Structured pruning instead removes components, such as channels or blocks, directly producing smaller dense models. Recent methods have improved structure selection through joint importance estimation~\cite{zheng2022savit}, progressive mask learning~\cite{shi2023upop}, self-supervised subnetworks~\cite{simoncini2025elastic}, elastic architectures~\cite{zhu2025eavit}, and activation-aware pruning~\cite{an2024flap}. Collectively, these works demonstrate that pretrained VFMs contain compact subnetworks with strong downstream performance.

\paragraph{Allocation and recovery.}
The remaining challenge is identifying which subnetwork to extract under a given budget. Budget-aware methods such as AMC~\cite{he2018amc} and NetAdapt~\cite{yang2018netadapt} search layer-wise capacity allocations, while most pruning methods rank candidate removals using importance or reconstruction error. Recovery methods such as EoRA~\cite{liu2024eora} then compensate the selected compressed model through low-rank reconstruction. In existing approaches, however, architecture selection and recovery remain largely separate stages. As a result, pruning decisions are made without considering how well removed structure can be recovered. 
\section{Methodology: Carve VFM}
\label{sec:method}

\textbf{Problem setup.}
Let $f_\theta$ denote a pretrained VFM with $L$ transformer layers and let
$\mathcal C=\{(x_i,y_i)\}_{i=1}^{n}$ be a small calibration set for a target
task. Given a parameter budget $B$, the objective is to extract a dense
subnetwork $g$ that satisfies the deployment budget while maximizing
\mbox{downstream performance},
\begin{equation}
g^\star
=
\arg\max_{P(g)\le B}
\operatorname{Acc}_{\mathrm{val}}(g).
\label{eq:specialization-objective}
\end{equation}
Direct optimization of Equation~\eqref{eq:specialization-objective} is intractable because the space of budget-feasible
subnetworks grows combinatorially with model size, while evaluating a single candidate requires extracting, repairing, and assessing a separate model.

Previous approaches typically approximate this objective by ranking channels, or other structures according to their individual importance and pruning
the least important ones. Under aggressive compression, however, importance
is insufficient. The quality of the extracted subnetwork depends jointly on how capacity is allocated across layers, which removal operators are applied, and how well the discarded channels can be recovered from the surviving ones. Because these decisions are strongly coupled, optimizing them independently generally leads to suboptimal results.

\begin{figure}[!t]
    \centering
    \includegraphics[width=\textwidth]{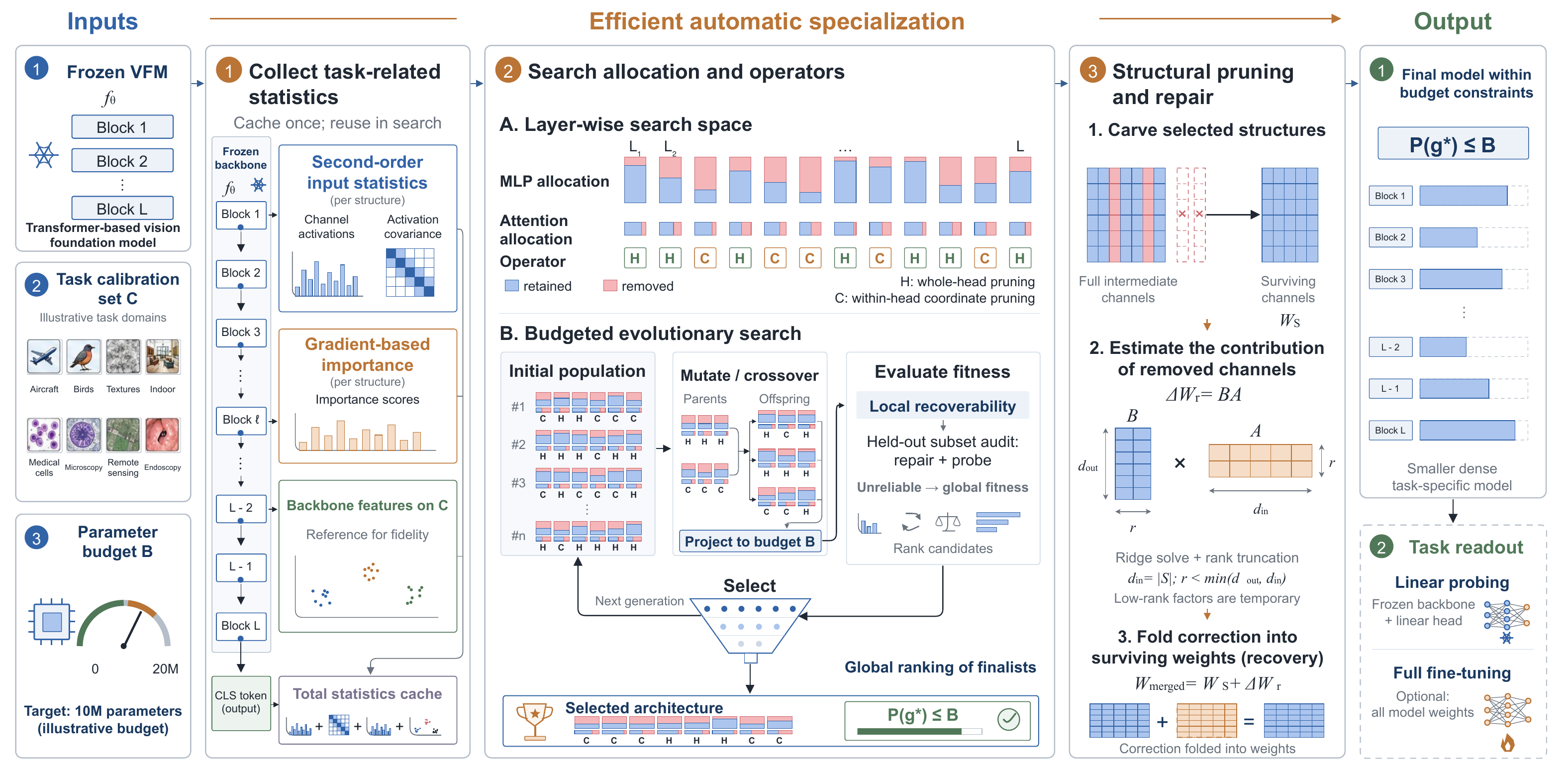}
    \caption{Overview of \emph{Carve VFM}. From a VFM, a calibration
    set, and a parameter budget $B$, the method (1) estimates removal damage,
    (2) searches layer-wise MLP retention and attention operators, and (3) carves and repairs the selected architecture. The output is a standard dense task-specific model within $B$.}
    \label{fig:method-overview}
\end{figure}

We address this interdependence with \emph{Carve VFM}, a budget-constrained architecture extraction framework that jointly optimizes allocation, removal
structure, and recoverability. Figure~\ref{fig:method-overview} summarizes its three-stage pipeline. Stage~1 estimates task-dependent removal statistics,
Stage~2 searches candidate architectures using an adaptive local-global fitness, and Stage~3 physically carves and repairs the selected architecture.
The resulting model is a small dense model within the target budget, optionally fine-tuned for better performance.

\label{sec:stage-stats}
\paragraph{Stage 1: Task-related statistics.}
A single forward/backward pass of the calibration set $\mathcal C$ through the frozen VFM caches the task-dependent statistics (Figure~\ref{fig:method-overview}, box~1). These include second-order input statistics for conditional reconstruction~\cite{frantar2022obc}, task-gradient
importance for identifying directions that most influence the downstream
objective, and reference activations for subsequent repair and held-out task evaluation. These quantities are computed only once and reused later throughout the search.

\label{sec:stage-search}
\paragraph{Stage 2: Budgeted subnetwork search.}
Stage~2 searches for the best budget-feasible subnetwork by jointly optimizing
capacity allocation across layers and the attention removal operator used in each layer. The resulting search space is explored through an evolutionary optimization process, while an adaptive local-global fitness is used for efficient evaluation. 

\paragraph{Search space.}
A candidate architecture assigns each transformer layer an independent MLP
retention ratio $s_l^{\mathrm{mlp}}$, an attention retention ratio
$s_l^{\mathrm{att}}$, and an attention removal operator
$o_l\in\{\mathrm H,\mathrm C\}$, while the overall parameter count is
constrained by the target budget $B$. The MLP ratio reduces the hidden
dimension by removing hidden channels, implemented by deleting the
corresponding rows of the MLP up-projection $W_{\mathrm{up}}$ and matching
columns of the down-projection $W_{\mathrm{down}}$.

Attention supports two complementary removal operators. \textbf{H} removes
complete attention heads. \textbf{C} instead preserves every head while
removing matched coordinates within the query/key pair ($W_Q,W_K$) and within
the value/output pair ($W_V,W_O$), with separate retained sets for the two
pairs. Consequently, H concentrates a given reduction in
fewer heads, whereas C distributes the same parameter reduction across all
heads, allowing compression even when no complete head \mbox{is expendable}.

\begin{figure}[!t]
    \centering
    \includegraphics[width=\linewidth]{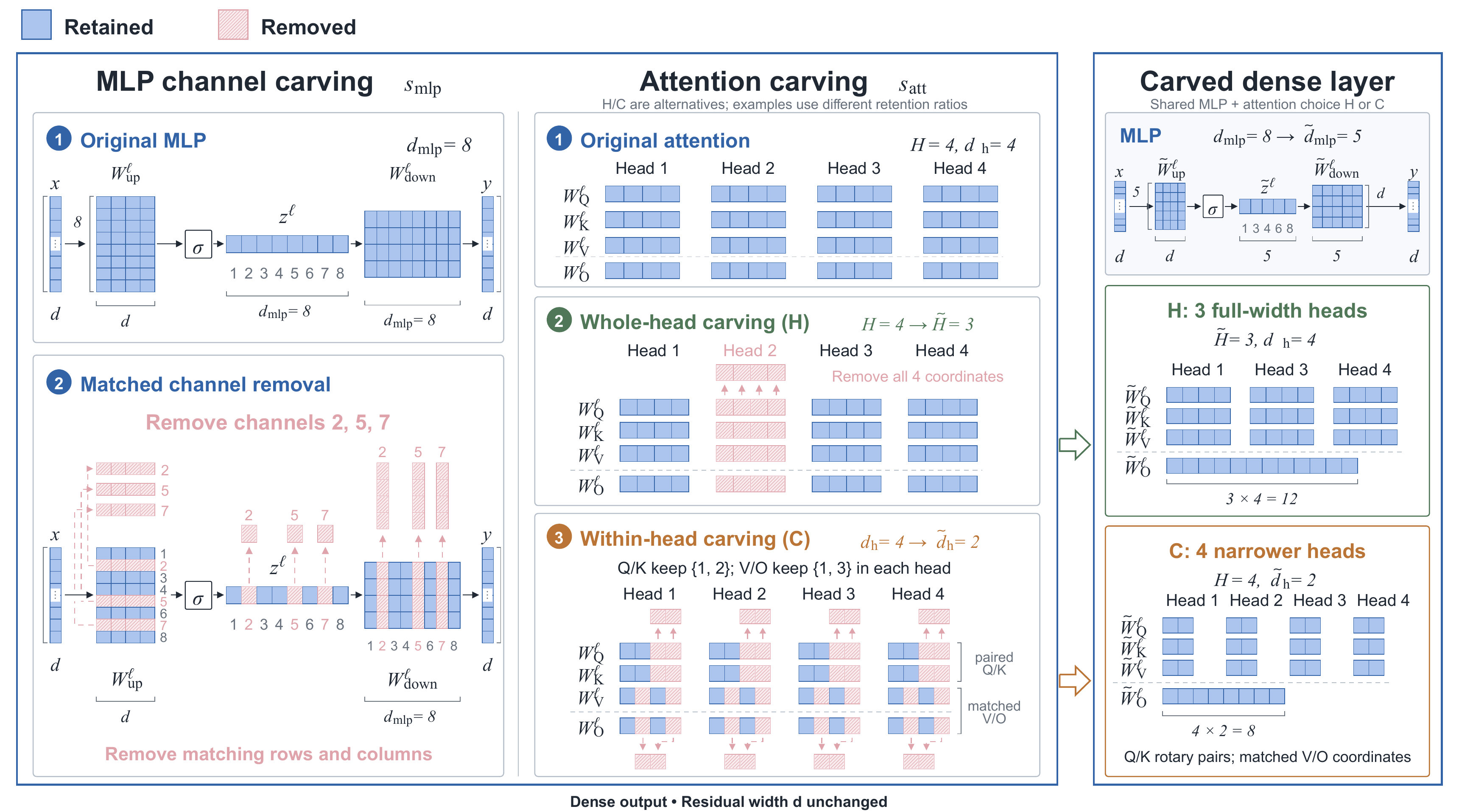}
    \caption{Structured pruning used in \emph{Carve VFM}. MLP carving removes hidden channels with matching rows and columns from its two projections. Attention removes
    either complete heads (H) or some channels within each head (C).}
    \label{fig:compression-operators}
\end{figure}

\paragraph{Search procedure.}
This mixed continuous--discrete space is explored using an evolutionary algorithm (Figure~\ref{fig:method-overview}, box~2B). Candidate subnetworks are initialized from diverse allocations, evolved through mutation and crossover, and projected onto the budget $B$ before evaluation. To make evaluations efficient, an adaptive local-global fitness is used to guide the search. Local recoverability first ranks the population, after which held-out accuracy audits a small rank-stratified subset. If the audit preserves the local ranking, exploration continues under the local fitness. Otherwise, if the rankings diverge or the audit discovers a substantially better candidate than predicted locally, the local score is deemed unreliable and held-out accuracy guides a capped continuation. In both cases, the global score ranks the final candidates.

\textit{Local fitness: recoverable damage.}
The local fitness estimates whether refitted surviving weights can reconstruct
removed functionality, unlike magnitude- or gradient-based importance. Held-out
evaluations of repaired candidates audit this search-time proxy.

Let $Z$ be a projection input, $W$ its downstream weight, and
$G=Z^\top Z/n+\lambda I$ the regularized Gram matrix. For a removable group $\mathcal I$ (an MLP channel or attention head), its local damage is
\begin{equation}
    d(\mathcal I)=\operatorname{tr}\!\left(
      W_{:\mathcal I}[G^{-1}_{\mathcal I\mathcal I}]^{-1}
      W_{:\mathcal I}^{\top}
    \right).
    \label{eq:local-damage}
\end{equation}
Here $[G^{-1}_{\mathcal I\mathcal I}]^{-1}$ is the conditional covariance of the removed group given the surviving inputs. Following OBC~\cite{frantar2022obc}, groups are removed iteratively, the surviving weights are refit, and $G^{-1}$ is updated by a Schur complement. This produces nested removal orders and cumulative damage curves $E_l^o(k)$ for every layer and operator, cached for MLP channels, attention heads, and V/O coordinates. Q/K rotary pairs are ranked separately by their perturbation of calibration logits, leaving their bilinear distortion to the global fitness. A candidate architecture is scored by Equation~\eqref{eq:local_score}, requiring neither model construction nor a forward pass.
\begin{equation}
\mathcal Q_{\mathrm{local}}(\mathbf a)
=
-\sum_l E_l^{o_l}(k_l).
\label{eq:local_score}
\end{equation}

\textit{Global fitness: task discriminability.}
Because the local score is additive across layers, it cannot capture cross-layer interactions or the preservation of task discriminability. We
therefore periodically audit a small, rank-stratified subset of candidates using held-out accuracy. After applying the repairs, we fit a regularized linear probe on one part of each of $T$ stratified calibration splits of $C$ and evaluate it on the others. This held-out accuracy is the search-time proxy for Equation~\eqref{eq:specialization-objective}. 

\label{sec:stage-carve}
\paragraph{Stage 3: Carving and repair.}
Stage~2 specifies a budget-feasible subnetwork; Stage~3 realizes its smaller tensor
dimensions and repairs the surviving weights without restoring any removed
dimension. The result therefore remains a dense model within the target budget $B$.

\textit{Carving: making the architecture smaller.}
A structural feature is usually produced by one projection and consumed by another. Carving therefore removes the same index from both sides of this producer--consumer path. For an MLP layer, keeping the hidden-channel set $\mathcal S_l^{\mathrm{mlp}}$ gives
an ordinary MLP with smaller hidden width $|\mathcal S_l^{\mathrm{mlp}}|$  as shown below.
\begin{equation}
\widetilde W_{\mathrm{up}}^l
=W_{\mathrm{up}}^l[\mathcal S_l^{\mathrm{mlp}},:],\qquad
\widetilde W_{\mathrm{down}}^l
=W_{\mathrm{down}}^l[:,\mathcal S_l^{\mathrm{mlp}}].
\label{eq}
\end{equation}

Attention follows the same paired-indexing principle, with the survivor sets
determined by the selected operator $o_l$:
\begin{equation}
\begin{aligned}
\widetilde W_p^l
&=W_p^l[\mathcal S_p^l,:],
&&p\in\{Q,K,V\},\\
\widetilde W_O^l
&=W_O^l[:,\mathcal S_O^l].
\end{aligned}
\label{eq:attn-carve}
\end{equation}
For whole-head carving ($o_l=\mathrm H$), all projections share the same
survivor set, corresponding to complete retained heads. For column carving
($o_l=\mathrm C$), Q/K and V/O use separate survivor sets, preserving rotary
pairs and matched value/output coordinates, respectively. See
Figure~\ref{fig:compression-operators} for an illustration of the two carving operators. In
all cases, the biases are sliced accordingly, producing
physically smaller dense projections.

\textit{Repair: recovering removed behavior.}
\looseness=-1
Carving removes entire structures and therefore part of each layer's output.
We recover this lost behavior by predicting the removed output from the
surviving activations. Similar to calibration-based low-rank
recovery~\cite{liu2024eora}, but formulated directly in the reduced coordinate space, the resulting correction can be merged exactly into the carved model.

Partition a linear projection into surviving indices $\mathcal S$ and removed
indices $\mathcal I$. The output lost by removing $\mathcal I$ is
$Y_{\mathcal I}=Z_{\mathcal I}W_{\mathcal I}^{\top}$. We reconstruct this
contribution by solving
\begin{equation}
 \begin{aligned}
    \Delta W^{\star}
      &=\arg\min_{\Delta W}
        \|Z_{\mathcal S}(\Delta W)^{\top}-Y_{\mathcal I}\|_F^2
        +\lambda\|\Delta W\|_F^2\\
      &=Y_{\mathcal I}^{\top}Z_{\mathcal S}
        (Z_{\mathcal S}^{\top}Z_{\mathcal S}+\lambda I)^{-1},
    \qquad
    \Delta W_r=\operatorname{SVD}_r(\Delta W^{\star}).
 \end{aligned}
    \label{eq:compensation-solve}
\end{equation}
The rank-$r$ reconstruction $\Delta W_r=U_r\Sigma_rV_r^\top$, using the leading
singular triplets, regularizes the solve without changing the reduced shape.
The correction merges as
$
W_{\mathrm{merged}}
=
W_{\mathcal S}
+
\Delta W_r,
$
with no additional inference parameters or operations. The same formulation
repairs the MLP down-projection, the V/O projections under operator C, and the
retained-head output projection under operator H; residual means are folded
into the existing biases. Q/K under operator C is repaired separately because
its effect is bilinear in the attention logits.

Unlike post-hoc recovery methods such as EoRA~\cite{liu2024eora}, Carve VFM
integrates repair into architecture search. Each cached curve refits the
surviving weights and updates $G^{-1}$ between group removals. The curves
guide Stage~2 local ranking; held-out evaluations of repaired candidates audit
that ranking before Stage~3 carves and repairs the selected architecture.

\paragraph{Output: dense subnetwork and optional fine-tuning.}
\label{sec:stage-output}
After search, pruning and recovery, the resulting carved model satisfies
$P(\mathbf a^\star)\leq B$ (Figure~\ref{fig:method-overview}, output). The primary protocol freezes this carved model and trains a linear head, while the full-fine-tuning variant optimizes it end-to-end. Both yield an ordinary dense transformer with task-specific widths.

\section{Experiments}
\label{sec:experiments}

\textbf{Experimental Setup}

\paragraph{Benchmarks and source models.}
We evaluate four source models: DINOv3 ViT-S (21.6M), ViT-B (85.7M), and ViT-L (303.1M)~\cite{simeoni2025dinov3}, and the image--text-pretrained MetaCLIP-2 ViT-B (85.8M)~\cite{chuang2025metaclip2}. The benchmark spans twelve tasks: CIFAR-10, CIFAR-100, and Caltech-256 for general recognition; FGVC-Aircraft, CUB-200, Stanford Cars, and Oxford-IIIT Pets for fine-grained recognition; DTD and MIT Indoor-67 for texture/scene recognition; and BloodMNIST,
PneumoniaMNIST, and OrganAMNIST for medical recognition
~\cite{krizhevsky2009cifar,griffin2007caltech256,maji2013aircraft,wah2011cub,krause2013cars,parkhi2012pets,cimpoi2014dtd,quattoni2009indoor,yang2023medmnist}.
We use standard train/test partitions, except for Caltech-256, where each seed
defines an 80/20 split, and report accuracy.

\paragraph{Protocol and comparisons.}
For each task, 512 training samples form the calibration set; the test set is never used for architecture selection. After carving and repair, the backbone is frozen, and a fresh linear classifier is trained on the full training set for ten epochs. Results are mean and standard deviation over three seeds. The historical comparison uses common source checkpoints, data splits, calibration sets, and parameter budgets, with Random, BlockDrop~\cite{wu2018blockdrop}, SNIP-Magnitude~\cite{lee2019snip}, SnapViT~\cite{simoncini2025elastic}, FLAP~\cite{an2024flap}, SVD-LLM~\cite{wang2025svd}, and CORP~\cite{zhang2026corp}. Table~\ref{tab:completed-baselines} additionally includes SliceGPT~\cite{ashkboos2024slicegpt} and available OSSCAR~\cite{meng2024osscar} results at the mild and tight budgets. Additional scoring baselines and intermediate-budget results are provided in the supplement, together with search, recovery, optimization, and hardware details.

% Generated by scripts/build_completed_baseline_tables.py. Do not edit numbers by hand.
% Requires xcolor[table], colortbl, array, tabularx, makecell, hhline, multirow, graphicx. One page using adjustbox; template margins unchanged. Requires adjustbox.
\begingroup
\definecolor{TableInk}{HTML}{25392E}
\definecolor{TableSide}{HTML}{EEF2E9}
\definecolor{TableHeat}{HTML}{78B84A}
\definecolor{TableGrid}{HTML}{CAD5D0}
\definecolor{TableMuted}{HTML}{66736B}
\definecolor{TableRef}{HTML}{EDF0F2}
\definecolor{TableMissing}{HTML}{F8F8F8}
\definecolor{TableCarve}{HTML}{F3DFC4}
\newcommand{\Score}[4]{\cellcolor{#3}#1}
\newcommand{\ScoreBest}[4]{\cellcolor{#3}\textbf{#1}}
\newcommand{\HeatKey}[2]{{\setlength{\fboxsep}{1pt}\colorbox{#1}{\strut #2}}}
\newlength{\TabLead}\setlength{\TabLead}{7.5pt}% row height
\newcommand{\TopStrut}{\rule{0pt}{\dimexpr0.7\TabLead+2pt\relax}}
\newcommand{\BotStrut}{\rule[\dimexpr-0.3\TabLead-1.5pt\relax]{0pt}{0pt}}
\newcommand{\MissingScore}[1]{\cellcolor{TableMissing}\textcolor{TableMuted}{--}}
\begin{table*}[p]
\caption[Linear probing accuracy at mild and tight compression ratios]{Linear probing accuracy (\%) at mild and tight compression ratios, mean over three seeds.
Shading (\HeatKey{TableHeat!0}{low}\,\HeatKey{TableHeat!25}{mid}\,\HeatKey{TableHeat!50}{high}) is normalized within each block; gray: full model or unavailable;
\textbf{bold}: best per source and budget (ties included). Standard deviations, additional budgets, and method details are reported in the appendix.}
\label{tab:completed-baselines}
\begin{center}
\fontsize{7}{\TabLead}\selectfont
\setlength{\tabcolsep}{1.5pt}
\renewcommand{\arraystretch}{1}
\setlength{\arrayrulewidth}{0.6pt}
\newcolumntype{S}{>{\centering\arraybackslash}X}
\arrayrulecolor{white}
\begin{adjustbox}{max totalsize={\textwidth}{0.92\textheight},center}
\begin{tabularx}{\textwidth}{@{}>{\columncolor{TableSide}}p{8pt}|>{\columncolor{TableSide}}p{27pt}@{\hspace{3pt}}l@{\hspace{3pt}}S|S|S!{\color{white}\vrule width 2pt}S|S|S|S!{\color{white}\vrule width 2pt}S|S!{\color{white}\vrule width 2pt}S|S|S!{\color{white}\vrule width 2pt}S@{}}
\arrayrulecolor{TableInk}\Xhline{0.9pt}\arrayrulecolor{white}
\multicolumn{2}{c|}{\textbf{Model}} & \textbf{Method} & \multicolumn{3}{c!{\color{white}\vrule width 2pt}}{\textbf{General}} & \multicolumn{4}{c!{\color{white}\vrule width 2pt}}{\textbf{Fine-grained}} & \multicolumn{2}{c!{\color{white}\vrule width 2pt}}{\textbf{Texture/scene}} & \multicolumn{3}{c!{\color{white}\vrule width 2pt}}{\textbf{Medical}} & \textbf{Mean} \\
 & \textbf{Size} & & C10 & C100 & Cal. & Air. & CUB & Cars & Pets & DTD & MIT & Blood & Pneu. & Org.A & \\
\arrayrulecolor{TableGrid}\Xhline{0.4pt}\arrayrulecolor{white}
 & \makebox[27pt][c]{21.6M} & \cellcolor{TableRef}\TopStrut\BotStrut Full & \Score{96.0}{0.1}{TableRef}{} & \Score{84.3}{0.2}{TableRef}{} & \Score{92.6}{0.2}{TableRef}{} & \Score{70.3}{0.1}{TableRef}{} & \Score{83.6}{0.1}{TableRef}{} & \Score{88.8}{0.3}{TableRef}{} & \Score{94.0}{0.4}{TableRef}{} & \Score{77.3}{0.1}{TableRef}{} & \Score{87.4}{0.2}{TableRef}{} & \Score{95.0}{0.1}{TableRef}{} & \Score{84.8}{0.2}{TableRef}{} & \Score{92.1}{0.1}{TableRef}{} & \Score{87.2}{0.0}{TableRef}{} \\
\arrayrulecolor{TableInk}\hhline{~---------------}\arrayrulecolor{white}
 &  & \TopStrut BlockDrop & \Score{65.6}{0.1}{TableHeat!20}{} & \Score{40.6}{0.5}{TableHeat!19}{} & \Score{44.3}{1.1}{TableHeat!22}{} & \Score{14.5}{0.3}{TableHeat!8}{} & \Score{12.1}{2.7}{TableHeat!6}{} & \Score{11.1}{0.2}{TableHeat!5}{} & \Score{21.2}{0.5}{TableHeat!9}{} & \Score{42.7}{0.3}{TableHeat!23}{} & \Score{40.2}{3.1}{TableHeat!21}{} & \Score{91.6}{0.1}{TableHeat!28}{} & \Score{81.6}{0.9}{TableHeat!0}{} & \Score{89.3}{0.2}{TableHeat!30}{} & \Score{46.2}{0.3}{TableHeat!15}{} \\
 &  & SNIP--Mag. & \Score{46.4}{0.8}{TableHeat!0}{} & \Score{14.9}{0.5}{TableHeat!0}{} & \Score{9.8}{0.9}{TableHeat!0}{} & \Score{4.1}{0.7}{TableHeat!0}{} & \Score{2.2}{0.5}{TableHeat!0}{} & \Score{1.8}{0.3}{TableHeat!0}{} & \Score{5.5}{0.8}{TableHeat!0}{} & \Score{15.5}{1.7}{TableHeat!0}{} & \Score{8.6}{1.5}{TableHeat!0}{} & \Score{85.1}{0.7}{TableHeat!0}{} & \Score{81.6}{0.3}{TableHeat!0}{} & \Score{82.6}{1.9}{TableHeat!0}{} & \Score{29.8}{0.9}{TableHeat!0}{} \\
 & \makebox[27pt][c]{21.6M} & SnapViT & \Score{52.5}{0.3}{TableHeat!6}{} & \Score{26.3}{0.4}{TableHeat!9}{} & \Score{31.3}{1.1}{TableHeat!14}{} & \Score{9.2}{0.2}{TableHeat!4}{} & \Score{6.2}{0.1}{TableHeat!3}{} & \Score{3.9}{0.2}{TableHeat!1}{} & \Score{13.9}{0.4}{TableHeat!5}{} & \Score{28.0}{0.5}{TableHeat!10}{} & \Score{27.2}{0.6}{TableHeat!12}{} & \Score{86.7}{0.4}{TableHeat!7}{} & \Score{82.4}{0.5}{TableHeat!5}{} & \Score{87.9}{0.1}{TableHeat!23}{} & \Score{38.0}{0.4}{TableHeat!7}{} \\
 & \makebox[27pt][c]{$\downarrow$} & FLAP & \Score{71.1}{0.1}{TableHeat!26}{} & \Score{52.1}{1.2}{TableHeat!28}{} & \Score{57.6}{0.6}{TableHeat!31}{} & \Score{33.2}{3.2}{TableHeat!22}{} & \Score{25.2}{0.3}{TableHeat!15}{} & \Score{39.7}{1.2}{TableHeat!22}{} & \Score{53.2}{2.3}{TableHeat!28}{} & \Score{54.9}{0.8}{TableHeat!33}{} & \Score{51.7}{1.7}{TableHeat!29}{} & \Score{94.3}{0.2}{TableHeat!39}{} & \Score{83.9}{1.8}{TableHeat!15}{} & \Score{92.7}{0.2}{TableHeat!45}{} & \Score{59.1}{0.1}{TableHeat!26}{} \\
 & \makebox[27pt][c]{\textbf{14M}} & SVD-LLM & \Score{92.5}{0.1}{TableHeat!48}{} & \Score{75.0}{0.2}{TableHeat!45}{} & \Score{76.4}{1.1}{TableHeat!43}{} & \Score{51.2}{0.9}{TableHeat!35}{} & \Score{61.6}{1.0}{TableHeat!39}{} & \Score{63.0}{4.3}{TableHeat!35}{} & \Score{79.4}{0.8}{TableHeat!43}{} & \Score{65.1}{0.8}{TableHeat!42}{} & \Score{74.7}{1.6}{TableHeat!44}{} & \Score{95.7}{0.1}{TableHeat!45}{} & \Score{88.9}{1.7}{TableHeat!48}{} & \Score{92.7}{0.2}{TableHeat!45}{} & \Score{76.3}{0.4}{TableHeat!42}{} \\
 &  & CORP & \Score{94.7}{0.1}{TableHeat!50}{} & \ScoreBest{81.1}{0.2}{TableHeat!50}{} & \Score{85.5}{0.4}{TableHeat!49}{} & \ScoreBest{70.9}{0.3}{TableHeat!50}{} & \Score{72.9}{0.5}{TableHeat!46}{} & \Score{83.0}{0.5}{TableHeat!47}{} & \Score{90.3}{0.0}{TableHeat!49}{} & \Score{57.9}{0.2}{TableHeat!36}{} & \Score{81.9}{0.3}{TableHeat!49}{} & \ScoreBest{96.8}{0.1}{TableHeat!50}{} & \Score{88.0}{1.0}{TableHeat!42}{} & \ScoreBest{93.9}{0.2}{TableHeat!50}{} & \Score{83.1}{0.0}{TableHeat!48}{} \\
 &  & SliceGPT & \Score{92.4}{0.2}{TableHeat!48}{} & \Score{75.7}{0.3}{TableHeat!46}{} & \Score{41.0}{1.1}{TableHeat!20}{} & \Score{7.9}{0.6}{TableHeat!3}{} & \Score{7.5}{0.6}{TableHeat!3}{} & \Score{5.4}{0.5}{TableHeat!2}{} & \Score{26.5}{0.9}{TableHeat!12}{} & \Score{59.5}{1.1}{TableHeat!37}{} & \Score{51.1}{1.3}{TableHeat!28}{} & \Score{95.3}{0.4}{TableHeat!44}{} & \Score{88.3}{1.0}{TableHeat!44}{} & \Score{92.7}{0.1}{TableHeat!45}{} & \Score{53.6}{0.3}{TableHeat!21}{} \\
 &  & OSSCAR & \Score{91.3}{0.1}{TableHeat!46}{} & \Score{73.5}{0.3}{TableHeat!44}{} & \Score{75.1}{0.8}{TableHeat!43}{} & \Score{57.5}{1.1}{TableHeat!40}{} & \Score{59.7}{0.1}{TableHeat!37}{} & \Score{75.6}{0.3}{TableHeat!42}{} & \Score{81.6}{0.3}{TableHeat!44}{} & \Score{69.8}{0.9}{TableHeat!45}{} & \Score{69.1}{6.1}{TableHeat!40}{} & \Score{96.1}{0.1}{TableHeat!47}{} & \Score{88.9}{0.8}{TableHeat!48}{} & \Score{93.0}{0.2}{TableHeat!46}{} & \Score{77.6}{0.6}{TableHeat!43}{} \\
 &  & \cellcolor{TableCarve}\BotStrut\textbf{Carve VFM} & \ScoreBest{94.8}{0.5}{TableHeat!50}{} & \Score{80.5}{0.5}{TableHeat!50}{} & \ScoreBest{86.5}{0.9}{TableHeat!50}{} & \Score{70.1}{0.4}{TableHeat!49}{} & \ScoreBest{79.1}{0.9}{TableHeat!50}{} & \ScoreBest{88.7}{0.5}{TableHeat!50}{} & \ScoreBest{91.3}{0.3}{TableHeat!50}{} & \ScoreBest{75.2}{0.3}{TableHeat!50}{} & \ScoreBest{83.9}{0.4}{TableHeat!50}{} & \Score{96.7}{0.2}{TableHeat!50}{} & \ScoreBest{89.2}{1.6}{TableHeat!50}{} & \Score{93.6}{0.0}{TableHeat!49}{} & \ScoreBest{85.8}{0.1}{TableHeat!50}{} \\
\arrayrulecolor{TableInk}\hhline{~---------------}\arrayrulecolor{white}
 &  & \TopStrut BlockDrop & \Score{51.9}{0.4}{TableHeat!14}{} & \Score{25.7}{0.2}{TableHeat!15}{} & \Score{19.8}{0.5}{TableHeat!18}{} & \Score{7.5}{0.2}{TableHeat!13}{} & \Score{5.4}{0.2}{TableHeat!12}{} & \Score{3.8}{0.2}{TableHeat!15}{} & \Score{12.3}{0.1}{TableHeat!15}{} & \Score{19.1}{0.0}{TableHeat!7}{} & \Score{19.3}{2.5}{TableHeat!18}{} & \Score{69.9}{0.7}{TableHeat!0}{} & \ScoreBest{85.4}{0.7}{TableHeat!50}{} & \Score{85.2}{0.2}{TableHeat!34}{} & \Score{33.8}{0.1}{TableHeat!14}{} \\
 &  & SNIP--Mag. & \Score{42.0}{2.5}{TableHeat!0}{} & \Score{14.2}{0.2}{TableHeat!0}{} & \Score{9.1}{0.8}{TableHeat!0}{} & \Score{2.4}{0.3}{TableHeat!0}{} & \Score{1.3}{0.0}{TableHeat!0}{} & \Score{1.0}{0.1}{TableHeat!0}{} & \Score{4.4}{0.6}{TableHeat!0}{} & \Score{15.1}{3.2}{TableHeat!0}{} & \Score{10.2}{3.8}{TableHeat!0}{} & \Score{76.1}{2.5}{TableHeat!13}{} & \Score{81.4}{1.8}{TableHeat!13}{} & \Score{74.2}{2.3}{TableHeat!0}{} & \Score{27.6}{1.5}{TableHeat!0}{} \\
 & \makebox[27pt][c]{21.6M} & SnapViT & \Score{44.0}{0.9}{TableHeat!3}{} & \Score{18.6}{0.4}{TableHeat!6}{} & \Score{16.5}{0.7}{TableHeat!13}{} & \Score{4.5}{0.4}{TableHeat!6}{} & \Score{3.8}{0.2}{TableHeat!7}{} & \Score{2.2}{0.1}{TableHeat!7}{} & \Score{7.1}{0.4}{TableHeat!5}{} & \Score{17.7}{0.1}{TableHeat!5}{} & \Score{15.8}{0.8}{TableHeat!11}{} & \Score{72.4}{0.4}{TableHeat!5}{} & \Score{82.3}{1.0}{TableHeat!21}{} & \Score{80.4}{0.3}{TableHeat!19}{} & \Score{30.4}{0.5}{TableHeat!6}{} \\
 & \makebox[27pt][c]{$\downarrow$} & FLAP & \Score{53.8}{0.7}{TableHeat!16}{} & \Score{26.7}{0.2}{TableHeat!17}{} & \Score{16.9}{0.5}{TableHeat!13}{} & \Score{4.8}{0.8}{TableHeat!6}{} & \Score{4.6}{0.1}{TableHeat!10}{} & \Score{2.2}{0.2}{TableHeat!7}{} & \Score{7.7}{1.2}{TableHeat!6}{} & \Score{23.2}{0.5}{TableHeat!14}{} & \Score{15.3}{0.4}{TableHeat!10}{} & \Score{73.8}{0.9}{TableHeat!8}{} & \Score{80.0}{0.6}{TableHeat!0}{} & \Score{83.1}{0.7}{TableHeat!27}{} & \Score{32.7}{0.0}{TableHeat!12}{} \\
 & \makebox[27pt][c]{\textbf{6M}} & SVD-LLM & \Score{75.1}{0.4}{TableHeat!45}{} & \Score{48.0}{0.3}{TableHeat!45}{} & \Score{19.3}{0.3}{TableHeat!17}{} & \Score{4.4}{0.2}{TableHeat!5}{} & \Score{4.6}{0.1}{TableHeat!10}{} & \Score{2.5}{0.0}{TableHeat!8}{} & \Score{10.7}{0.6}{TableHeat!12}{} & \Score{28.0}{1.2}{TableHeat!22}{} & \Score{21.1}{0.3}{TableHeat!22}{} & \Score{90.1}{0.5}{TableHeat!43}{} & \Score{83.5}{2.1}{TableHeat!32}{} & \Score{89.4}{0.2}{TableHeat!47}{} & \Score{39.7}{0.3}{TableHeat!28}{} \\
 &  & CORP & \Score{57.1}{0.5}{TableHeat!21}{} & \Score{29.7}{0.4}{TableHeat!20}{} & \Score{20.4}{0.6}{TableHeat!19}{} & \Score{5.8}{0.4}{TableHeat!9}{} & \Score{5.8}{0.3}{TableHeat!13}{} & \Score{2.7}{0.1}{TableHeat!9}{} & \Score{10.1}{0.2}{TableHeat!11}{} & \Score{16.0}{0.7}{TableHeat!2}{} & \Score{14.6}{1.2}{TableHeat!9}{} & \Score{79.7}{1.0}{TableHeat!21}{} & \Score{82.1}{2.8}{TableHeat!19}{} & \Score{86.5}{0.9}{TableHeat!38}{} & \Score{34.2}{0.2}{TableHeat!15}{} \\
 &  & SliceGPT & \ScoreBest{78.4}{0.3}{TableHeat!50}{} & \ScoreBest{52.0}{0.4}{TableHeat!50}{} & \Score{26.4}{0.7}{TableHeat!30}{} & \Score{7.6}{0.4}{TableHeat!14}{} & \Score{4.3}{0.6}{TableHeat!9}{} & \Score{3.0}{0.1}{TableHeat!11}{} & \Score{19.5}{0.5}{TableHeat!28}{} & \Score{25.2}{0.9}{TableHeat!18}{} & \Score{18.7}{0.5}{TableHeat!17}{} & \Score{80.4}{0.9}{TableHeat!22}{} & \Score{80.7}{1.5}{TableHeat!6}{} & \Score{90.1}{0.2}{TableHeat!49}{} & \Score{40.5}{0.1}{TableHeat!30}{} \\
 &  & OSSCAR & \Score{57.7}{0.7}{TableHeat!22}{} & \Score{31.4}{0.2}{TableHeat!23}{} & \Score{27.7}{1.8}{TableHeat!32}{} & \Score{12.4}{0.6}{TableHeat!26}{} & \Score{10.9}{0.2}{TableHeat!28}{} & \Score{5.4}{0.3}{TableHeat!24}{} & \Score{18.5}{0.8}{TableHeat!26}{} & \Score{41.6}{1.0}{TableHeat!46}{} & \Score{24.5}{0.4}{TableHeat!29}{} & \Score{89.7}{0.3}{TableHeat!43}{} & \Score{82.2}{1.0}{TableHeat!20}{} & \Score{86.2}{0.4}{TableHeat!37}{} & \Score{40.7}{0.1}{TableHeat!30}{} \\
\multirow{-19}{8pt}{\centering\rotatebox{90}{\textbf{DINOv3 ViT-S/16}}} &  & \cellcolor{TableCarve}\BotStrut\textbf{Carve VFM} & \Score{77.1}{2.0}{TableHeat!48}{} & \Score{51.3}{6.0}{TableHeat!49}{} & \ScoreBest{38.4}{1.2}{TableHeat!50}{} & \ScoreBest{21.4}{2.9}{TableHeat!50}{} & \ScoreBest{18.5}{4.8}{TableHeat!50}{} & \ScoreBest{10.2}{1.7}{TableHeat!50}{} & \ScoreBest{31.1}{7.1}{TableHeat!50}{} & \ScoreBest{43.8}{1.7}{TableHeat!50}{} & \ScoreBest{34.8}{1.8}{TableHeat!50}{} & \ScoreBest{93.2}{0.9}{TableHeat!50}{} & \Score{83.9}{0.5}{TableHeat!36}{} & \ScoreBest{90.4}{1.0}{TableHeat!50}{} & \ScoreBest{49.5}{0.7}{TableHeat!50}{} \\
\arrayrulecolor{TableInk}\Xhline{0.9pt}\arrayrulecolor{white}
 & \makebox[27pt][c]{85.7M} & \cellcolor{TableRef}\TopStrut\BotStrut Full & \Score{98.2}{0.1}{TableRef}{} & \Score{88.9}{0.1}{TableRef}{} & \Score{95.9}{0.3}{TableRef}{} & \Score{81.8}{0.2}{TableRef}{} & \Score{89.2}{0.1}{TableRef}{} & \Score{94.3}{0.2}{TableRef}{} & \Score{95.9}{0.2}{TableRef}{} & \Score{80.0}{0.2}{TableRef}{} & \Score{89.7}{0.5}{TableRef}{} & \Score{95.7}{0.1}{TableRef}{} & \Score{83.8}{0.2}{TableRef}{} & \Score{93.7}{0.1}{TableRef}{} & \Score{90.6}{0.0}{TableRef}{} \\
\arrayrulecolor{TableInk}\hhline{~---------------}\arrayrulecolor{white}
 &  & \TopStrut BlockDrop & \Score{73.7}{0.4}{TableHeat!0}{} & \Score{48.5}{0.5}{TableHeat!3}{} & \Score{55.1}{0.3}{TableHeat!8}{} & \Score{26.8}{0.4}{TableHeat!5}{} & \Score{12.6}{0.2}{TableHeat!0}{} & \Score{15.9}{0.3}{TableHeat!3}{} & \Score{23.3}{0.1}{TableHeat!0}{} & \Score{55.4}{5.1}{TableHeat!15}{} & \Score{54.7}{0.3}{TableHeat!11}{} & \Score{94.8}{0.3}{TableHeat!7}{} & \Score{84.0}{1.5}{TableHeat!0}{} & \Score{89.8}{0.2}{TableHeat!0}{} & \Score{52.9}{0.4}{TableHeat!3}{} \\
 &  & SNIP--Mag. & \Score{74.3}{1.9}{TableHeat!1}{} & \Score{46.3}{1.3}{TableHeat!0}{} & \Score{47.9}{2.6}{TableHeat!0}{} & \Score{22.9}{1.4}{TableHeat!2}{} & \Score{14.5}{1.3}{TableHeat!1}{} & \Score{10.3}{0.4}{TableHeat!0}{} & \Score{22.6}{4.6}{TableHeat!0}{} & \Score{43.8}{2.0}{TableHeat!0}{} & \Score{44.5}{4.4}{TableHeat!0}{} & \Score{94.4}{0.9}{TableHeat!0}{} & \Score{88.1}{0.9}{TableHeat!39}{} & \Score{92.9}{0.3}{TableHeat!33}{} & \Score{50.2}{1.8}{TableHeat!0}{} \\
 & \makebox[27pt][c]{85.7M} & SnapViT & \Score{75.7}{0.4}{TableHeat!4}{} & \Score{52.1}{0.3}{TableHeat!7}{} & \Score{60.1}{0.8}{TableHeat!13}{} & \Score{20.8}{0.2}{TableHeat!0}{} & \Score{13.5}{0.3}{TableHeat!1}{} & \Score{15.7}{0.2}{TableHeat!3}{} & \Score{33.5}{0.1}{TableHeat!7}{} & \Score{45.5}{0.6}{TableHeat!2}{} & \Score{45.2}{1.3}{TableHeat!1}{} & \Score{95.2}{0.3}{TableHeat!14}{} & \Score{87.0}{0.7}{TableHeat!29}{} & \Score{92.0}{0.3}{TableHeat!23}{} & \Score{53.0}{0.4}{TableHeat!3}{} \\
 & \makebox[27pt][c]{$\downarrow$} & FLAP & \Score{83.9}{0.4}{TableHeat!22}{} & \Score{65.5}{0.0}{TableHeat!24}{} & \Score{71.4}{0.4}{TableHeat!26}{} & \Score{66.5}{0.7}{TableHeat!37}{} & \Score{59.0}{2.5}{TableHeat!31}{} & \Score{82.0}{0.7}{TableHeat!43}{} & \Score{74.4}{0.2}{TableHeat!35}{} & \Score{71.4}{1.0}{TableHeat!37}{} & \Score{74.7}{1.2}{TableHeat!33}{} & \Score{96.8}{0.4}{TableHeat!41}{} & \Score{84.6}{3.8}{TableHeat!6}{} & \Score{94.0}{0.3}{TableHeat!45}{} & \Score{77.0}{0.3}{TableHeat!33}{} \\
 & \makebox[27pt][c]{\textbf{60M}} & SVD-LLM & \Score{97.0}{0.2}{TableHeat!49}{} & \Score{84.9}{0.2}{TableHeat!48}{} & \Score{93.2}{0.1}{TableHeat!49}{} & \Score{80.5}{0.3}{TableHeat!48}{} & \Score{86.5}{0.2}{TableHeat!49}{} & \Score{93.7}{0.2}{TableHeat!50}{} & \Score{95.3}{0.2}{TableHeat!50}{} & \Score{79.1}{0.3}{TableHeat!47}{} & \Score{88.5}{0.4}{TableHeat!48}{} & \Score{97.1}{0.2}{TableHeat!47}{} & \Score{87.2}{2.7}{TableHeat!31}{} & \Score{93.7}{0.1}{TableHeat!41}{} & \Score{89.7}{0.2}{TableHeat!49}{} \\
 &  & CORP & \Score{96.4}{1.2}{TableHeat!48}{} & \Score{85.3}{0.3}{TableHeat!49}{} & \Score{93.7}{0.3}{TableHeat!50}{} & \Score{81.7}{0.5}{TableHeat!49}{} & \Score{80.7}{0.2}{TableHeat!45}{} & \Score{88.2}{0.1}{TableHeat!47}{} & \Score{94.6}{0.2}{TableHeat!49}{} & \Score{59.0}{4.8}{TableHeat!20}{} & \Score{89.0}{0.3}{TableHeat!49}{} & \ScoreBest{97.3}{0.3}{TableHeat!50}{} & \Score{87.0}{3.2}{TableHeat!29}{} & \Score{94.2}{0.2}{TableHeat!47}{} & \Score{87.3}{0.7}{TableHeat!46}{} \\
 &  & SliceGPT & \Score{96.6}{0.1}{TableHeat!48}{} & \Score{83.3}{0.3}{TableHeat!46}{} & \Score{89.6}{0.3}{TableHeat!45}{} & \Score{78.1}{0.7}{TableHeat!46}{} & \Score{83.8}{0.3}{TableHeat!47}{} & \Score{91.8}{0.0}{TableHeat!49}{} & \Score{93.4}{0.2}{TableHeat!48}{} & \Score{76.6}{0.3}{TableHeat!44}{} & \Score{87.2}{1.0}{TableHeat!47}{} & \Score{96.0}{0.2}{TableHeat!28}{} & \Score{87.3}{2.5}{TableHeat!32}{} & \Score{93.5}{0.2}{TableHeat!40}{} & \Score{88.1}{0.2}{TableHeat!47}{} \\
 &  & OSSCAR & \Score{97.2}{0.1}{TableHeat!50}{} & \Score{85.3}{0.1}{TableHeat!49}{} & \Score{91.6}{0.2}{TableHeat!48}{} & \Score{76.8}{0.7}{TableHeat!45}{} & \Score{84.1}{0.5}{TableHeat!47}{} & \Score{91.5}{0.4}{TableHeat!49}{} & \Score{94.8}{0.2}{TableHeat!49}{} & \Score{77.7}{0.9}{TableHeat!45}{} & \Score{87.2}{0.4}{TableHeat!47}{} & \Score{96.3}{0.0}{TableHeat!33}{} & \Score{86.5}{2.0}{TableHeat!24}{} & \Score{94.1}{0.2}{TableHeat!46}{} & \Score{88.6}{0.1}{TableHeat!47}{} \\
 &  & \cellcolor{TableCarve}\BotStrut\textbf{Carve VFM} & \ScoreBest{97.3}{0.4}{TableHeat!50}{} & \ScoreBest{86.2}{0.4}{TableHeat!50}{} & \ScoreBest{93.8}{0.6}{TableHeat!50}{} & \ScoreBest{82.9}{0.7}{TableHeat!50}{} & \ScoreBest{87.9}{0.4}{TableHeat!50}{} & \ScoreBest{94.0}{0.1}{TableHeat!50}{} & \ScoreBest{95.7}{0.0}{TableHeat!50}{} & \ScoreBest{81.3}{0.5}{TableHeat!50}{} & \ScoreBest{89.9}{0.2}{TableHeat!50}{} & \ScoreBest{97.3}{0.0}{TableHeat!50}{} & \ScoreBest{89.2}{0.8}{TableHeat!50}{} & \ScoreBest{94.5}{0.4}{TableHeat!50}{} & \ScoreBest{90.8}{0.2}{TableHeat!50}{} \\
\arrayrulecolor{TableInk}\hhline{~---------------}\arrayrulecolor{white}
 &  & \TopStrut BlockDrop & \Score{32.7}{0.9}{TableHeat!0}{} & \Score{10.0}{0.4}{TableHeat!0}{} & \Score{17.9}{0.3}{TableHeat!5}{} & \Score{5.2}{0.2}{TableHeat!2}{} & \Score{2.4}{0.0}{TableHeat!0}{} & \Score{2.8}{0.2}{TableHeat!5}{} & \Score{5.3}{0.3}{TableHeat!0}{} & \Score{11.9}{0.3}{TableHeat!0}{} & \Score{13.5}{0.4}{TableHeat!0}{} & \Score{70.8}{0.1}{TableHeat!0}{} & \Score{80.6}{1.4}{TableHeat!7}{} & \Score{66.8}{0.4}{TableHeat!0}{} & \Score{26.7}{0.1}{TableHeat!0}{} \\
 &  & SNIP--Mag. & \Score{48.1}{1.3}{TableHeat!15}{} & \Score{18.8}{0.7}{TableHeat!9}{} & \Score{16.0}{2.2}{TableHeat!1}{} & \Score{4.6}{0.3}{TableHeat!1}{} & \Score{4.6}{0.2}{TableHeat!7}{} & \Score{2.2}{0.1}{TableHeat!2}{} & \Score{7.3}{0.5}{TableHeat!3}{} & \Score{17.9}{1.0}{TableHeat!8}{} & \Score{16.6}{0.5}{TableHeat!5}{} & \Score{76.6}{1.3}{TableHeat!12}{} & \Score{83.0}{0.6}{TableHeat!20}{} & \Score{78.0}{1.1}{TableHeat!22}{} & \Score{31.1}{0.8}{TableHeat!8}{} \\
 & \makebox[27pt][c]{85.7M} & SnapViT & \Score{46.1}{0.8}{TableHeat!13}{} & \Score{16.5}{0.7}{TableHeat!6}{} & \Score{15.2}{0.9}{TableHeat!0}{} & \Score{4.0}{0.2}{TableHeat!0}{} & \Score{3.2}{0.2}{TableHeat!3}{} & \Score{1.8}{0.1}{TableHeat!0}{} & \Score{6.0}{0.2}{TableHeat!1}{} & \Score{15.1}{1.0}{TableHeat!4}{} & \Score{13.6}{1.0}{TableHeat!0}{} & \Score{75.7}{0.7}{TableHeat!11}{} & \Score{79.4}{0.6}{TableHeat!0}{} & \Score{73.8}{0.9}{TableHeat!13}{} & \Score{29.2}{0.6}{TableHeat!5}{} \\
 & \makebox[27pt][c]{$\downarrow$} & FLAP & \Score{57.3}{0.7}{TableHeat!24}{} & \Score{29.8}{0.4}{TableHeat!20}{} & \Score{18.6}{0.9}{TableHeat!6}{} & \Score{5.4}{0.2}{TableHeat!2}{} & \Score{3.8}{0.4}{TableHeat!5}{} & \Score{3.1}{0.2}{TableHeat!6}{} & \Score{7.5}{0.7}{TableHeat!3}{} & \Score{18.2}{0.2}{TableHeat!8}{} & \Score{17.7}{0.8}{TableHeat!6}{} & \Score{79.9}{0.5}{TableHeat!20}{} & \Score{81.9}{1.3}{TableHeat!14}{} & \Score{88.8}{0.8}{TableHeat!42}{} & \Score{34.3}{0.2}{TableHeat!14}{} \\
 & \makebox[27pt][c]{\textbf{20M}} & SVD-LLM & \Score{84.3}{0.2}{TableHeat!50}{} & \Score{59.9}{0.5}{TableHeat!50}{} & \Score{42.7}{1.1}{TableHeat!46}{} & \Score{17.9}{1.0}{TableHeat!24}{} & \Score{13.3}{0.4}{TableHeat!37}{} & \Score{12.1}{0.7}{TableHeat!47}{} & \Score{26.6}{0.4}{TableHeat!32}{} & \Score{44.4}{0.8}{TableHeat!43}{} & \Score{38.7}{1.3}{TableHeat!39}{} & \ScoreBest{94.1}{0.1}{TableHeat!50}{} & \Score{85.0}{2.1}{TableHeat!32}{} & \Score{92.6}{0.3}{TableHeat!50}{} & \Score{51.0}{0.1}{TableHeat!44}{} \\
 &  & CORP & \Score{46.0}{5.6}{TableHeat!13}{} & \Score{22.1}{0.6}{TableHeat!12}{} & \Score{24.3}{1.0}{TableHeat!15}{} & \Score{6.9}{0.2}{TableHeat!5}{} & \Score{5.1}{1.2}{TableHeat!9}{} & \Score{3.4}{0.3}{TableHeat!7}{} & \Score{9.5}{1.3}{TableHeat!6}{} & \Score{21.2}{0.6}{TableHeat!12}{} & \Score{20.3}{1.8}{TableHeat!10}{} & \Score{86.2}{1.8}{TableHeat!33}{} & \Score{80.6}{2.3}{TableHeat!7}{} & \Score{83.9}{0.3}{TableHeat!33}{} & \Score{34.1}{0.7}{TableHeat!13}{} \\
 &  & SliceGPT & \ScoreBest{84.8}{0.4}{TableHeat!50}{} & \ScoreBest{60.2}{0.1}{TableHeat!50}{} & \Score{27.3}{0.4}{TableHeat!20}{} & \Score{6.1}{0.3}{TableHeat!4}{} & \Score{4.2}{0.3}{TableHeat!6}{} & \Score{3.4}{0.2}{TableHeat!8}{} & \Score{12.9}{0.4}{TableHeat!11}{} & \Score{22.9}{1.6}{TableHeat!14}{} & \Score{19.3}{0.7}{TableHeat!9}{} & \Score{92.3}{0.2}{TableHeat!46}{} & \ScoreBest{88.2}{0.5}{TableHeat!50}{} & \Score{92.2}{0.2}{TableHeat!49}{} & \Score{42.8}{0.2}{TableHeat!29}{} \\
 &  & OSSCAR & \Score{66.3}{1.3}{TableHeat!32}{} & \Score{32.8}{0.1}{TableHeat!23}{} & \Score{37.0}{0.4}{TableHeat!37}{} & \Score{21.6}{0.4}{TableHeat!31}{} & \Score{12.5}{0.7}{TableHeat!34}{} & \Score{6.4}{0.8}{TableHeat!21}{} & \Score{26.8}{0.4}{TableHeat!32}{} & \Score{46.2}{0.2}{TableHeat!45}{} & \Score{34.0}{0.8}{TableHeat!31}{} & \Score{90.1}{0.9}{TableHeat!41}{} & \Score{85.6}{1.4}{TableHeat!35}{} & \Score{87.3}{0.4}{TableHeat!39}{} & \Score{45.5}{0.3}{TableHeat!34}{} \\
\multirow{-19}{8pt}{\centering\rotatebox{90}{\textbf{DINOv3 ViT-B/16}}} &  & \cellcolor{TableCarve}\BotStrut\textbf{Carve VFM} & \Score{82.6}{1.6}{TableHeat!48}{} & \Score{59.0}{1.3}{TableHeat!49}{} & \ScoreBest{45.0}{1.9}{TableHeat!50}{} & \ScoreBest{32.4}{5.4}{TableHeat!50}{} & \ScoreBest{17.3}{7.5}{TableHeat!50}{} & \ScoreBest{12.7}{6.2}{TableHeat!50}{} & \ScoreBest{39.0}{6.2}{TableHeat!50}{} & \ScoreBest{50.1}{2.0}{TableHeat!50}{} & \ScoreBest{46.2}{3.7}{TableHeat!50}{} & \Score{93.1}{0.8}{TableHeat!48}{} & \Score{81.2}{2.7}{TableHeat!10}{} & \ScoreBest{92.8}{0.8}{TableHeat!50}{} & \ScoreBest{54.3}{1.1}{TableHeat!50}{} \\
\arrayrulecolor{TableInk}\Xhline{0.9pt}\arrayrulecolor{white}
 & \makebox[27pt][c]{85.8M} & \cellcolor{TableRef}\TopStrut\BotStrut Full & \Score{97.5}{0.1}{TableRef}{} & \Score{86.4}{0.0}{TableRef}{} & \Score{96.0}{0.3}{TableRef}{} & \Score{69.5}{0.2}{TableRef}{} & \Score{81.1}{0.2}{TableRef}{} & \Score{91.8}{0.2}{TableRef}{} & \Score{92.0}{0.3}{TableRef}{} & \Score{81.0}{0.4}{TableRef}{} & \Score{92.6}{0.1}{TableRef}{} & \Score{94.7}{0.3}{TableRef}{} & \Score{83.9}{1.0}{TableRef}{} & \Score{88.1}{0.5}{TableRef}{} & \Score{87.9}{0.0}{TableRef}{} \\
\arrayrulecolor{TableInk}\hhline{~---------------}\arrayrulecolor{white}
 &  & \TopStrut BlockDrop & \Score{81.0}{0.1}{TableHeat!31}{} & \Score{56.3}{0.4}{TableHeat!25}{} & \Score{45.8}{0.2}{TableHeat!13}{} & \Score{4.7}{1.0}{TableHeat!0}{} & \Score{7.4}{0.2}{TableHeat!1}{} & \Score{6.4}{0.0}{TableHeat!1}{} & \Score{17.1}{0.3}{TableHeat!3}{} & \Score{33.5}{0.4}{TableHeat!12}{} & \Score{39.8}{0.7}{TableHeat!15}{} & \Score{89.0}{0.5}{TableHeat!35}{} & \Score{82.9}{1.9}{TableHeat!5}{} & \Score{87.7}{0.2}{TableHeat!11}{} & \Score{46.0}{0.2}{TableHeat!9}{} \\
 &  & SNIP--Mag. & \Score{63.7}{1.0}{TableHeat!10}{} & \Score{36.2}{4.7}{TableHeat!7}{} & \Score{29.1}{4.6}{TableHeat!0}{} & \Score{8.9}{1.2}{TableHeat!3}{} & \Score{5.5}{0.4}{TableHeat!0}{} & \Score{4.3}{0.6}{TableHeat!0}{} & \Score{12.9}{3.9}{TableHeat!1}{} & \Score{18.8}{0.9}{TableHeat!0}{} & \Score{18.9}{0.4}{TableHeat!0}{} & \Score{73.9}{8.9}{TableHeat!0}{} & \Score{83.7}{1.5}{TableHeat!12}{} & \Score{86.7}{1.3}{TableHeat!0}{} & \Score{36.9}{2.5}{TableHeat!0}{} \\
 & \makebox[27pt][c]{85.8M} & SnapViT & \Score{55.1}{1.3}{TableHeat!0}{} & \Score{28.8}{0.6}{TableHeat!0}{} & \Score{28.7}{0.9}{TableHeat!0}{} & \Score{8.1}{0.4}{TableHeat!3}{} & \Score{6.4}{0.1}{TableHeat!1}{} & \Score{4.6}{0.0}{TableHeat!0}{} & \Score{12.0}{0.5}{TableHeat!0}{} & \Score{22.4}{0.2}{TableHeat!3}{} & \Score{25.0}{0.7}{TableHeat!4}{} & \Score{83.5}{1.9}{TableHeat!22}{} & \Score{82.3}{0.2}{TableHeat!0}{} & \Score{87.4}{0.7}{TableHeat!8}{} & \Score{37.0}{0.0}{TableHeat!0}{} \\
 & \makebox[27pt][c]{$\downarrow$} & FLAP & \Score{84.0}{0.6}{TableHeat!34}{} & \Score{59.7}{1.1}{TableHeat!28}{} & \Score{43.3}{1.0}{TableHeat!11}{} & \Score{24.6}{2.0}{TableHeat!17}{} & \Score{27.4}{3.4}{TableHeat!15}{} & \Score{42.7}{5.3}{TableHeat!23}{} & \Score{52.5}{0.9}{TableHeat!25}{} & \Score{52.7}{0.3}{TableHeat!29}{} & \Score{49.3}{0.6}{TableHeat!21}{} & \Score{92.4}{1.1}{TableHeat!43}{} & \Score{86.7}{1.7}{TableHeat!37}{} & \Score{88.2}{0.3}{TableHeat!17}{} & \Score{58.6}{0.6}{TableHeat!22}{} \\
 & \makebox[27pt][c]{\textbf{60M}} & SVD-LLM & \Score{94.8}{0.2}{TableHeat!47}{} & \Score{78.8}{0.4}{TableHeat!45}{} & \Score{82.2}{0.5}{TableHeat!41}{} & \Score{52.0}{0.3}{TableHeat!39}{} & \Score{65.3}{0.5}{TableHeat!41}{} & \Score{77.4}{0.0}{TableHeat!43}{} & \Score{85.3}{0.5}{TableHeat!46}{} & \Score{73.2}{0.3}{TableHeat!46}{} & \Score{83.8}{0.2}{TableHeat!46}{} & \Score{95.0}{0.1}{TableHeat!49}{} & \ScoreBest{88.2}{1.1}{TableHeat!50}{} & \ScoreBest{91.1}{0.1}{TableHeat!50}{} & \Score{80.6}{0.2}{TableHeat!44}{} \\
 &  & CORP & \Score{96.8}{0.3}{TableHeat!50}{} & \Score{83.2}{0.7}{TableHeat!49}{} & \Score{92.2}{0.4}{TableHeat!49}{} & \ScoreBest{65.0}{0.5}{TableHeat!50}{} & \Score{71.7}{0.2}{TableHeat!46}{} & \Score{84.9}{0.4}{TableHeat!47}{} & \Score{88.7}{0.3}{TableHeat!48}{} & \Score{60.1}{0.1}{TableHeat!35}{} & \Score{89.5}{0.2}{TableHeat!50}{} & \ScoreBest{95.4}{0.5}{TableHeat!50}{} & \Score{85.4}{3.1}{TableHeat!26}{} & \Score{90.0}{0.3}{TableHeat!38}{} & \Score{83.6}{0.4}{TableHeat!47}{} \\
 &  & SliceGPT & \Score{96.8}{0.1}{TableHeat!50}{} & \Score{82.0}{0.2}{TableHeat!48}{} & \Score{91.1}{0.8}{TableHeat!48}{} & \Score{59.9}{1.3}{TableHeat!46}{} & \Score{74.7}{0.7}{TableHeat!48}{} & \Score{88.9}{0.1}{TableHeat!50}{} & \Score{90.3}{0.5}{TableHeat!49}{} & \Score{77.4}{0.8}{TableHeat!49}{} & \Score{88.7}{0.4}{TableHeat!49}{} & \Score{93.8}{0.3}{TableHeat!46}{} & \Score{83.8}{3.2}{TableHeat!13}{} & \Score{88.1}{1.2}{TableHeat!15}{} & \Score{84.6}{0.2}{TableHeat!48}{} \\
 &  & OSSCAR & \Score{96.5}{0.3}{TableHeat!49}{} & \Score{82.0}{0.6}{TableHeat!48}{} & \Score{90.9}{0.5}{TableHeat!48}{} & \Score{62.4}{0.7}{TableHeat!48}{} & \Score{74.0}{0.1}{TableHeat!47}{} & \Score{87.3}{0.5}{TableHeat!49}{} & \Score{90.4}{0.5}{TableHeat!49}{} & \Score{76.2}{0.6}{TableHeat!48}{} & \Score{87.7}{0.3}{TableHeat!48}{} & \Score{93.6}{0.8}{TableHeat!46}{} & \Score{87.7}{2.2}{TableHeat!46}{} & \Score{89.2}{1.1}{TableHeat!28}{} & \Score{84.8}{0.3}{TableHeat!48}{} \\
 &  & \cellcolor{TableCarve}\BotStrut\textbf{Carve VFM} & \ScoreBest{97.2}{0.1}{TableHeat!50}{} & \ScoreBest{84.1}{0.4}{TableHeat!50}{} & \ScoreBest{93.2}{0.4}{TableHeat!50}{} & \Score{64.8}{1.2}{TableHeat!50}{} & \ScoreBest{77.7}{0.8}{TableHeat!50}{} & \ScoreBest{89.2}{0.1}{TableHeat!50}{} & \ScoreBest{91.7}{0.3}{TableHeat!50}{} & \ScoreBest{78.2}{0.9}{TableHeat!50}{} & \ScoreBest{90.1}{0.4}{TableHeat!50}{} & \ScoreBest{95.4}{0.0}{TableHeat!50}{} & \Score{88.1}{2.0}{TableHeat!49}{} & \Score{89.6}{0.6}{TableHeat!33}{} & \ScoreBest{86.6}{0.2}{TableHeat!50}{} \\
\arrayrulecolor{TableInk}\hhline{~---------------}\arrayrulecolor{white}
 &  & \TopStrut BlockDrop & \Score{54.0}{1.0}{TableHeat!4}{} & \Score{26.0}{0.2}{TableHeat!1}{} & \Score{20.8}{0.9}{TableHeat!0}{} & \Score{2.9}{0.4}{TableHeat!0}{} & \Score{1.7}{0.5}{TableHeat!0}{} & \Score{2.5}{0.1}{TableHeat!0}{} & \Score{6.8}{2.4}{TableHeat!0}{} & \Score{12.2}{7.6}{TableHeat!0}{} & \Score{18.1}{0.3}{TableHeat!0}{} & \Score{78.4}{0.4}{TableHeat!0}{} & \Score{79.1}{2.0}{TableHeat!0}{} & \Score{85.5}{1.2}{TableHeat!11}{} & \Score{32.3}{0.8}{TableHeat!0}{} \\
 &  & SNIP--Mag. & \Score{56.2}{0.5}{TableHeat!7}{} & \Score{26.1}{0.6}{TableHeat!1}{} & \Score{21.3}{1.0}{TableHeat!1}{} & \Score{6.2}{0.2}{TableHeat!10}{} & \Score{4.7}{0.2}{TableHeat!7}{} & \Score{3.8}{0.1}{TableHeat!3}{} & \Score{9.7}{1.0}{TableHeat!3}{} & \Score{20.5}{0.6}{TableHeat!13}{} & \Score{20.7}{1.7}{TableHeat!5}{} & \Score{83.7}{3.2}{TableHeat!21}{} & \Score{84.5}{1.2}{TableHeat!36}{} & \Score{86.2}{0.8}{TableHeat!16}{} & \Score{35.3}{0.9}{TableHeat!7}{} \\
 & \makebox[27pt][c]{85.8M} & SnapViT & \Score{51.4}{0.8}{TableHeat!0}{} & \Score{25.6}{0.6}{TableHeat!0}{} & \Score{22.3}{0.5}{TableHeat!3}{} & \Score{5.5}{0.2}{TableHeat!8}{} & \Score{5.2}{0.1}{TableHeat!8}{} & \Score{3.8}{0.2}{TableHeat!3}{} & \Score{9.6}{0.7}{TableHeat!3}{} & \Score{19.7}{0.4}{TableHeat!12}{} & \Score{20.7}{0.7}{TableHeat!5}{} & \Score{87.2}{1.3}{TableHeat!35}{} & \Score{84.1}{0.7}{TableHeat!33}{} & \Score{87.0}{0.6}{TableHeat!22}{} & \Score{35.2}{0.1}{TableHeat!7}{} \\
 & \makebox[27pt][c]{$\downarrow$} & FLAP & \Score{63.2}{0.9}{TableHeat!17}{} & \Score{35.7}{0.5}{TableHeat!17}{} & \Score{25.6}{1.2}{TableHeat!10}{} & \Score{7.1}{0.5}{TableHeat!13}{} & \Score{4.9}{0.1}{TableHeat!8}{} & \Score{4.7}{0.1}{TableHeat!4}{} & \Score{13.1}{0.0}{TableHeat!7}{} & \Score{28.9}{0.9}{TableHeat!27}{} & \Score{25.1}{1.2}{TableHeat!14}{} & \Score{89.9}{1.1}{TableHeat!45}{} & \ScoreBest{86.6}{0.1}{TableHeat!50}{} & \Score{90.1}{0.4}{TableHeat!45}{} & \Score{39.6}{0.1}{TableHeat!17}{} \\
 & \makebox[27pt][c]{\textbf{20M}} & SVD-LLM & \Score{77.7}{0.1}{TableHeat!38}{} & \Score{50.9}{0.6}{TableHeat!42}{} & \Score{31.9}{0.8}{TableHeat!24}{} & \Score{9.5}{0.5}{TableHeat!20}{} & \Score{6.6}{0.5}{TableHeat!12}{} & \Score{4.3}{0.5}{TableHeat!4}{} & \Score{12.7}{0.1}{TableHeat!6}{} & \Score{33.1}{0.8}{TableHeat!33}{} & \Score{28.6}{0.9}{TableHeat!20}{} & \Score{90.1}{0.6}{TableHeat!46}{} & \Score{85.4}{2.5}{TableHeat!42}{} & \Score{90.7}{0.9}{TableHeat!49}{} & \Score{43.5}{0.3}{TableHeat!26}{} \\
 &  & CORP & \Score{62.2}{0.7}{TableHeat!16}{} & \Score{34.1}{0.4}{TableHeat!14}{} & \Score{25.4}{0.3}{TableHeat!10}{} & \Score{6.3}{0.1}{TableHeat!10}{} & \Score{5.4}{0.3}{TableHeat!9}{} & \Score{3.0}{0.2}{TableHeat!1}{} & \Score{10.5}{1.0}{TableHeat!4}{} & \Score{20.2}{0.8}{TableHeat!13}{} & \Score{21.9}{1.1}{TableHeat!7}{} & \Score{90.2}{0.2}{TableHeat!46}{} & \Score{84.5}{0.8}{TableHeat!36}{} & \Score{89.9}{0.9}{TableHeat!43}{} & \Score{37.8}{0.1}{TableHeat!13}{} \\
 &  & SliceGPT & \ScoreBest{85.8}{0.3}{TableHeat!50}{} & \ScoreBest{56.0}{1.0}{TableHeat!50}{} & \ScoreBest{43.7}{0.3}{TableHeat!50}{} & \ScoreBest{19.4}{1.0}{TableHeat!50}{} & \ScoreBest{22.8}{0.3}{TableHeat!50}{} & \ScoreBest{27.4}{0.4}{TableHeat!50}{} & \ScoreBest{53.1}{1.1}{TableHeat!50}{} & \ScoreBest{43.7}{1.4}{TableHeat!50}{} & \Score{35.2}{1.3}{TableHeat!33}{} & \Score{88.2}{0.8}{TableHeat!39}{} & \Score{82.1}{2.6}{TableHeat!20}{} & \Score{84.0}{1.0}{TableHeat!0}{} & \ScoreBest{53.4}{0.4}{TableHeat!50}{} \\
 &  & OSSCAR & \Score{66.9}{0.5}{TableHeat!23}{} & \Score{38.5}{0.7}{TableHeat!21}{} & \Score{28.2}{0.7}{TableHeat!16}{} & \Score{10.7}{1.7}{TableHeat!24}{} & \Score{5.3}{0.4}{TableHeat!9}{} & \Score{5.2}{0.2}{TableHeat!5}{} & \Score{20.3}{1.6}{TableHeat!15}{} & \Score{40.1}{0.6}{TableHeat!44}{} & \Score{29.2}{1.1}{TableHeat!21}{} & \Score{83.8}{2.3}{TableHeat!21}{} & \Score{81.0}{3.3}{TableHeat!13}{} & \Score{86.6}{0.6}{TableHeat!19}{} & \Score{41.3}{0.4}{TableHeat!21}{} \\
\multirow{-19}{8pt}{\centering\rotatebox{90}{\textbf{MetaCLIP-2 ViT-B/16}}} &  & \cellcolor{TableCarve}\BotStrut\textbf{Carve VFM} & \Score{76.6}{1.3}{TableHeat!37}{} & \Score{42.5}{5.1}{TableHeat!28}{} & \Score{36.6}{1.1}{TableHeat!34}{} & \Score{16.5}{0.9}{TableHeat!41}{} & \Score{15.9}{1.0}{TableHeat!34}{} & \Score{7.8}{1.9}{TableHeat!11}{} & \Score{34.5}{1.4}{TableHeat!30}{} & \Score{42.3}{4.4}{TableHeat!48}{} & \ScoreBest{43.9}{1.2}{TableHeat!50}{} & \ScoreBest{91.1}{0.3}{TableHeat!50}{} & \Score{85.1}{2.4}{TableHeat!40}{} & \ScoreBest{90.8}{2.2}{TableHeat!50}{} & \Score{48.6}{0.5}{TableHeat!39}{} \\
\arrayrulecolor{TableInk}\Xhline{0.9pt}\arrayrulecolor{white}
 & \makebox[27pt][c]{303.1M} & \cellcolor{TableRef}\TopStrut\BotStrut Full & \Score{99.1}{0.0}{TableRef}{} & \Score{93.1}{0.1}{TableRef}{} & \Score{97.3}{0.1}{TableRef}{} & \Score{80.4}{0.1}{TableRef}{} & \Score{90.6}{0.1}{TableRef}{} & \Score{94.4}{0.1}{TableRef}{} & \Score{96.7}{0.1}{TableRef}{} & \Score{82.7}{0.2}{TableRef}{} & \Score{90.7}{0.1}{TableRef}{} & \Score{95.9}{0.0}{TableRef}{} & \Score{87.3}{0.6}{TableRef}{} & \Score{92.3}{0.1}{TableRef}{} & \Score{91.7}{0.0}{TableRef}{} \\
\arrayrulecolor{TableInk}\hhline{~---------------}\arrayrulecolor{white}
 &  & \TopStrut BlockDrop & \Score{77.8}{0.5}{TableHeat!14}{} & \Score{56.4}{0.2}{TableHeat!13}{} & \Score{76.5}{0.7}{TableHeat!28}{} & \Score{41.8}{5.1}{TableHeat!14}{} & \Score{46.7}{2.7}{TableHeat!21}{} & \Score{58.6}{0.5}{TableHeat!27}{} & \Score{65.9}{0.2}{TableHeat!28}{} & \Score{64.8}{0.1}{TableHeat!28}{} & \Score{70.1}{0.9}{TableHeat!26}{} & \Score{95.6}{0.2}{TableHeat!32}{} & \Score{86.1}{0.8}{TableHeat!10}{} & \Score{90.8}{0.4}{TableHeat!0}{} & \Score{69.3}{0.2}{TableHeat!23}{} \\
 &  & SNIP--Mag. & \Score{69.6}{0.7}{TableHeat!0}{} & \Score{44.5}{1.3}{TableHeat!0}{} & \Score{50.3}{3.1}{TableHeat!0}{} & \Score{25.8}{1.0}{TableHeat!0}{} & \Score{15.3}{1.2}{TableHeat!0}{} & \Score{17.2}{1.3}{TableHeat!0}{} & \Score{26.7}{4.6}{TableHeat!0}{} & \Score{40.4}{1.6}{TableHeat!0}{} & \Score{47.9}{3.0}{TableHeat!0}{} & \Score{94.6}{0.3}{TableHeat!19}{} & \Score{86.4}{0.7}{TableHeat!13}{} & \Score{92.1}{0.8}{TableHeat!23}{} & \Score{50.9}{0.6}{TableHeat!0}{} \\
 & \makebox[27pt][c]{303.1M} & SnapViT & \Score{72.2}{1.0}{TableHeat!4}{} & \Score{46.5}{0.3}{TableHeat!2}{} & \Score{61.8}{0.8}{TableHeat!12}{} & \Score{26.8}{0.5}{TableHeat!1}{} & \Score{15.9}{0.4}{TableHeat!0}{} & \Score{18.6}{0.1}{TableHeat!1}{} & \Score{28.0}{0.5}{TableHeat!1}{} & \Score{45.8}{0.1}{TableHeat!6}{} & \Score{55.5}{1.0}{TableHeat!9}{} & \Score{93.1}{0.1}{TableHeat!0}{} & \Score{85.2}{0.9}{TableHeat!0}{} & \Score{92.9}{0.3}{TableHeat!38}{} & \Score{53.5}{0.1}{TableHeat!3}{} \\
 & \makebox[27pt][c]{$\downarrow$} & FLAP & \Score{93.4}{0.3}{TableHeat!41}{} & \Score{77.5}{0.2}{TableHeat!36}{} & \Score{88.7}{0.1}{TableHeat!42}{} & \Score{73.7}{0.7}{TableHeat!43}{} & \Score{81.2}{1.0}{TableHeat!44}{} & \Score{90.7}{0.5}{TableHeat!48}{} & \Score{92.6}{0.1}{TableHeat!47}{} & \Score{78.5}{0.5}{TableHeat!44}{} & \Score{88.7}{0.3}{TableHeat!47}{} & \Score{95.7}{0.4}{TableHeat!33}{} & \Score{86.2}{2.0}{TableHeat!11}{} & \Score{91.6}{0.1}{TableHeat!14}{} & \Score{86.6}{0.2}{TableHeat!44}{} \\
 & \makebox[27pt][c]{\textbf{200M}} & SVD-LLM & \ScoreBest{98.9}{0.0}{TableHeat!50}{} & \Score{90.7}{0.1}{TableHeat!50}{} & \Score{96.1}{0.1}{TableHeat!50}{} & \Score{80.6}{1.2}{TableHeat!49}{} & \Score{89.7}{0.1}{TableHeat!50}{} & \ScoreBest{94.1}{0.1}{TableHeat!50}{} & \ScoreBest{96.5}{0.0}{TableHeat!50}{} & \Score{83.2}{0.3}{TableHeat!50}{} & \ScoreBest{91.0}{0.5}{TableHeat!50}{} & \Score{96.2}{0.4}{TableHeat!40}{} & \ScoreBest{89.9}{1.9}{TableHeat!50}{} & \Score{93.1}{0.2}{TableHeat!41}{} & \ScoreBest{91.7}{0.3}{TableHeat!50}{} \\
 &  & CORP & \ScoreBest{98.9}{0.0}{TableHeat!50}{} & \Score{89.5}{2.8}{TableHeat!48}{} & \ScoreBest{96.5}{0.1}{TableHeat!50}{} & \Score{79.8}{2.4}{TableHeat!48}{} & \Score{83.1}{0.1}{TableHeat!46}{} & \Score{88.7}{0.2}{TableHeat!46}{} & \Score{95.8}{0.2}{TableHeat!49}{} & \Score{63.8}{0.1}{TableHeat!27}{} & \Score{90.6}{0.2}{TableHeat!50}{} & \ScoreBest{97.0}{0.3}{TableHeat!50}{} & \Score{88.2}{1.1}{TableHeat!32}{} & \Score{93.0}{0.4}{TableHeat!39}{} & \Score{88.8}{0.2}{TableHeat!46}{} \\
 &  & SliceGPT & \Score{98.7}{0.1}{TableHeat!50}{} & \Score{89.2}{0.3}{TableHeat!48}{} & \Score{94.0}{0.3}{TableHeat!47}{} & \Score{78.5}{0.5}{TableHeat!47}{} & \Score{87.4}{0.2}{TableHeat!48}{} & \Score{93.2}{0.3}{TableHeat!49}{} & \Score{95.5}{0.0}{TableHeat!49}{} & \Score{80.3}{0.1}{TableHeat!47}{} & \Score{89.5}{0.2}{TableHeat!48}{} & \Score{94.4}{0.4}{TableHeat!17}{} & \Score{89.2}{0.9}{TableHeat!43}{} & \Score{92.1}{0.6}{TableHeat!23}{} & \Score{90.2}{0.1}{TableHeat!48}{} \\
 &  & OSSCAR & \Score{97.8}{0.2}{TableHeat!48}{} & \Score{86.6}{0.2}{TableHeat!45}{} & \Score{92.1}{0.4}{TableHeat!45}{} & \Score{71.0}{0.7}{TableHeat!40}{} & \Score{74.2}{3.9}{TableHeat!40}{} & \Score{82.0}{1.0}{TableHeat!42}{} & \Score{92.4}{0.8}{TableHeat!47}{} & \Score{75.0}{0.5}{TableHeat!40}{} & \Score{89.0}{0.4}{TableHeat!48}{} & \Score{95.3}{0.5}{TableHeat!29}{} & \Score{86.6}{0.1}{TableHeat!15}{} & \Score{92.2}{0.2}{TableHeat!26}{} & \Score{86.2}{0.5}{TableHeat!43}{} \\
 &  & \cellcolor{TableCarve}\BotStrut\textbf{Carve VFM} & \ScoreBest{98.9}{0.1}{TableHeat!50}{} & \ScoreBest{90.9}{0.3}{TableHeat!50}{} & \Score{96.3}{0.2}{TableHeat!50}{} & \ScoreBest{81.6}{0.6}{TableHeat!50}{} & \ScoreBest{89.8}{0.7}{TableHeat!50}{} & \Score{94.0}{0.3}{TableHeat!50}{} & \ScoreBest{96.5}{0.1}{TableHeat!50}{} & \ScoreBest{83.3}{0.9}{TableHeat!50}{} & \Score{90.8}{0.4}{TableHeat!50}{} & \Score{96.5}{0.7}{TableHeat!44}{} & \Score{87.9}{1.7}{TableHeat!29}{} & \ScoreBest{93.6}{0.2}{TableHeat!50}{} & \ScoreBest{91.7}{0.5}{TableHeat!50}{} \\
\arrayrulecolor{TableInk}\hhline{~---------------}\arrayrulecolor{white}
 &  & \TopStrut BlockDrop & \Score{54.6}{4.7}{TableHeat!7}{} & \Score{32.8}{0.3}{TableHeat!11}{} & \Score{25.6}{0.7}{TableHeat!6}{} & \Score{8.4}{1.0}{TableHeat!3}{} & \Score{7.6}{0.9}{TableHeat!3}{} & \Score{5.4}{0.2}{TableHeat!2}{} & \Score{14.6}{1.5}{TableHeat!4}{} & \Score{25.1}{0.1}{TableHeat!7}{} & \Score{15.6}{1.6}{TableHeat!0}{} & \Score{90.8}{0.3}{TableHeat!42}{} & \Score{84.7}{1.5}{TableHeat!38}{} & \Score{85.0}{0.1}{TableHeat!13}{} & \Score{37.5}{0.3}{TableHeat!6}{} \\
 &  & SNIP--Mag. & \Score{49.2}{1.3}{TableHeat!2}{} & \Score{22.7}{1.1}{TableHeat!3}{} & \Score{20.5}{1.3}{TableHeat!1}{} & \Score{5.6}{0.2}{TableHeat!1}{} & \Score{6.4}{0.4}{TableHeat!2}{} & \Score{3.5}{0.2}{TableHeat!1}{} & \Score{10.0}{0.2}{TableHeat!2}{} & \Score{18.4}{0.6}{TableHeat!2}{} & \Score{23.1}{1.7}{TableHeat!6}{} & \Score{76.5}{5.4}{TableHeat!15}{} & \Score{81.0}{1.0}{TableHeat!5}{} & \Score{82.2}{0.3}{TableHeat!0}{} & \Score{33.3}{0.3}{TableHeat!2}{} \\
 & \makebox[27pt][c]{303.1M} & SnapViT & \Score{47.5}{2.0}{TableHeat!0}{} & \Score{19.4}{0.9}{TableHeat!0}{} & \Score{18.7}{0.6}{TableHeat!0}{} & \Score{4.4}{0.2}{TableHeat!0}{} & \Score{3.7}{0.3}{TableHeat!0}{} & \Score{2.6}{0.4}{TableHeat!0}{} & \Score{7.3}{0.7}{TableHeat!0}{} & \Score{16.5}{0.4}{TableHeat!0}{} & \Score{17.2}{1.3}{TableHeat!1}{} & \Score{68.8}{0.5}{TableHeat!0}{} & \Score{81.7}{1.1}{TableHeat!11}{} & \Score{82.4}{1.0}{TableHeat!1}{} & \Score{30.8}{0.3}{TableHeat!0}{} \\
 & \makebox[27pt][c]{$\downarrow$} & FLAP & \Score{77.4}{0.2}{TableHeat!31}{} & \Score{52.4}{0.3}{TableHeat!27}{} & \Score{45.4}{0.6}{TableHeat!22}{} & \Score{29.3}{0.6}{TableHeat!18}{} & \Score{11.4}{0.4}{TableHeat!5}{} & \Score{12.5}{0.8}{TableHeat!6}{} & \Score{28.9}{0.5}{TableHeat!13}{} & \Score{51.5}{0.7}{TableHeat!30}{} & \Score{40.8}{0.7}{TableHeat!19}{} & \Score{92.9}{0.3}{TableHeat!46}{} & \Score{84.0}{0.6}{TableHeat!32}{} & \Score{92.8}{0.1}{TableHeat!49}{} & \Score{51.6}{0.1}{TableHeat!19}{} \\
 & \makebox[27pt][c]{\textbf{100M}} & SVD-LLM & \ScoreBest{96.1}{0.0}{TableHeat!50}{} & \ScoreBest{80.0}{0.2}{TableHeat!50}{} & \Score{78.5}{0.5}{TableHeat!49}{} & \ScoreBest{74.0}{0.6}{TableHeat!50}{} & \ScoreBest{81.5}{0.2}{TableHeat!50}{} & \ScoreBest{91.8}{0.1}{TableHeat!50}{} & \ScoreBest{92.0}{0.2}{TableHeat!50}{} & \Score{69.8}{1.0}{TableHeat!46}{} & \ScoreBest{82.1}{0.3}{TableHeat!50}{} & \ScoreBest{94.9}{0.3}{TableHeat!50}{} & \ScoreBest{86.1}{2.2}{TableHeat!50}{} & \Score{92.0}{0.4}{TableHeat!45}{} & \ScoreBest{84.9}{0.2}{TableHeat!50}{} \\
 &  & CORP & \Score{71.0}{8.0}{TableHeat!24}{} & \Score{37.1}{15.6}{TableHeat!15}{} & \Score{57.9}{3.5}{TableHeat!32}{} & \Score{36.9}{10.0}{TableHeat!23}{} & \Score{33.5}{2.4}{TableHeat!19}{} & \Score{59.6}{2.7}{TableHeat!32}{} & \Score{58.1}{3.6}{TableHeat!30}{} & \Score{45.7}{1.1}{TableHeat!25}{} & \Score{57.1}{1.4}{TableHeat!31}{} & \Score{91.8}{0.6}{TableHeat!44}{} & \Score{80.4}{3.5}{TableHeat!0}{} & \Score{89.5}{3.4}{TableHeat!33}{} & \Score{59.9}{0.8}{TableHeat!27}{} \\
 &  & SliceGPT & \Score{95.8}{0.2}{TableHeat!50}{} & \Score{77.5}{0.1}{TableHeat!48}{} & \Score{67.1}{0.8}{TableHeat!39}{} & \Score{68.2}{1.0}{TableHeat!46}{} & \Score{58.1}{4.8}{TableHeat!35}{} & \Score{86.4}{0.5}{TableHeat!47}{} & \Score{86.4}{0.2}{TableHeat!47}{} & \Score{47.0}{0.4}{TableHeat!26}{} & \Score{78.7}{1.3}{TableHeat!47}{} & \Score{92.1}{0.4}{TableHeat!45}{} & \Score{84.1}{0.7}{TableHeat!32}{} & \Score{91.3}{0.7}{TableHeat!42}{} & \Score{77.7}{0.3}{TableHeat!43}{} \\
 &  & OSSCAR & \Score{84.9}{0.5}{TableHeat!38}{} & \Score{62.0}{0.1}{TableHeat!35}{} & \Score{61.3}{0.8}{TableHeat!35}{} & \Score{47.2}{1.3}{TableHeat!31}{} & \Score{27.7}{2.8}{TableHeat!15}{} & \Score{56.5}{3.8}{TableHeat!30}{} & \Score{69.9}{4.7}{TableHeat!37}{} & \Score{63.4}{0.5}{TableHeat!41}{} & \Score{56.2}{2.4}{TableHeat!31}{} & \Score{92.4}{0.1}{TableHeat!45}{} & \Score{84.4}{1.4}{TableHeat!35}{} & \Score{91.5}{0.5}{TableHeat!43}{} & \Score{66.4}{0.5}{TableHeat!33}{} \\
\multirow{-19}{8pt}{\centering\rotatebox{90}{\textbf{DINOv3 ViT-L/16}}} &  & \cellcolor{TableCarve}\BotStrut\textbf{Carve VFM} & \Score{94.2}{0.9}{TableHeat!48}{} & \Score{76.9}{4.2}{TableHeat!47}{} & \ScoreBest{80.1}{1.3}{TableHeat!50}{} & \Score{73.6}{1.6}{TableHeat!50}{} & \Score{76.4}{5.4}{TableHeat!47}{} & \Score{89.0}{1.8}{TableHeat!48}{} & \Score{90.6}{0.8}{TableHeat!49}{} & \ScoreBest{74.0}{1.0}{TableHeat!50}{} & \Score{80.3}{1.9}{TableHeat!49}{} & \Score{92.7}{2.9}{TableHeat!46}{} & \Score{84.2}{1.2}{TableHeat!33}{} & \ScoreBest{93.1}{0.4}{TableHeat!50}{} & \Score{83.7}{1.9}{TableHeat!49}{} \\
\arrayrulecolor{TableInk}\Xhline{0.9pt}\arrayrulecolor{white}
\end{tabularx}
\end{adjustbox}
\end{center}
\end{table*}
% Protocol and color definitions are documented in COMPLETED_BASELINES_README.md.
% New methods: vision adaptations with validation selection; historical rows use test selection.
% Bold marks the highest displayed compressed score, including ties at one decimal.
% This descriptive ranking spans different protocols; it is not a matched-protocol comparison.

\endgroup
\arrayrulecolor{black}

\paragraph{Recoverability becomes critical under aggressive compression.}
Among dense structured-pruning methods with complete results in Table~\ref{tab:completed-baselines}, Carve VFM has the highest reported twelve-task mean in seven of eight source--budget settings. The exception is MetaCLIP-2 at 20M parameters, where SliceGPT reaches 53.4\% versus 48.6\% for Carve VFM. Checkpoint-selection protocols for SliceGPT and OSSCAR differ from those of the other baselines, so these rankings are descriptive. The supplement reports the complete twelve-setting grid for the other baselines. As budgets tighten, useful structures may also be removed; performance then depends on preserving or reconstructing their task-relevant contribution. These comparisons motivate considering allocation and recoverability jointly, but do not isolate their individual contributions.

\paragraph{The benefit transfers across representations and domains.}
The advantage is consistent across both self-supervised DINOv3 and image--text-pretrained MetaCLIP-2, and from ViT-S through ViT-L, suggesting that recoverability is largely independent of the pretrained representation. The gains are largest on fine-grained and texture-rich recognition tasks, where small semantic differences require preserving highly task-specific capacity. The gains are smaller and more variable on the medical benchmarks, where the source models transfer less consistently.

\textbf{Carve VFM vs.\ factorization.} SVD-LLM offers a complementary approach through low-rank factorization. Across the twelve source--budget settings in the supplement, Carve VFM has the higher reported twelve-task mean in nine, SVD-LLM in two, and the two tie in one at the reported one-decimal precision. SVD-LLM leads for DINOv3 ViT-B at 35M and ViT-L at 100M; the tie occurs for ViT-L at 200M. Carve VFM returns conventional dense projections, whereas SVD-LLM executes low-rank factors, motivating the deployment comparison below.

\paragraph{Deployment efficiency.}

\begin{figure}[!t]
\centering
\begin{minipage}[t]{0.50\linewidth}
\vspace{0pt}
\centering
    \includegraphics[width=0.90\linewidth]{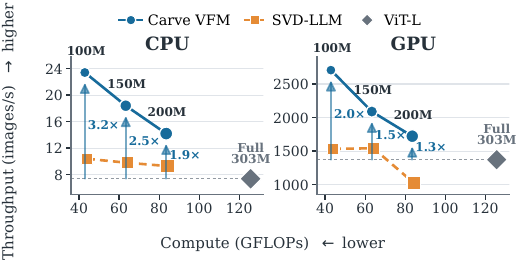}
    \caption{ViT-L efficiency of Carve VFM, uncompressed ViT-L,
    and SVD-LLM on CPU (AMD EPYC 9754) and GPU (RTX PRO 6000
    Blackwell). Marker area denotes peak memory.}
    \label{fig:vitl-deployment}
\par\medskip
\centering
    \includegraphics[width=0.90\linewidth]{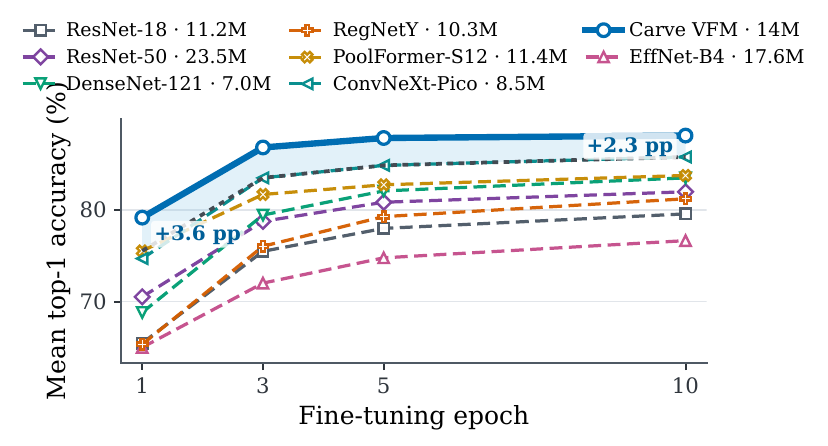}
    \caption{Average full-fine-tuning accuracy over twelve tasks. Model sizes in the legend; shading marks the gap between Carve VFM and the strongest baseline at each checkpoint.}
    \label{fig:full-convergence}
\end{minipage}\hfill
\begin{minipage}[t]{0.46\linewidth}
\vspace{0pt}
\centering
    \includegraphics[width=0.95\linewidth]{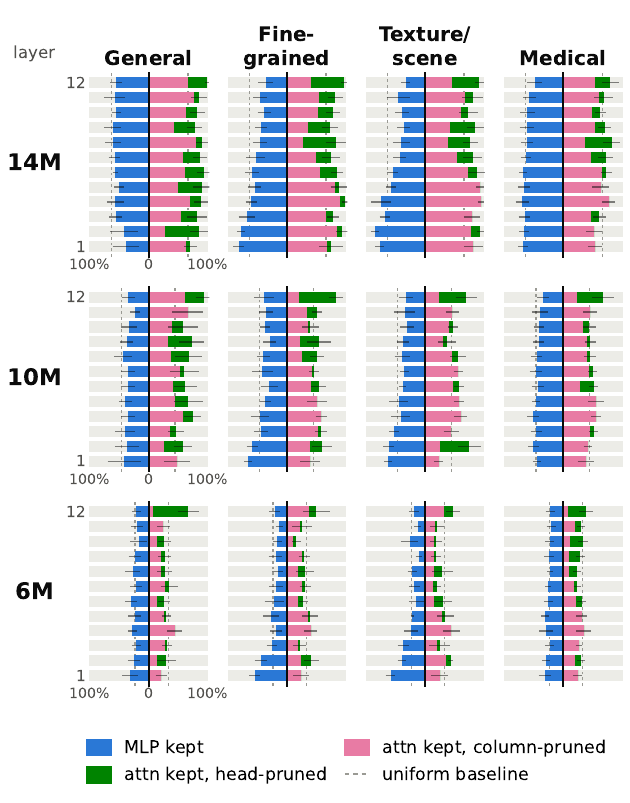}
    \caption{Architectural patterns discovered by Carve VFM across task categories. Rows are encoder layers; bars show retained MLP and attention, separated into head and column pruning. Dashed lines are uniform allocation baseline, and whiskers show variation across tasks and seeds.}
    \label{fig:arch-fingerprint}
\end{minipage}
\end{figure}

Figure~\ref{fig:vitl-deployment} compares \emph{Carve VFM} with the original ViT-L and the parameter-matched low-rank baseline SVD-LLM. Other structured pruning methods already produce conventional dense networks with the same execution pattern as \emph{Carve VFM} and therefore do not provide a distinct deployment comparison.

\emph{Carve VFM} consistently shifts the deployment trade-off toward higher throughput, lower compute, and lower memory. Compared with the original ViT-L, it improves CPU throughput by 1.9--3.2$\times$ and GPU throughput by 1.3--2.0$\times$, while reducing peak GPU memory by 26--51\%. At matched parameter budgets, it also delivers 1.5--2.3$\times$ higher CPU throughput and 1.4--1.8$\times$ higher GPU throughput than SVD-LLM, while using 7--17\% less GPU memory. This advantage stems from the compression mechanism itself: SVD-LLM replaces each projection with two serial low-rank operations, whereas \emph{Carve VFM} physically narrows the dense operators, reducing both arithmetic and memory traffic without introducing additional layers.

\paragraph{Full fine-tuning and convergence.}

We fully fine-tune the carved 14M ViT-S specialist for each task and compare it with natively pretrained compact architectures under a ten-epoch schedule. Table~\ref{tab:full-ft-native} reports a best-checkpoint macro-average of 88.1\% for the 14M specialist and 85.8\% for ConvNeXt-Pico, the strongest evaluated native compact model, a difference of 2.3 percentage points. These models have different parameter counts: 14M and 8.5M, respectively. Figure~\ref{fig:full-convergence} compares the reported learning curves. Full optimization settings, per-dataset results, and comparisons with all eight compact baselines are provided in the supplement.

\paragraph{Architecture patterns discovered by Carve VFM.}
Figure~\ref{fig:arch-fingerprint} shows that Carve VFM consistently discovers
task-dependent subnetworks rather than uniformly distributing the capacity. Texture and scene recognition retain more early-layer MLP capacity, with fine-grained tasks exhibiting the same tendency more moderately, indicating that these domains benefit from preserving richer low-level features. In contrast, medical specialists remain much closer to a uniform allocation, suggesting that task-relevant information is distributed more evenly across depth. A second pattern is remarkably consistent across domains: most parameter savings come from reducing MLP capacity, while attention is compressed more conservatively. In particular, whole-head removal is rare in the early layers, where Carve VFM instead favors column pruning, distributing the reduction across all heads rather than eliminating complete ones. These results reveal that Carve VFM discovers distinct compression regimes across depth.

\paragraph{When is global fitness needed?}
The adaptive audit provides a complementary diagnostic. For DINOv3 ViT-S/B, the search switched from local recoverability to cross-validated task accuracy in 26\% of mild-budget runs and 51\% of tight-budget runs; this rose to 79\% for MetaCLIP-2 and 89\% for ViT-L. Cached local statistics therefore guide
exploration efficiently when their ranking is stable, while global task
evidence becomes more important as compression, representation shift,
or model scale makes that ranking unreliable.

\paragraph{Decoupling search from repair.}
To determine whether Carve’s search improves performance beyond repair alone, we compare uniform allocations and FLAP subnetworks with and without the same repair against full Carve VFM. We evaluate DINOv3-S at 14M and 6M parameters on CIFAR-100, CUB-200, and BloodMNIST. At these budgets, repair raises mean accuracy from 31.7/31.5\% to 78.2/47.7\% for uniform allocation and from 28.2/26.6\% to 63.5/36.7\% for FLAP subnetworks. Across these datasets, the corresponding Carve entries in Table~\ref{tab:completed-baselines} average 85.4/54.3\%, 7.2/6.6 percentage points above repaired uniform allocation. Thus, repair provides most of the gain, while searching allocations and attention operators adds further improvements on average.

\section{Conclusion}

We introduced \emph{Carve VFM}, a framework for extracting compact,
task-specific models from pretrained VFMs under an explicit parameter
budget, without backbone retraining. A single search
jointly decides how capacity is allocated across layers, which
structures to remove, and how to recover their lost contribution, producing standard dense models competitive with evaluated compression methods and
compact architectures, with gains varying by source, task, and budget.
Our evaluation centers on image classification. Natural next steps
include dense prediction and multimodal foundation models, dynamic
deployment budgets that adapt to shifting device constraints, and
continual carving that re-allocates capacity as new tasks arrive.
Relaxing the requirement for labeled calibration data through
self-supervised surrogates would further \mbox{broaden applicability}.

\FloatBarrier
\label{migration:main-text-end}

\clearpage
\subsection*{AI use statement}
\label{sec:ai-use}
Generative AI tools were used only to assist with manuscript drafting and
grammatical and language editing. They were not used to develop the method,
generate data, implement the research methods, execute experiments, or analyze
experimental results. The authors take responsibility for the final manuscript
and all reported claims and results.

\subsection*{Ethics statement}
Carve VFM is evaluated on existing image-classification benchmarks, including
three medical-image benchmarks. The experiments assess classification accuracy
and computational efficiency. These benchmark results do not establish clinical
safety or effectiveness; compressed models would require application-specific
validation before clinical use. The reported evaluation does not measure
demographic fairness or privacy, so accuracy and efficiency gains should not be
interpreted as evidence of improvements in those properties.

\subsection*{Reproducibility statement}
Section~\ref{sec:method} describes the architecture search, recoverability
scoring, and closed-form repair procedure.
Appendix~\ref{sec:supp-protocol} reports the source checkpoints, datasets,
data splits, calibration sampling, and shared evaluation protocol.
Appendix~\ref{sec:supp-method} provides the search and repair settings,
random seeds, software versions, and computing hardware.
Appendix~\ref{sec:supp-full-results} reports per-dataset results with
standard deviations across three seeds. The full-fine-tuning and deployment
measurement protocols are provided in
Appendices~\ref{sec:supp-full-finetuning} and~\ref{sec:supp-deployment},
respectively.

\clearpage
\bibliographystyle{assets/plainnat}
\bibliography{resources/main}

\clearpage
\beginappendix
\suppressfloats[t]
\label{sec:supp-overview}

These appendices report results and implementation details omitted from the
main paper for space: complete three-budget, per-dataset comparisons for every
source model; the full fine-tuning comparison; documented search and repair
settings; measured wall-clock and deployment costs; and limitations and future
work. Unless stated otherwise, accuracy is reported as
mean$\pm$sample standard deviation over seeds 17, 42, and 91.

\section{Evaluation Protocol}
\label{sec:supp-protocol}

\subsection{Source Models, Tasks, and Budgets}

We evaluate DINOv3 ViT-S/16 (21.6M parameters), ViT-B/16 (85.7M), and
ViT-L/16 (303.1M), and MetaCLIP-2 Worldwide ViT-B/16 (85.8M). The DINOv3
models use the LVD-1689M pretrained checkpoints; MetaCLIP-2 uses its released
Worldwide B/16 checkpoint. Table~\ref{tab:supp-source-configs} lists the
three dense parameter budgets evaluated for each source.

\begin{table}[!ht]
\caption{Source models, dense parameter budgets, and final repair rank.}
\label{tab:supp-source-configs}
\centering
\normalsize
\setlength{\tabcolsep}{3.6pt}
\begin{tabular}{@{}lrrrc@{}}
\toprule
Source & \multicolumn{3}{c}{Budgets (M)} & Repair rank \\
\cmidrule(lr){2-4}
 & Mild & Intermediate & Tight & $r$ \\
\midrule
DINOv3 ViT-S/16 & 14 & 10 & 6 & 128 \\
DINOv3 ViT-B/16 & 60 & 35 & 20 & 128 \\
MetaCLIP-2 ViT-B/16 & 60 & 35 & 20 & 256 \\
DINOv3 ViT-L/16 & 200 & 150 & 100 & 256 \\
\bottomrule
\end{tabular}

\end{table}

The twelve tasks are CIFAR-10, CIFAR-100, and Caltech-256 (general
recognition); FGVC-Aircraft, CUB-200, Stanford Cars, and Oxford-IIIT Pets
(fine-grained recognition); DTD and MIT Indoor-67 (texture and scene
recognition); and BloodMNIST, PneumoniaMNIST, and OrganAMNIST (medical
recognition). We use the standard train/test split for every dataset except
Caltech-256, for which each seed defines an 80/20 split. All inputs are
resized to $224\times224$ using the preprocessing associated with the source
checkpoint.

\subsection{Calibration and Model Selection}

For each task and seed, 512 class-balanced examples are drawn from the
training split to form the compression calibration set. All matched methods
share the same examples and data split. Carve VFM further partitions this set
internally when held-out task accuracy is required; no test image is used to
guide search or choose an architecture.

After compression, the backbone is frozen and a new linear classifier is
trained on the full training split for ten epochs. AdamW uses batch size 32,
learning rate $10^{-3}$, and weight decay $10^{-4}$. We evaluate epochs 1, 3,
5, and 10. SliceGPT and OSSCAR use a different checkpoint-selection protocol
from the protocol-matched baseline grid; their downstream training budget
remains ten epochs, and their per-dataset results are reported separately in
Table~\ref{tab:supp-additional-baselines}.

\subsection{Baselines}

The common comparison contains seven structured methods. Random removes
budget-matched structures uniformly at random. BlockDrop~\cite{wu2018blockdrop}
removes complete transformer blocks. SNIP--Magnitude~\cite{lee2019snip} ranks
structures using sensitivity at initialization followed by magnitude
aggregation. SnapViT~\cite{simoncini2025elastic} applies its structured ViT
scoring rule. FLAP~\cite{an2024flap} uses activation fluctuation to allocate
structured removals. SVD-LLM~\cite{wang2025svd} replaces projections with
budget-matched low-rank factors. CORP~\cite{zhang2026corp} uses compensated
one-shot structured pruning. All methods start from the same source checkpoint
and use identical calibration data, seeds, dense parameter budgets, and
frozen-backbone linear-probing protocol.

Where a complete protocol-matched grid exists, the source-specific tables also
include scoring baselines omitted from the main paper: Magnitude~\cite{li2017pruning},
LAMP~\cite{lee2021lamp}, Wanda~\cite{sun2024wanda}, and Taylor~\cite{molchanov2019importance} for DINOv3
ViT-S/B, and Magnitude and LAMP for MetaCLIP-2 ViT-B. We omit UPop and EA-ViT
because no completed three-seed, twelve-task grid is available under the
current checkpoint, budget, and linear-probing protocol. This rule avoids mixing
results obtained under different training regimes with the controlled
comparison.

\section{Complete Per-Dataset Results}
\label{sec:supp-full-results}

The main paper displays only the mild and tight budgets and a compact
baseline set. Tables~\ref{tab:supp-dinov3-vits}--\ref{tab:supp-dinov3-vitl}
restore the intermediate budget and every completed protocol-matched
baseline. ``Mean'' gives equal weight to the twelve dataset means rather than
to their numbers of examples. Values are mean$\pm$sample standard deviation
over seeds 17, 42, and 91. Accuracies and standard deviations are displayed to
one decimal, while macro means are computed from the unrounded per-task values.
Each source table includes only methods with a complete protocol-matched grid;
bold marks the best result within each budget.

Table~\ref{tab:supp-additional-baselines} reports the separate per-dataset
results for SliceGPT and OSSCAR, including the standard deviations omitted
from the main table.

\begin{landscape}
\begin{table}[p]
\caption{Complete per-dataset frozen-backbone linear-probing accuracy (\%) for DINOv3 ViT-S/16 at three dense parameter budgets.}
\label{tab:supp-dinov3-vits}
\centering
\footnotesize
\setlength{\tabcolsep}{2.2pt}
\renewcommand{\arraystretch}{0.74}
\begin{adjustbox}{width=\linewidth,max totalheight=.82\textheight,keepaspectratio}
\begin{tabular}{@{}ll*{12}{r}@{\hspace{2.5pt}}r@{}}
\toprule
Budget & Method
 & \multicolumn{3}{c}{General}
 & \multicolumn{4}{c}{Fine-grained}
 & \multicolumn{2}{c}{Texture/scene}
 & \multicolumn{3}{c}{Medical} & \textbf{Mean} \\
\cmidrule(lr){3-5}\cmidrule(lr){6-9}\cmidrule(lr){10-11}
\cmidrule(lr){12-14}\cmidrule(l){15-15}
 & & C10 & C100 & Cal. & Air. & CUB & Cars & Pets & DTD & MIT
 & Blood & Pneu. & OrganA & \\
\midrule
21.6M & Linear probe & 96.0\std{0.1} & 84.3\std{0.2} & 92.6\std{0.2} & 70.3\std{0.1} & 83.6\std{0.1} & 88.8\std{0.3} & 94.0\std{0.4} & 77.3\std{0.1} & 87.4\std{0.2} & 95.0\std{0.1} & 84.8\std{0.2} & 92.1\std{0.1} & 87.2\std{0.0} \\
\cmidrule(lr){1-15}
\multirow{12}{*}{\textbf{14M}} & Random & 57.6\std{1.9} & 30.9\std{1.9} & 29.1\std{4.3} & 7.7\std{1.2} & 6.0\std{1.1} & 3.9\std{0.9} & 13.1\std{2.4} & 25.1\std{3.1} & 26.0\std{2.6} & 84.8\std{1.3} & 83.4\std{1.5} & 87.9\std{1.5} & 38.0\std{2.0} \\
 & Magnitude & 42.6\std{0.1} & 17.3\std{0.6} & 14.4\std{1.0} & 3.9\std{0.2} & 2.5\std{0.1} & 1.9\std{0.1} & 7.3\std{0.3} & 13.3\std{0.3} & 12.2\std{0.5} & 79.3\std{0.2} & 82.5\std{0.8} & 78.2\std{0.2} & 29.6\std{0.4} \\
 & LAMP & 44.4\std{0.2} & 19.4\std{0.2} & 19.7\std{1.0} & 5.5\std{0.0} & 4.1\std{0.2} & 2.8\std{0.1} & 9.6\std{0.4} & 17.2\std{0.1} & 16.5\std{0.3} & 85.5\std{0.2} & 82.3\std{1.3} & 83.5\std{0.2} & 32.5\std{0.4} \\
 & SNIP--Mag. & 46.4\std{0.8} & 14.9\std{0.5} & 9.8\std{0.9} & 4.1\std{0.7} & 2.2\std{0.5} & 1.8\std{0.3} & 5.5\std{0.8} & 15.5\std{1.7} & 8.6\std{1.5} & 85.1\std{0.7} & 81.6\std{0.3} & 82.6\std{1.9} & 29.8\std{0.9} \\
 & Wanda & 70.6\std{0.6} & 47.1\std{1.9} & 52.2\std{2.4} & 19.1\std{0.8} & 16.6\std{0.2} & 11.1\std{1.1} & 31.0\std{1.1} & 43.7\std{0.1} & 42.4\std{0.9} & 92.4\std{0.1} & 85.1\std{1.7} & 93.3\std{0.1} & 50.4\std{0.9} \\
 & Taylor & 59.5\std{13.2} & 39.3\std{0.2} & 40.1\std{2.1} & 7.6\std{7.1} & 5.3\std{4.5} & 5.6\std{4.4} & 18.9\std{4.5} & 32.2\std{6.6} & 16.9\std{8.9} & 79.6\std{16.8} & 83.7\std{2.6} & 88.9\std{3.4} & 39.8\std{6.2} \\
 & SnapViT & 52.5\std{0.3} & 26.3\std{0.4} & 31.3\std{1.1} & 9.2\std{0.2} & 6.2\std{0.1} & 3.9\std{0.2} & 13.9\std{0.4} & 28.0\std{0.5} & 27.2\std{0.6} & 86.7\std{0.4} & 82.4\std{0.5} & 87.9\std{0.1} & 38.0\std{0.4} \\
 & BlockDrop & 65.6\std{0.1} & 40.6\std{0.5} & 44.3\std{1.1} & 14.5\std{0.3} & 12.1\std{2.7} & 11.1\std{0.2} & 21.2\std{0.5} & 42.7\std{0.3} & 40.2\std{3.1} & 91.6\std{0.1} & 81.6\std{0.9} & 89.3\std{0.2} & 46.2\std{0.3} \\
 & FLAP & 71.1\std{0.1} & 52.1\std{1.2} & 57.6\std{0.6} & 33.2\std{3.2} & 25.2\std{0.3} & 39.7\std{1.2} & 53.2\std{2.3} & 54.9\std{0.8} & 51.7\std{1.7} & 94.3\std{0.2} & 83.9\std{1.8} & 92.7\std{0.2} & 59.1\std{0.1} \\
 & SVD-LLM & 92.5\std{0.1} & 75.0\std{0.2} & 76.4\std{1.1} & 51.2\std{0.9} & 61.6\std{1.0} & 63.0\std{4.3} & 79.4\std{0.8} & 65.1\std{0.8} & 74.7\std{1.6} & 95.7\std{0.1} & 88.9\std{1.7} & 92.7\std{0.2} & 76.3\std{0.4} \\
 & CORP & 94.7\std{0.1} & \textbf{81.1\std{0.2}} & 85.5\std{0.4} & \textbf{70.9\std{0.3}} & 72.9\std{0.5} & 83.0\std{0.5} & 90.3\std{0.0} & 57.9\std{0.2} & 81.9\std{0.3} & \textbf{96.8\std{0.1}} & 88.0\std{1.0} & \textbf{93.9\std{0.2}} & 83.1\std{0.0} \\
 & \textbf{Carve VFM} & \textbf{94.8\std{0.5}} & 80.5\std{0.5} & \textbf{86.5\std{0.9}} & 70.1\std{0.4} & \textbf{79.1\std{0.9}} & \textbf{88.7\std{0.5}} & \textbf{91.3\std{0.3}} & \textbf{75.2\std{0.3}} & \textbf{83.9\std{0.4}} & 96.7\std{0.2} & \textbf{89.2\std{1.6}} & 93.6\std{0.0} & \textbf{85.8\std{0.1}} \\
\cmidrule(lr){1-15}
\multirow{12}{*}{\textbf{10M}} & Random & 41.6\std{1.3} & 17.0\std{1.8} & 13.7\std{2.9} & 3.7\std{0.6} & 2.8\std{0.5} & 1.8\std{0.5} & 6.8\std{1.2} & 14.0\std{1.1} & 11.0\std{2.9} & 71.2\std{10.4} & 78.9\std{2.6} & 77.8\std{1.4} & 28.4\std{2.3} \\
 & Magnitude & 38.9\std{0.4} & 14.3\std{0.5} & 12.7\std{0.8} & 3.7\std{0.3} & 2.8\std{0.1} & 1.7\std{0.1} & 6.4\std{0.3} & 12.8\std{0.8} & 11.5\std{0.8} & 75.2\std{0.2} & 82.7\std{0.6} & 70.1\std{0.6} & 27.7\std{0.5} \\
 & LAMP & 41.8\std{0.2} & 16.5\std{0.5} & 17.4\std{1.0} & 4.6\std{0.2} & 3.3\std{0.2} & 2.5\std{0.3} & 8.2\std{0.1} & 16.1\std{0.5} & 14.3\std{0.4} & 81.9\std{0.3} & 83.2\std{0.4} & 79.9\std{0.6} & 30.8\std{0.4} \\
 & SNIP--Mag. & 43.5\std{1.3} & 14.8\std{1.9} & 11.6\std{0.4} & 3.7\std{0.9} & 1.7\std{0.2} & 1.1\std{0.4} & 5.2\std{0.4} & 14.4\std{0.2} & 10.0\std{2.5} & 80.6\std{1.0} & 81.5\std{1.5} & 80.3\std{1.4} & 29.0\std{1.0} \\
 & Wanda & 62.2\std{0.2} & 35.1\std{0.9} & 30.5\std{0.8} & 10.3\std{0.2} & 7.8\std{0.4} & 5.0\std{0.1} & 14.7\std{0.5} & 26.2\std{0.3} & 29.6\std{0.7} & 86.5\std{0.5} & 83.1\std{0.9} & 87.5\std{0.4} & 39.9\std{0.5} \\
 & Taylor & 44.3\std{11.3} & 25.5\std{3.6} & 22.5\std{7.4} & 5.1\std{1.8} & 3.6\std{0.5} & 2.7\std{0.3} & 9.7\std{3.4} & 20.4\std{7.7} & 14.9\std{7.2} & 77.6\std{2.2} & 79.3\std{5.9} & 83.4\std{8.4} & 32.4\std{5.0} \\
 & SnapViT & 42.1\std{0.3} & 16.5\std{0.4} & 21.4\std{0.7} & 6.3\std{0.4} & 4.3\std{0.1} & 2.9\std{0.1} & 10.2\std{0.4} & 22.4\std{0.3} & 19.9\std{0.5} & 77.2\std{0.5} & 81.5\std{0.8} & 82.6\std{0.2} & 32.3\std{0.4} \\
 & BlockDrop & 57.2\std{0.1} & 29.5\std{2.1} & 29.9\std{0.5} & 10.5\std{0.2} & 8.2\std{0.2} & 5.8\std{0.1} & 17.1\std{0.3} & 25.6\std{0.4} & 26.5\std{0.6} & 87.0\std{0.2} & 78.3\std{1.0} & 87.1\std{0.1} & 38.6\std{0.1} \\
 & FLAP & 65.1\std{0.3} & 39.4\std{0.1} & 32.5\std{0.8} & 15.4\std{0.2} & 9.7\std{0.2} & 6.5\std{0.1} & 25.8\std{0.7} & 40.5\std{0.9} & 31.7\std{0.8} & 89.7\std{0.4} & 80.4\std{0.6} & 91.7\std{0.2} & 44.0\std{0.1} \\
 & SVD-LLM & 87.7\std{0.0} & 66.2\std{0.2} & 32.0\std{0.5} & 17.9\std{0.6} & 7.9\std{0.3} & 4.6\std{0.3} & 41.6\std{3.4} & 46.2\std{1.2} & 36.3\std{0.5} & 95.0\std{0.3} & 86.8\std{2.4} & 93.2\std{0.1} & 51.3\std{0.4} \\
 & CORP & 87.1\std{0.8} & 66.4\std{0.7} & 54.7\std{0.9} & 48.1\std{0.4} & 37.5\std{5.5} & 33.1\std{1.6} & 51.4\std{3.1} & 41.9\std{0.3} & 59.3\std{0.5} & 94.5\std{0.3} & 85.1\std{1.1} & 92.8\std{0.4} & 62.6\std{0.4} \\
 & \textbf{Carve VFM} & \textbf{91.1\std{1.0}} & \textbf{73.6\std{1.4}} & \textbf{70.7\std{1.0}} & \textbf{52.5\std{2.7}} & \textbf{63.3\std{1.9}} & \textbf{74.6\std{0.9}} & \textbf{80.3\std{0.7}} & \textbf{67.0\std{1.8}} & \textbf{70.4\std{3.9}} & \textbf{96.2\std{0.2}} & \textbf{87.4\std{1.1}} & \textbf{93.5\std{0.5}} & \textbf{76.7\std{0.8}} \\
\cmidrule(lr){1-15}
\multirow{12}{*}{\textbf{6M}} & Random & 38.6\std{1.8} & 14.0\std{0.9} & 12.3\std{0.9} & 3.0\std{0.2} & 2.4\std{0.1} & 1.7\std{0.1} & 6.4\std{0.3} & 11.9\std{0.8} & 10.0\std{0.9} & 66.3\std{8.7} & 78.0\std{2.0} & 74.1\std{3.2} & 26.6\std{1.7} \\
 & Magnitude & 40.3\std{0.3} & 14.5\std{0.3} & 14.0\std{0.1} & 4.3\std{0.2} & 2.7\std{0.1} & 2.0\std{0.1} & 6.7\std{0.5} & 14.3\std{0.0} & 12.4\std{1.0} & 72.2\std{0.5} & 79.8\std{1.1} & 75.9\std{0.8} & 28.3\std{0.4} \\
 & LAMP & 41.9\std{0.4} & 17.5\std{0.3} & 17.1\std{0.9} & 5.2\std{0.1} & 3.8\std{0.1} & 2.3\std{0.1} & 7.5\std{0.3} & 16.0\std{0.2} & 14.3\std{0.7} & 81.5\std{0.2} & 81.1\std{2.0} & 80.5\std{0.3} & 30.7\std{0.5} \\
 & SNIP--Mag. & 42.0\std{2.5} & 14.2\std{0.2} & 9.1\std{0.8} & 2.4\std{0.3} & 1.3\std{0.0} & 1.0\std{0.1} & 4.4\std{0.6} & 15.1\std{3.2} & 10.2\std{3.8} & 76.1\std{2.5} & 81.4\std{1.8} & 74.2\std{2.3} & 27.6\std{1.5} \\
 & Wanda & 55.1\std{0.2} & 27.6\std{0.5} & 18.6\std{0.7} & 4.8\std{0.2} & 4.8\std{0.2} & 3.1\std{0.2} & 9.3\std{0.4} & 17.2\std{0.3} & 18.2\std{0.9} & 79.5\std{0.9} & 81.4\std{0.8} & 83.0\std{0.5} & 33.5\std{0.5} \\
 & Taylor & 42.0\std{8.7} & 18.1\std{5.8} & 17.4\std{2.5} & 3.7\std{0.7} & 3.0\std{1.1} & 1.7\std{0.5} & 7.8\std{1.2} & 12.7\std{4.4} & 16.5\std{0.3} & 67.1\std{5.1} & 78.7\std{5.0} & 80.9\std{3.7} & 29.1\std{3.2} \\
 & SnapViT & 44.0\std{0.9} & 18.6\std{0.4} & 16.5\std{0.7} & 4.5\std{0.4} & 3.8\std{0.2} & 2.2\std{0.1} & 7.1\std{0.4} & 17.7\std{0.1} & 15.8\std{0.8} & 72.4\std{0.4} & 82.3\std{1.0} & 80.4\std{0.3} & 30.4\std{0.5} \\
 & BlockDrop & 51.9\std{0.4} & 25.7\std{0.2} & 19.8\std{0.5} & 7.5\std{0.2} & 5.4\std{0.2} & 3.8\std{0.2} & 12.3\std{0.1} & 19.1\std{0.0} & 19.3\std{2.5} & 69.9\std{0.7} & \textbf{85.4\std{0.7}} & 85.2\std{0.2} & 33.8\std{0.1} \\
 & FLAP & 53.8\std{0.7} & 26.7\std{0.2} & 16.9\std{0.5} & 4.8\std{0.8} & 4.6\std{0.1} & 2.2\std{0.2} & 7.7\std{1.2} & 23.2\std{0.5} & 15.3\std{0.4} & 73.8\std{0.9} & 80.0\std{0.6} & 83.1\std{0.7} & 32.7\std{0.0} \\
 & SVD-LLM & 75.1\std{0.4} & 48.0\std{0.3} & 19.3\std{0.3} & 4.4\std{0.2} & 4.6\std{0.1} & 2.5\std{0.0} & 10.7\std{0.6} & 28.0\std{1.2} & 21.1\std{0.3} & 90.1\std{0.5} & 83.5\std{2.1} & 89.4\std{0.2} & 39.7\std{0.3} \\
 & CORP & 57.1\std{0.5} & 29.7\std{0.4} & 20.4\std{0.6} & 5.8\std{0.4} & 5.8\std{0.3} & 2.7\std{0.1} & 10.1\std{0.2} & 16.0\std{0.7} & 14.6\std{1.2} & 79.7\std{1.0} & 82.1\std{2.8} & 86.5\std{0.9} & 34.2\std{0.2} \\
 & \textbf{Carve VFM} & \textbf{77.1\std{2.0}} & \textbf{51.3\std{6.0}} & \textbf{38.4\std{1.2}} & \textbf{21.4\std{2.9}} & \textbf{18.5\std{4.8}} & \textbf{10.2\std{1.7}} & \textbf{31.1\std{7.1}} & \textbf{43.8\std{1.7}} & \textbf{34.8\std{1.8}} & \textbf{93.2\std{0.9}} & 83.9\std{0.5} & \textbf{90.4\std{1.0}} & \textbf{49.5\std{0.7}} \\
\bottomrule
\end{tabular}
\end{adjustbox}

\end{table}
\end{landscape}

\begin{landscape}
\begin{table}[p]
\caption{Complete per-dataset frozen-backbone linear-probing accuracy (\%) for DINOv3 ViT-B/16 at three dense parameter budgets.}
\label{tab:supp-dinov3-vitb}
\centering
\footnotesize
\setlength{\tabcolsep}{2.2pt}
\renewcommand{\arraystretch}{0.74}
\begin{adjustbox}{width=\linewidth,max totalheight=.82\textheight,keepaspectratio}
\begin{tabular}{@{}ll*{12}{r}@{\hspace{2.5pt}}r@{}}
\toprule
Budget & Method
 & \multicolumn{3}{c}{General}
 & \multicolumn{4}{c}{Fine-grained}
 & \multicolumn{2}{c}{Texture/scene}
 & \multicolumn{3}{c}{Medical} & \textbf{Mean} \\
\cmidrule(lr){3-5}\cmidrule(lr){6-9}\cmidrule(lr){10-11}
\cmidrule(lr){12-14}\cmidrule(l){15-15}
 & & C10 & C100 & Cal. & Air. & CUB & Cars & Pets & DTD & MIT
 & Blood & Pneu. & OrganA & \\
\midrule
85.7M & Linear probe & 98.2\std{0.1} & 88.9\std{0.1} & 95.9\std{0.3} & 81.8\std{0.2} & 89.2\std{0.1} & 94.3\std{0.2} & 95.9\std{0.2} & 80.0\std{0.2} & 89.7\std{0.5} & 95.7\std{0.1} & 83.8\std{0.2} & 93.7\std{0.1} & 90.6\std{0.0} \\
\cmidrule(lr){1-15}
\multirow{12}{*}{\textbf{60M}} & Random & 65.9\std{6.0} & 38.7\std{6.1} & 52.6\std{9.5} & 17.8\std{3.8} & 13.0\std{3.5} & 12.7\std{5.2} & 28.3\std{8.7} & 45.2\std{4.7} & 47.9\std{8.0} & 93.0\std{1.6} & 86.4\std{1.8} & 89.7\std{1.3} & 49.3\std{5.0} \\
 & Magnitude & 54.4\std{0.8} & 26.9\std{0.4} & 35.6\std{1.0} & 9.6\std{0.3} & 5.6\std{0.1} & 5.7\std{0.2} & 14.4\std{0.2} & 35.7\std{0.4} & 38.3\std{0.6} & 89.1\std{0.9} & 85.8\std{0.7} & 87.7\std{0.2} & 40.7\std{0.5} \\
 & LAMP & 56.1\std{0.3} & 28.6\std{0.3} & 39.1\std{0.5} & 12.9\std{0.1} & 7.4\std{0.3} & 7.3\std{0.2} & 15.3\std{0.6} & 37.4\std{0.2} & 40.4\std{1.0} & 85.2\std{0.8} & 84.3\std{1.3} & 88.0\std{0.3} & 41.8\std{0.5} \\
 & SNIP--Mag. & 74.3\std{1.9} & 46.3\std{1.3} & 47.9\std{2.6} & 22.9\std{1.4} & 14.5\std{1.3} & 10.3\std{0.4} & 22.6\std{4.6} & 43.8\std{2.0} & 44.5\std{4.4} & 94.4\std{0.9} & 88.1\std{0.9} & 92.9\std{0.3} & 50.2\std{1.8} \\
 & Wanda & 74.5\std{0.9} & 51.2\std{1.6} & 54.9\std{0.5} & 22.8\std{1.0} & 14.9\std{0.2} & 12.3\std{0.8} & 27.9\std{0.1} & 49.8\std{0.6} & 50.1\std{0.9} & 94.0\std{0.2} & 87.2\std{1.5} & 93.3\std{0.8} & 52.7\std{0.8} \\
 & Taylor & 78.1\std{4.6} & 49.1\std{10.1} & 60.6\std{6.6} & 36.0\std{15.1} & 29.3\std{10.6} & 38.3\std{8.9} & 33.4\std{6.2} & 51.9\std{2.0} & 52.1\std{4.6} & 95.4\std{0.1} & 87.4\std{2.3} & 92.4\std{0.3} & 58.7\std{5.9} \\
 & SnapViT & 75.7\std{0.4} & 52.1\std{0.3} & 60.1\std{0.8} & 20.8\std{0.2} & 13.5\std{0.3} & 15.7\std{0.2} & 33.5\std{0.1} & 45.5\std{0.6} & 45.2\std{1.3} & 95.2\std{0.3} & 87.0\std{0.7} & 92.0\std{0.3} & 53.0\std{0.4} \\
 & BlockDrop & 73.7\std{0.4} & 48.5\std{0.5} & 55.1\std{0.3} & 26.8\std{0.4} & 12.6\std{0.2} & 15.9\std{0.3} & 23.3\std{0.1} & 55.4\std{5.1} & 54.7\std{0.3} & 94.8\std{0.3} & 84.0\std{1.5} & 89.8\std{0.2} & 52.9\std{0.4} \\
 & FLAP & 83.9\std{0.4} & 65.5\std{0.0} & 71.4\std{0.4} & 66.5\std{0.7} & 59.0\std{2.5} & 82.0\std{0.7} & 74.4\std{0.2} & 71.4\std{1.0} & 74.7\std{1.2} & 96.8\std{0.4} & 84.6\std{3.8} & 94.0\std{0.3} & 77.0\std{0.3} \\
 & SVD-LLM & 97.0\std{0.2} & 84.9\std{0.2} & 93.2\std{0.1} & 80.5\std{0.3} & 86.5\std{0.2} & 93.7\std{0.2} & 95.3\std{0.2} & 79.1\std{0.3} & 88.5\std{0.4} & 97.1\std{0.2} & 87.2\std{2.7} & 93.7\std{0.1} & 89.7\std{0.2} \\
 & CORP & 96.4\std{1.2} & 85.3\std{0.3} & 93.7\std{0.3} & 81.7\std{0.5} & 80.7\std{0.2} & 88.2\std{0.1} & 94.6\std{0.2} & 59.0\std{4.8} & 89.0\std{0.3} & \textbf{97.3\std{0.3}} & 87.0\std{3.2} & 94.2\std{0.2} & 87.3\std{0.7} \\
 & \textbf{Carve VFM} & \textbf{97.3\std{0.4}} & \textbf{86.2\std{0.4}} & \textbf{93.8\std{0.6}} & \textbf{82.9\std{0.7}} & \textbf{87.9\std{0.4}} & \textbf{94.0\std{0.1}} & \textbf{95.7\std{0.0}} & \textbf{81.3\std{0.5}} & \textbf{89.9\std{0.2}} & \textbf{97.3\std{0.0}} & \textbf{89.2\std{0.8}} & \textbf{94.5\std{0.4}} & \textbf{90.8\std{0.2}} \\
\cmidrule(lr){1-15}
\multirow{12}{*}{\textbf{35M}} & Random & 49.7\std{1.3} & 21.1\std{0.7} & 22.7\std{1.5} & 5.9\std{0.3} & 4.8\std{0.2} & 3.1\std{0.1} & 8.7\std{0.5} & 21.1\std{1.4} & 20.7\std{3.0} & 80.1\std{2.4} & 82.7\std{1.2} & 82.5\std{1.3} & 33.6\std{1.1} \\
 & Magnitude & 42.8\std{0.8} & 17.9\std{0.4} & 22.1\std{0.7} & 7.1\std{0.0} & 4.8\std{0.1} & 3.5\std{0.2} & 9.0\std{0.3} & 19.3\std{0.5} & 22.0\std{0.7} & 85.4\std{0.6} & 84.7\std{0.5} & 80.4\std{0.7} & 33.3\std{0.5} \\
 & LAMP & 45.1\std{1.1} & 17.6\std{0.6} & 22.1\std{0.8} & 5.7\std{0.1} & 4.6\std{0.4} & 3.4\std{0.2} & 8.7\std{0.3} & 19.3\std{0.5} & 22.5\std{0.2} & 79.7\std{0.7} & 81.9\std{1.0} & 81.2\std{1.5} & 32.7\std{0.6} \\
 & SNIP--Mag. & 58.2\std{1.3} & 28.0\std{1.7} & 25.8\std{2.5} & 7.8\std{0.1} & 7.2\std{0.1} & 4.1\std{0.1} & 11.4\std{1.0} & 27.5\std{1.5} & 26.2\std{0.6} & 87.4\std{1.5} & 84.5\std{1.3} & 86.0\std{0.8} & 37.8\std{1.0} \\
 & Wanda & 58.4\std{0.3} & 29.2\std{1.1} & 22.8\std{0.9} & 6.6\std{0.4} & 5.1\std{0.3} & 2.9\std{0.2} & 9.2\std{0.6} & 21.3\std{1.0} & 21.5\std{0.2} & 83.9\std{1.2} & 83.1\std{0.5} & 86.6\std{1.1} & 35.9\std{0.6} \\
 & Taylor & 54.5\std{1.1} & 25.9\std{2.2} & 25.1\std{2.1} & 7.6\std{1.6} & 6.0\std{0.2} & 3.4\std{0.3} & 9.8\std{1.2} & 26.0\std{0.4} & 25.2\std{1.2} & 85.4\std{1.6} & 82.4\std{0.3} & 85.6\std{1.2} & 36.4\std{1.1} \\
 & SnapViT & 56.5\std{0.2} & 27.5\std{1.0} & 26.6\std{0.9} & 7.0\std{0.1} & 5.6\std{0.3} & 3.3\std{0.1} & 9.6\std{0.4} & 24.7\std{0.5} & 23.1\std{0.7} & 85.4\std{0.1} & 81.7\std{1.5} & 83.2\std{0.7} & 36.2\std{0.5} \\
 & BlockDrop & 33.9\std{0.8} & 12.4\std{0.2} & 19.7\std{0.4} & 6.5\std{0.4} & 1.7\std{0.2} & 4.1\std{0.2} & 5.3\std{0.7} & 30.4\std{0.3} & 29.8\std{0.6} & 61.5\std{1.1} & 82.6\std{0.9} & 69.3\std{0.6} & 29.8\std{0.1} \\
 & FLAP & 71.9\std{0.2} & 47.3\std{1.1} & 31.6\std{1.2} & 15.9\std{0.7} & 7.3\std{0.1} & 6.4\std{0.2} & 15.2\std{0.9} & 37.0\std{2.4} & 27.5\std{0.5} & 93.8\std{0.1} & 84.4\std{3.0} & 93.2\std{0.4} & 44.3\std{0.3} \\
 & SVD-LLM & \textbf{93.9\std{0.1}} & \textbf{76.0\std{0.3}} & \textbf{73.6\std{1.3}} & 67.9\std{0.4} & \textbf{72.7\std{0.2}} & \textbf{85.9\std{0.4}} & \textbf{87.8\std{0.2}} & 68.7\std{0.8} & 76.6\std{0.2} & 96.1\std{0.6} & \textbf{86.7\std{2.7}} & 93.6\std{0.3} & \textbf{81.6\std{0.2}} \\
 & CORP & 80.2\std{10.8} & 73.0\std{0.9} & 67.9\std{0.6} & 66.7\std{0.2} & 44.0\std{17.0} & 79.6\std{0.3} & 51.1\std{24.2} & 48.9\std{0.3} & 64.7\std{3.8} & \textbf{96.2\std{0.3}} & 85.7\std{2.3} & \textbf{94.7\std{0.2}} & 71.1\std{0.4} \\
 & \textbf{Carve VFM} & 93.3\std{1.0} & 71.4\std{4.9} & 69.4\std{3.5} & \textbf{69.0\std{1.0}} & 69.8\std{2.1} & 85.3\std{4.5} & 83.3\std{4.3} & \textbf{69.9\std{2.7}} & \textbf{77.6\std{2.3}} & 95.8\std{0.9} & 85.7\std{1.3} & 94.4\std{0.3} & 80.4\std{1.4} \\
\cmidrule(lr){1-15}
\multirow{12}{*}{\textbf{20M}} & Random & 41.7\std{1.3} & 14.6\std{1.0} & 15.0\std{2.1} & 3.7\std{0.5} & 3.4\std{0.7} & 2.0\std{0.4} & 6.0\std{0.7} & 14.2\std{1.6} & 14.2\std{1.3} & 69.1\std{2.9} & 80.5\std{1.2} & 75.5\std{0.7} & 28.3\std{1.2} \\
 & Magnitude & 41.2\std{1.1} & 15.5\std{0.3} & 17.6\std{1.0} & 4.6\std{0.1} & 3.6\std{0.5} & 2.4\std{0.1} & 7.2\std{0.3} & 15.7\std{0.9} & 15.5\std{0.6} & 77.9\std{0.8} & 82.0\std{1.1} & 73.2\std{0.3} & 29.7\std{0.6} \\
 & LAMP & 40.4\std{1.2} & 13.6\std{0.7} & 15.3\std{1.0} & 4.0\std{0.2} & 3.3\std{0.2} & 2.1\std{0.2} & 6.6\std{0.3} & 14.8\std{0.7} & 14.8\std{0.6} & 66.7\std{1.2} & 80.3\std{0.5} & 76.0\std{1.4} & 28.2\std{0.7} \\
 & SNIP--Mag. & 48.1\std{1.3} & 18.8\std{0.7} & 16.0\std{2.2} & 4.6\std{0.3} & 4.6\std{0.2} & 2.2\std{0.1} & 7.3\std{0.5} & 17.9\std{1.0} & 16.6\std{0.5} & 76.6\std{1.3} & 83.0\std{0.6} & 78.0\std{1.1} & 31.1\std{0.8} \\
 & Wanda & 47.9\std{0.8} & 17.6\std{0.7} & 15.0\std{0.8} & 3.8\std{0.2} & 2.9\std{0.2} & 1.9\std{0.2} & 6.4\std{0.4} & 13.2\std{0.4} & 15.0\std{0.6} & 68.3\std{0.3} & 82.3\std{0.9} & 77.1\std{1.4} & 29.3\std{0.6} \\
 & Taylor & 44.7\std{1.3} & 15.3\std{2.0} & 15.1\std{1.1} & 4.3\std{0.7} & 3.9\std{0.2} & 1.9\std{0.4} & 6.6\std{0.6} & 16.4\std{0.7} & 14.7\std{0.5} & 72.1\std{0.6} & 81.7\std{0.2} & 78.4\std{2.2} & 29.6\std{0.9} \\
 & SnapViT & 46.1\std{0.8} & 16.5\std{0.7} & 15.2\std{0.9} & 4.0\std{0.2} & 3.2\std{0.2} & 1.8\std{0.1} & 6.0\std{0.2} & 15.1\std{1.0} & 13.6\std{1.0} & 75.7\std{0.7} & 79.4\std{0.6} & 73.8\std{0.9} & 29.2\std{0.6} \\
 & BlockDrop & 32.7\std{0.9} & 10.0\std{0.4} & 17.9\std{0.3} & 5.2\std{0.2} & 2.4\std{0.0} & 2.8\std{0.2} & 5.3\std{0.3} & 11.9\std{0.3} & 13.5\std{0.4} & 70.8\std{0.1} & 80.6\std{1.4} & 66.8\std{0.4} & 26.7\std{0.1} \\
 & FLAP & 57.3\std{0.7} & 29.8\std{0.4} & 18.6\std{0.9} & 5.4\std{0.2} & 3.8\std{0.4} & 3.1\std{0.2} & 7.5\std{0.7} & 18.2\std{0.2} & 17.7\std{0.8} & 79.9\std{0.5} & 81.9\std{1.3} & 88.8\std{0.8} & 34.3\std{0.2} \\
 & SVD-LLM & \textbf{84.3\std{0.2}} & \textbf{59.9\std{0.5}} & 42.7\std{1.1} & 17.9\std{1.0} & 13.3\std{0.4} & 12.1\std{0.7} & 26.6\std{0.4} & 44.4\std{0.8} & 38.7\std{1.3} & \textbf{94.1\std{0.1}} & \textbf{85.0\std{2.1}} & 92.6\std{0.3} & 51.0\std{0.1} \\
 & CORP & 46.0\std{5.6} & 22.1\std{0.6} & 24.3\std{1.0} & 6.9\std{0.2} & 5.1\std{1.2} & 3.4\std{0.3} & 9.5\std{1.3} & 21.2\std{0.6} & 20.3\std{1.8} & 86.2\std{1.8} & 80.6\std{2.3} & 83.9\std{0.3} & 34.1\std{0.7} \\
 & \textbf{Carve VFM} & 82.6\std{1.6} & 59.0\std{1.3} & \textbf{45.0\std{1.9}} & \textbf{32.4\std{5.4}} & \textbf{17.3\std{7.5}} & \textbf{12.7\std{6.2}} & \textbf{39.0\std{6.2}} & \textbf{50.1\std{2.0}} & \textbf{46.2\std{3.7}} & 93.1\std{0.8} & 81.2\std{2.7} & \textbf{92.8\std{0.8}} & \textbf{54.3\std{1.1}} \\
\bottomrule
\end{tabular}
\end{adjustbox}

\end{table}
\end{landscape}

\begin{landscape}
\begin{table}[p]
\caption{Complete per-dataset frozen-backbone linear-probing accuracy (\%) for MetaCLIP-2 ViT-B/16 at three dense parameter budgets.}
\label{tab:supp-metaclip2-vitb}
\centering
\footnotesize
\setlength{\tabcolsep}{2.2pt}
\renewcommand{\arraystretch}{0.84}
\begin{adjustbox}{width=\linewidth,max totalheight=.82\textheight,keepaspectratio}
\begin{tabular}{@{}ll*{12}{r}@{\hspace{2.5pt}}r@{}}
\toprule
Budget & Method
 & \multicolumn{3}{c}{General}
 & \multicolumn{4}{c}{Fine-grained}
 & \multicolumn{2}{c}{Texture/scene}
 & \multicolumn{3}{c}{Medical} & \textbf{Mean} \\
\cmidrule(lr){3-5}\cmidrule(lr){6-9}\cmidrule(lr){10-11}
\cmidrule(lr){12-14}\cmidrule(l){15-15}
 & & C10 & C100 & Cal. & Air. & CUB & Cars & Pets & DTD & MIT
 & Blood & Pneu. & OrganA & \\
\midrule
85.8M & Linear probe & 97.5\std{0.1} & 86.4\std{0.0} & 96.0\std{0.3} & 69.5\std{0.2} & 81.1\std{0.2} & 91.8\std{0.2} & 92.0\std{0.3} & 81.0\std{0.4} & 92.6\std{0.1} & 94.7\std{0.3} & 83.9\std{1.0} & 88.1\std{0.5} & 87.9\std{0.0} \\
\cmidrule(lr){1-15}
\multirow{10}{*}{\textbf{60M}} & Random & 55.4\std{2.7} & 28.9\std{2.4} & 31.4\std{3.5} & 10.6\std{3.7} & 7.8\std{2.2} & 5.5\std{1.7} & 14.1\std{1.7} & 26.7\std{3.0} & 28.4\std{3.9} & 90.2\std{2.6} & 83.4\std{2.4} & 84.4\std{1.8} & 38.9\std{2.1} \\
 & Magnitude & 59.5\std{0.2} & 34.9\std{0.7} & 33.8\std{1.2} & 11.5\std{0.6} & 7.0\std{0.1} & 6.0\std{0.1} & 17.1\std{0.8} & 30.8\std{0.1} & 34.6\std{0.3} & 87.9\std{1.5} & 83.8\std{1.0} & 88.1\std{0.5} & 41.2\std{0.3} \\
 & LAMP & 60.5\std{0.1} & 35.5\std{0.7} & 32.3\std{0.8} & 10.8\std{0.0} & 7.3\std{0.3} & 7.3\std{0.2} & 19.2\std{0.2} & 30.6\std{1.2} & 36.9\std{0.8} & 91.3\std{0.9} & 87.5\std{1.7} & 89.3\std{0.3} & 42.4\std{0.1} \\
 & SNIP--Mag. & 63.7\std{1.0} & 36.2\std{4.7} & 29.1\std{4.6} & 8.9\std{1.2} & 5.5\std{0.4} & 4.3\std{0.6} & 12.9\std{3.9} & 18.8\std{0.9} & 18.9\std{0.4} & 73.9\std{8.9} & 83.7\std{1.5} & 86.7\std{1.3} & 36.9\std{2.5} \\
 & SnapViT & 55.1\std{1.3} & 28.8\std{0.6} & 28.7\std{0.9} & 8.1\std{0.4} & 6.4\std{0.1} & 4.6\std{0.0} & 12.0\std{0.5} & 22.4\std{0.2} & 25.0\std{0.7} & 83.5\std{1.9} & 82.3\std{0.2} & 87.4\std{0.7} & 37.0\std{0.0} \\
 & BlockDrop & 81.0\std{0.1} & 56.3\std{0.4} & 45.8\std{0.2} & 4.7\std{1.0} & 7.4\std{0.2} & 6.4\std{0.0} & 17.1\std{0.3} & 33.5\std{0.4} & 39.8\std{0.7} & 89.0\std{0.5} & 82.9\std{1.9} & 87.7\std{0.2} & 46.0\std{0.2} \\
 & FLAP & 84.0\std{0.6} & 59.7\std{1.1} & 43.3\std{1.0} & 24.6\std{2.0} & 27.4\std{3.4} & 42.7\std{5.3} & 52.5\std{0.9} & 52.7\std{0.3} & 49.3\std{0.6} & 92.4\std{1.1} & 86.7\std{1.7} & 88.2\std{0.3} & 58.6\std{0.6} \\
 & SVD-LLM & 94.8\std{0.2} & 78.8\std{0.4} & 82.2\std{0.5} & 52.0\std{0.3} & 65.3\std{0.5} & 77.4\std{0.0} & 85.3\std{0.5} & 73.2\std{0.3} & 83.8\std{0.2} & 95.0\std{0.1} & \textbf{88.2\std{1.1}} & \textbf{91.1\std{0.1}} & 80.6\std{0.2} \\
 & CORP & 96.8\std{0.3} & 83.2\std{0.7} & 92.2\std{0.4} & \textbf{65.0\std{0.5}} & 71.7\std{0.2} & 84.9\std{0.4} & 88.7\std{0.3} & 60.1\std{0.1} & 89.5\std{0.2} & \textbf{95.4\std{0.5}} & 85.4\std{3.1} & 90.0\std{0.3} & 83.6\std{0.4} \\
 & \textbf{Carve VFM} & \textbf{97.2\std{0.1}} & \textbf{84.1\std{0.4}} & \textbf{93.2\std{0.4}} & 64.8\std{1.2} & \textbf{77.7\std{0.8}} & \textbf{89.2\std{0.1}} & \textbf{91.7\std{0.3}} & \textbf{78.2\std{0.9}} & \textbf{90.1\std{0.4}} & \textbf{95.4\std{0.0}} & 88.1\std{2.0} & 89.6\std{0.6} & \textbf{86.6\std{0.2}} \\
\cmidrule(lr){1-15}
\multirow{10}{*}{\textbf{35M}} & Random & 57.3\std{1.7} & 31.7\std{2.3} & 27.3\std{1.7} & 7.6\std{0.6} & 5.9\std{0.4} & 4.7\std{0.3} & 13.2\std{1.2} & 25.4\std{1.2} & 26.3\std{1.0} & 87.7\std{3.0} & 84.8\std{1.4} & 88.9\std{0.4} & 38.4\std{0.6} \\
 & Magnitude & 56.9\std{0.8} & 30.7\std{0.5} & 27.6\std{1.4} & 9.3\std{0.3} & 5.6\std{0.2} & 5.0\std{0.0} & 12.5\std{0.8} & 25.3\std{1.0} & 27.5\std{1.0} & 85.4\std{1.4} & 85.3\std{0.5} & 88.7\std{0.3} & 38.3\std{0.2} \\
 & LAMP & 58.5\std{1.0} & 31.0\std{0.7} & 27.0\std{0.7} & 8.7\std{0.1} & 5.4\std{0.1} & 4.9\std{0.3} & 13.4\std{0.7} & 25.8\std{0.4} & 27.5\std{2.0} & 88.8\std{1.3} & 87.1\std{1.3} & 89.3\std{0.7} & 39.0\std{0.1} \\
 & SNIP--Mag. & 59.0\std{1.1} & 31.6\std{1.8} & 22.6\std{2.9} & 6.2\std{0.4} & 4.4\std{0.2} & 5.0\std{0.2} & 9.8\std{1.3} & 21.1\std{0.9} & 21.1\std{4.0} & 79.0\std{5.2} & 84.9\std{0.8} & 88.0\std{2.0} & 36.1\std{1.7} \\
 & SnapViT & 52.9\std{0.3} & 26.8\std{0.6} & 23.7\std{0.9} & 7.1\std{0.3} & 6.1\std{0.2} & 4.3\std{0.2} & 10.9\std{0.5} & 23.3\std{0.5} & 22.0\std{1.0} & 85.9\std{1.6} & 84.2\std{0.7} & 86.6\std{0.2} & 36.1\std{0.2} \\
 & BlockDrop & 64.6\std{0.1} & 37.9\std{0.3} & 31.6\std{0.8} & 4.2\std{0.1} & 4.8\std{0.2} & 3.8\std{0.4} & 7.9\std{3.5} & 24.8\std{9.8} & 30.2\std{0.1} & 89.3\std{0.4} & 79.0\std{1.5} & 86.6\std{0.9} & 38.7\std{1.0} \\
 & FLAP & 63.2\std{0.6} & 39.4\std{0.8} & 27.5\std{0.7} & 11.0\std{0.7} & 6.5\std{0.4} & 4.2\std{0.4} & 16.1\std{1.6} & 32.4\std{0.8} & 23.3\std{0.4} & 86.8\std{1.2} & \textbf{88.0\std{0.3}} & 87.6\std{0.3} & 40.5\std{0.2} \\
 & SVD-LLM & 89.2\std{0.4} & 66.9\std{0.7} & 48.3\std{1.2} & 21.7\std{0.7} & 21.3\std{1.5} & 10.3\std{0.3} & 38.2\std{0.8} & 50.1\std{1.7} & 49.5\std{2.3} & \textbf{94.8\std{0.4}} & 83.7\std{2.0} & 91.3\std{0.8} & 55.4\std{0.5} \\
 & CORP & 89.0\std{0.1} & 66.1\std{0.4} & 56.8\std{0.9} & 33.5\std{0.6} & 42.1\std{0.2} & 44.5\std{1.0} & 59.9\std{0.5} & 44.2\std{1.1} & 56.3\std{2.7} & 94.6\std{0.7} & 86.9\std{4.0} & 91.4\std{0.7} & 63.8\std{0.2} \\
 & \textbf{Carve VFM} & \textbf{92.0\std{0.5}} & \textbf{70.4\std{0.8}} & \textbf{68.6\std{2.0}} & \textbf{39.5\std{0.9}} & \textbf{52.3\std{4.1}} & \textbf{60.4\std{6.2}} & \textbf{75.0\std{0.1}} & \textbf{54.4\std{9.7}} & \textbf{74.1\std{1.2}} & 93.9\std{0.0} & 85.4\std{2.4} & \textbf{92.2\std{0.2}} & \textbf{71.5\std{1.0}} \\
\cmidrule(lr){1-15}
\multirow{10}{*}{\textbf{20M}} & Random & 52.6\std{1.1} & 25.1\std{1.0} & 20.7\std{1.8} & 5.2\std{0.2} & 3.9\std{0.2} & 2.9\std{0.4} & 9.5\std{1.0} & 19.5\std{0.4} & 19.3\std{1.4} & 80.8\std{2.6} & 82.5\std{1.5} & 86.8\std{1.8} & 34.1\std{0.8} \\
 & Magnitude & 54.1\std{0.2} & 26.5\std{0.1} & 23.1\std{1.0} & 6.8\std{0.2} & 4.4\std{0.2} & 3.4\std{0.2} & 10.8\std{0.6} & 20.8\std{0.4} & 21.9\std{0.9} & 80.3\std{1.6} & 82.7\std{0.4} & 87.6\std{0.4} & 35.2\std{0.2} \\
 & LAMP & 54.7\std{0.2} & 27.3\std{0.6} & 22.8\std{0.6} & 6.5\std{0.2} & 4.5\std{0.2} & 3.6\std{0.4} & 10.8\std{0.4} & 20.7\std{0.5} & 22.1\std{1.2} & 81.1\std{1.5} & 82.9\std{0.8} & 87.4\std{0.8} & 35.4\std{0.1} \\
 & SNIP--Mag. & 56.2\std{0.5} & 26.1\std{0.6} & 21.3\std{1.0} & 6.2\std{0.2} & 4.7\std{0.2} & 3.8\std{0.1} & 9.7\std{1.0} & 20.5\std{0.6} & 20.7\std{1.7} & 83.7\std{3.2} & 84.5\std{1.2} & 86.2\std{0.8} & 35.3\std{0.9} \\
 & SnapViT & 51.4\std{0.8} & 25.6\std{0.6} & 22.3\std{0.5} & 5.5\std{0.2} & 5.2\std{0.1} & 3.8\std{0.2} & 9.6\std{0.7} & 19.7\std{0.4} & 20.7\std{0.7} & 87.2\std{1.3} & 84.1\std{0.7} & 87.0\std{0.6} & 35.2\std{0.1} \\
 & BlockDrop & 54.0\std{1.0} & 26.0\std{0.2} & 20.8\std{0.9} & 2.9\std{0.4} & 1.7\std{0.5} & 2.5\std{0.1} & 6.8\std{2.4} & 12.2\std{7.6} & 18.1\std{0.3} & 78.4\std{0.4} & 79.1\std{2.0} & 85.5\std{1.2} & 32.3\std{0.8} \\
 & FLAP & 63.2\std{0.9} & 35.7\std{0.5} & 25.6\std{1.2} & 7.1\std{0.5} & 4.9\std{0.1} & 4.7\std{0.1} & 13.1\std{0.0} & 28.9\std{0.9} & 25.1\std{1.2} & 89.9\std{1.1} & \textbf{86.6\std{0.1}} & 90.1\std{0.4} & 39.6\std{0.1} \\
 & SVD-LLM & \textbf{77.7\std{0.1}} & \textbf{50.9\std{0.6}} & 31.9\std{0.8} & 9.5\std{0.5} & 6.6\std{0.5} & 4.3\std{0.5} & 12.7\std{0.1} & 33.1\std{0.8} & 28.6\std{0.9} & 90.1\std{0.6} & 85.4\std{2.5} & 90.7\std{0.9} & 43.5\std{0.3} \\
 & CORP & 62.2\std{0.7} & 34.1\std{0.4} & 25.4\std{0.3} & 6.3\std{0.1} & 5.4\std{0.3} & 3.0\std{0.2} & 10.5\std{1.0} & 20.2\std{0.8} & 21.9\std{1.1} & 90.2\std{0.2} & 84.5\std{0.8} & 89.9\std{0.9} & 37.8\std{0.1} \\
 & \textbf{Carve VFM} & 76.6\std{1.3} & 42.5\std{5.1} & \textbf{36.6\std{1.1}} & \textbf{16.5\std{0.9}} & \textbf{15.9\std{1.0}} & \textbf{7.8\std{1.9}} & \textbf{34.5\std{1.4}} & \textbf{42.3\std{4.4}} & \textbf{43.9\std{1.2}} & \textbf{91.1\std{0.3}} & 85.1\std{2.4} & \textbf{90.8\std{2.2}} & \textbf{48.6\std{0.5}} \\
\bottomrule
\end{tabular}
\end{adjustbox}

\end{table}
\end{landscape}

\begin{landscape}
\begin{table}[p]
\caption{Complete per-dataset frozen-backbone linear-probing accuracy (\%) for DINOv3 ViT-L/16 at three dense parameter budgets.}
\label{tab:supp-dinov3-vitl}
\centering
\footnotesize
\setlength{\tabcolsep}{2.2pt}
\renewcommand{\arraystretch}{0.88}
\begin{adjustbox}{width=\linewidth,max totalheight=.82\textheight,keepaspectratio}
\begin{tabular}{@{}ll*{12}{r}@{\hspace{2.5pt}}r@{}}
\toprule
Budget & Method
 & \multicolumn{3}{c}{General}
 & \multicolumn{4}{c}{Fine-grained}
 & \multicolumn{2}{c}{Texture/scene}
 & \multicolumn{3}{c}{Medical} & \textbf{Mean} \\
\cmidrule(lr){3-5}\cmidrule(lr){6-9}\cmidrule(lr){10-11}
\cmidrule(lr){12-14}\cmidrule(l){15-15}
 & & C10 & C100 & Cal. & Air. & CUB & Cars & Pets & DTD & MIT
 & Blood & Pneu. & OrganA & \\
\midrule
303.1M & Linear probe & 99.1\std{0.0} & 93.1\std{0.1} & 97.3\std{0.1} & 80.4\std{0.1} & 90.6\std{0.1} & 94.4\std{0.1} & 96.7\std{0.1} & 82.7\std{0.2} & 90.7\std{0.1} & 95.9\std{0.0} & 87.3\std{0.6} & 92.3\std{0.1} & 91.7\std{0.0} \\
\cmidrule(lr){1-15}
\multirow{8}{*}{\textbf{200M}} & Random & 67.5\std{0.5} & 41.7\std{0.6} & 54.6\std{0.7} & 20.4\std{3.2} & 13.4\std{0.5} & 11.4\std{2.1} & 28.1\std{1.2} & 42.9\std{1.0} & 47.4\std{2.3} & 93.1\std{1.0} & 85.5\std{0.9} & 90.7\std{1.4} & 49.7\std{0.6} \\
 & SNIP--Mag. & 69.6\std{0.7} & 44.5\std{1.3} & 50.3\std{3.1} & 25.8\std{1.0} & 15.3\std{1.2} & 17.2\std{1.3} & 26.7\std{4.6} & 40.4\std{1.6} & 47.9\std{3.0} & 94.6\std{0.3} & 86.4\std{0.7} & 92.1\std{0.8} & 50.9\std{0.6} \\
 & SnapViT & 72.2\std{1.0} & 46.5\std{0.3} & 61.8\std{0.8} & 26.8\std{0.5} & 15.9\std{0.4} & 18.6\std{0.1} & 28.0\std{0.5} & 45.8\std{0.1} & 55.5\std{1.0} & 93.1\std{0.1} & 85.2\std{0.9} & 92.9\std{0.3} & 53.5\std{0.1} \\
 & BlockDrop & 77.8\std{0.5} & 56.4\std{0.2} & 76.5\std{0.7} & 41.8\std{5.1} & 46.7\std{2.7} & 58.6\std{0.5} & 65.9\std{0.2} & 64.8\std{0.1} & 70.1\std{0.9} & 95.6\std{0.2} & 86.1\std{0.8} & 90.8\std{0.4} & 69.3\std{0.2} \\
 & FLAP & 93.4\std{0.3} & 77.5\std{0.2} & 88.7\std{0.1} & 73.7\std{0.7} & 81.2\std{1.0} & 90.7\std{0.5} & 92.6\std{0.1} & 78.5\std{0.5} & 88.7\std{0.3} & 95.7\std{0.4} & 86.2\std{2.0} & 91.6\std{0.1} & 86.6\std{0.2} \\
 & SVD-LLM & \textbf{98.9\std{0.0}} & 90.7\std{0.1} & 96.1\std{0.1} & 80.6\std{1.2} & 89.7\std{0.1} & \textbf{94.1\std{0.1}} & \textbf{96.5\std{0.0}} & 83.2\std{0.3} & \textbf{91.0\std{0.5}} & 96.2\std{0.4} & \textbf{89.9\std{1.9}} & 93.1\std{0.2} & \textbf{91.7\std{0.3}} \\
 & CORP & \textbf{98.9\std{0.0}} & 89.5\std{2.8} & \textbf{96.5\std{0.1}} & 79.8\std{2.4} & 83.1\std{0.1} & 88.7\std{0.2} & 95.8\std{0.2} & 63.8\std{0.1} & 90.6\std{0.2} & \textbf{97.0\std{0.3}} & 88.2\std{1.1} & 93.0\std{0.4} & 88.8\std{0.2} \\
 & \textbf{Carve VFM} & \textbf{98.9\std{0.1}} & \textbf{90.9\std{0.3}} & 96.3\std{0.2} & \textbf{81.6\std{0.6}} & \textbf{89.8\std{0.7}} & 94.0\std{0.3} & \textbf{96.5\std{0.1}} & \textbf{83.3\std{0.9}} & 90.8\std{0.4} & 96.5\std{0.7} & 87.9\std{1.7} & \textbf{93.6\std{0.2}} & \textbf{91.7\std{0.5}} \\
\cmidrule(lr){1-15}
\multirow{8}{*}{\textbf{150M}} & Random & 52.3\std{2.6} & 23.7\std{2.5} & 25.6\std{2.5} & 6.3\std{1.2} & 5.6\std{1.3} & 3.7\std{0.5} & 9.9\std{1.2} & 21.7\std{2.6} & 23.1\std{0.6} & 77.9\std{5.0} & 82.3\std{1.7} & 84.0\std{0.8} & 34.7\std{1.5} \\
 & SNIP--Mag. & 60.2\std{0.6} & 31.7\std{1.5} & 33.1\std{1.0} & 11.7\std{1.0} & 11.2\std{0.7} & 7.2\std{0.2} & 14.2\std{0.9} & 29.1\std{0.8} & 33.5\std{1.6} & 90.3\std{0.9} & 83.4\std{1.5} & 89.5\std{0.8} & 41.3\std{0.2} \\
 & SnapViT & 60.9\std{1.1} & 33.5\std{0.6} & 36.6\std{0.7} & 11.0\std{0.1} & 8.4\std{0.5} & 6.5\std{0.4} & 13.7\std{0.3} & 30.5\std{0.5} & 33.4\std{0.9} & 86.9\std{0.4} & 84.0\std{1.4} & 89.7\std{0.3} & 41.3\std{0.1} \\
 & BlockDrop & 62.1\std{7.6} & 41.6\std{0.5} & 34.2\std{0.2} & 8.7\std{3.4} & 9.0\std{1.1} & 8.9\std{0.3} & 19.0\std{0.6} & 38.4\std{0.2} & 27.9\std{2.4} & 92.6\std{0.3} & 84.6\std{0.7} & 87.1\std{0.1} & 42.8\std{0.8} \\
 & FLAP & 84.1\std{0.5} & 62.0\std{0.3} & 70.4\std{0.6} & 64.3\std{0.7} & 60.6\std{0.9} & 83.9\std{1.1} & 78.6\std{0.1} & 72.3\std{0.1} & 71.5\std{0.7} & 95.5\std{0.4} & 85.3\std{1.3} & 92.0\std{0.0} & 76.7\std{0.2} \\
 & SVD-LLM & \textbf{98.4\std{0.0}} & 88.1\std{0.1} & 93.6\std{0.3} & 79.5\std{1.4} & 88.2\std{0.3} & \textbf{94.0\std{0.1}} & \textbf{95.9\std{0.1}} & 80.2\std{0.5} & 90.2\std{0.2} & 95.8\std{0.7} & \textbf{88.2\std{2.1}} & 92.8\std{0.2} & 90.4\std{0.3} \\
 & CORP & \textbf{98.4\std{0.1}} & 88.2\std{0.3} & 93.1\std{0.3} & 78.0\std{2.2} & 77.8\std{4.2} & 88.4\std{0.1} & 94.4\std{1.0} & 62.1\std{0.9} & 89.8\std{0.1} & \textbf{96.4\std{0.2}} & 86.9\std{1.1} & 93.1\std{0.3} & 87.2\std{0.3} \\
 & \textbf{Carve VFM} & 98.1\std{0.5} & \textbf{88.3\std{1.5}} & \textbf{93.7\std{0.5}} & \textbf{81.2\std{0.7}} & \textbf{88.5\std{0.4}} & 93.6\std{0.1} & 95.7\std{0.3} & \textbf{82.2\std{0.6}} & \textbf{90.4\std{0.6}} & 95.9\std{0.3} & 86.8\std{2.7} & \textbf{93.6\std{0.4}} & \textbf{90.7\std{0.7}} \\
\cmidrule(lr){1-15}
\multirow{8}{*}{\textbf{100M}} & Random & 40.4\std{2.7} & 13.5\std{0.7} & 12.8\std{0.2} & 2.8\std{0.4} & 2.4\std{0.2} & 1.5\std{0.1} & 5.6\std{0.2} & 12.5\std{0.6} & 11.3\std{1.3} & 62.2\std{3.0} & 78.9\std{3.2} & 75.4\std{0.4} & 26.6\std{0.7} \\
 & SNIP--Mag. & 49.2\std{1.3} & 22.7\std{1.1} & 20.5\std{1.3} & 5.6\std{0.2} & 6.4\std{0.4} & 3.5\std{0.2} & 10.0\std{0.2} & 18.4\std{0.6} & 23.1\std{1.7} & 76.5\std{5.4} & 81.0\std{1.0} & 82.2\std{0.3} & 33.3\std{0.3} \\
 & SnapViT & 47.5\std{2.0} & 19.4\std{0.9} & 18.7\std{0.6} & 4.4\std{0.2} & 3.7\std{0.3} & 2.6\std{0.4} & 7.3\std{0.7} & 16.5\std{0.4} & 17.2\std{1.3} & 68.8\std{0.5} & 81.7\std{1.1} & 82.4\std{1.0} & 30.8\std{0.3} \\
 & BlockDrop & 54.6\std{4.7} & 32.8\std{0.3} & 25.6\std{0.7} & 8.4\std{1.0} & 7.6\std{0.9} & 5.4\std{0.2} & 14.6\std{1.5} & 25.1\std{0.1} & 15.6\std{1.6} & 90.8\std{0.3} & 84.7\std{1.5} & 85.0\std{0.1} & 37.5\std{0.3} \\
 & FLAP & 77.4\std{0.2} & 52.4\std{0.3} & 45.4\std{0.6} & 29.3\std{0.6} & 11.4\std{0.4} & 12.5\std{0.8} & 28.9\std{0.5} & 51.5\std{0.7} & 40.8\std{0.7} & 92.9\std{0.3} & 84.0\std{0.6} & 92.8\std{0.1} & 51.6\std{0.1} \\
 & SVD-LLM & \textbf{96.1\std{0.0}} & \textbf{80.0\std{0.2}} & 78.5\std{0.5} & \textbf{74.0\std{0.6}} & \textbf{81.5\std{0.2}} & \textbf{91.8\std{0.1}} & \textbf{92.0\std{0.2}} & 69.8\std{1.0} & \textbf{82.1\std{0.3}} & \textbf{94.9\std{0.3}} & \textbf{86.1\std{2.2}} & 92.0\std{0.4} & \textbf{84.9\std{0.2}} \\
 & CORP & 71.0\std{8.0} & 37.1\std{15.6} & 57.9\std{3.5} & 36.9\std{10.0} & 33.5\std{2.4} & 59.6\std{2.7} & 58.1\std{3.6} & 45.7\std{1.1} & 57.1\std{1.4} & 91.8\std{0.6} & 80.4\std{3.5} & 89.5\std{3.4} & 59.9\std{0.8} \\
 & \textbf{Carve VFM} & 94.2\std{0.9} & 76.9\std{4.2} & \textbf{80.1\std{1.3}} & 73.6\std{1.6} & 76.4\std{5.4} & 89.0\std{1.8} & 90.6\std{0.8} & \textbf{74.0\std{1.0}} & 80.3\std{1.9} & 92.7\std{2.9} & 84.2\std{1.2} & \textbf{93.1\std{0.4}} & 83.7\std{1.9} \\
\bottomrule
\end{tabular}
\end{adjustbox}

\end{table}
\end{landscape}

\begin{landscape}
\begin{table}[p]
\caption{Per-dataset linear-probing accuracy (\%) for SliceGPT and OSSCAR at mild and tight budgets.}
\label{tab:supp-additional-baselines}
\centering
\footnotesize
\setlength{\tabcolsep}{2.2pt}
\renewcommand{\arraystretch}{1.0}
\begin{adjustbox}{width=\linewidth,max totalheight=.82\textheight,keepaspectratio}
\begin{tabular}{@{}ll*{12}{r}@{\hspace{2.5pt}}r@{}}
\toprule
Budget & Method
 & \multicolumn{3}{c}{General}
 & \multicolumn{4}{c}{Fine-grained}
 & \multicolumn{2}{c}{Texture/scene}
 & \multicolumn{3}{c}{Medical} & \textbf{Mean} \\
\cmidrule(lr){3-5}\cmidrule(lr){6-9}\cmidrule(lr){10-11}
\cmidrule(lr){12-14}\cmidrule(l){15-15}
 & & C10 & C100 & Cal. & Air. & CUB & Cars & Pets & DTD & MIT
 & Blood & Pneu. & OrganA & \\
\midrule
\multicolumn{15}{l}{\textbf{DINOv3 ViT-S/16}} \\
14M & SliceGPT & 92.4\std{0.2} & 75.7\std{0.3} & 41.0\std{1.1} & 7.9\std{0.6} & 7.5\std{0.6} & 5.4\std{0.5} & 26.5\std{0.9} & 59.5\std{1.1} & 51.1\std{1.3} & 95.3\std{0.4} & 88.3\std{1.0} & 92.7\std{0.1} & 53.6\std{0.3} \\
14M & OSSCAR & 91.3\std{0.1} & 73.5\std{0.3} & 75.1\std{0.8} & 57.5\std{1.1} & 59.7\std{0.1} & 75.6\std{0.3} & 81.6\std{0.3} & 69.8\std{0.9} & 69.1\std{6.1} & 96.1\std{0.1} & 88.9\std{0.8} & 93.0\std{0.2} & 77.6\std{0.6} \\
6M & SliceGPT & 78.4\std{0.3} & 52.0\std{0.4} & 26.4\std{0.7} & 7.6\std{0.4} & 4.3\std{0.6} & 3.0\std{0.1} & 19.5\std{0.5} & 25.2\std{0.9} & 18.7\std{0.5} & 80.4\std{0.9} & 80.7\std{1.5} & 90.1\std{0.2} & 40.5\std{0.1} \\
6M & OSSCAR & 57.7\std{0.7} & 31.4\std{0.2} & 27.7\std{1.8} & 12.4\std{0.6} & 10.9\std{0.2} & 5.4\std{0.3} & 18.5\std{0.8} & 41.6\std{1.0} & 24.5\std{0.4} & 89.7\std{0.3} & 82.2\std{1.0} & 86.2\std{0.4} & 40.7\std{0.1} \\
\midrule
\multicolumn{15}{l}{\textbf{DINOv3 ViT-B/16}} \\
60M & SliceGPT & 96.6\std{0.1} & 83.3\std{0.3} & 89.6\std{0.3} & 78.1\std{0.7} & 83.8\std{0.3} & 91.8\std{0.0} & 93.4\std{0.2} & 76.6\std{0.3} & 87.2\std{1.0} & 96.0\std{0.2} & 87.3\std{2.5} & 93.5\std{0.2} & 88.1\std{0.2} \\
60M & OSSCAR & 97.2\std{0.1} & 85.3\std{0.1} & 91.6\std{0.2} & 76.8\std{0.7} & 84.1\std{0.5} & 91.5\std{0.4} & 94.8\std{0.2} & 77.7\std{0.9} & 87.2\std{0.4} & 96.3\std{0.0} & 86.5\std{2.0} & 94.1\std{0.2} & 88.6\std{0.1} \\
20M & SliceGPT & 84.8\std{0.4} & 60.2\std{0.1} & 27.3\std{0.4} & 6.1\std{0.3} & 4.2\std{0.3} & 3.4\std{0.2} & 12.9\std{0.4} & 22.9\std{1.6} & 19.3\std{0.7} & 92.3\std{0.2} & 88.2\std{0.5} & 92.2\std{0.2} & 42.8\std{0.2} \\
20M & OSSCAR & 66.3\std{1.3} & 32.8\std{0.1} & 37.0\std{0.4} & 21.6\std{0.4} & 12.5\std{0.7} & 6.4\std{0.8} & 26.8\std{0.4} & 46.2\std{0.2} & 34.0\std{0.8} & 90.1\std{0.9} & 85.6\std{1.4} & 87.3\std{0.4} & 45.5\std{0.3} \\
\midrule
\multicolumn{15}{l}{\textbf{MetaCLIP-2 ViT-B/16}} \\
60M & SliceGPT & 96.8\std{0.1} & 82.0\std{0.2} & 91.1\std{0.8} & 59.9\std{1.3} & 74.7\std{0.7} & 88.9\std{0.1} & 90.3\std{0.5} & 77.4\std{0.8} & 88.7\std{0.4} & 93.8\std{0.3} & 83.8\std{3.2} & 88.1\std{1.2} & 84.6\std{0.2} \\
60M & OSSCAR & 96.5\std{0.3} & 82.0\std{0.6} & 90.9\std{0.5} & 62.4\std{0.7} & 74.0\std{0.1} & 87.3\std{0.5} & 90.4\std{0.5} & 76.2\std{0.6} & 87.7\std{0.3} & 93.6\std{0.8} & 87.7\std{2.2} & 89.2\std{1.1} & 84.8\std{0.3} \\
20M & SliceGPT & 85.8\std{0.3} & 56.0\std{1.0} & 43.7\std{0.3} & 19.4\std{1.0} & 22.8\std{0.3} & 27.4\std{0.4} & 53.1\std{1.1} & 43.7\std{1.4} & 35.2\std{1.3} & 88.2\std{0.8} & 82.1\std{2.6} & 84.0\std{1.0} & 53.4\std{0.4} \\
20M & OSSCAR & 66.9\std{0.5} & 38.5\std{0.7} & 28.2\std{0.7} & 10.7\std{1.7} & 5.3\std{0.4} & 5.2\std{0.2} & 20.3\std{1.6} & 40.1\std{0.6} & 29.2\std{1.1} & 83.8\std{2.3} & 81.0\std{3.3} & 86.6\std{0.6} & 41.3\std{0.4} \\
\midrule
\multicolumn{15}{l}{\textbf{DINOv3 ViT-L/16}} \\
200M & SliceGPT & 98.7\std{0.1} & 89.2\std{0.3} & 94.0\std{0.3} & 78.5\std{0.5} & 87.4\std{0.2} & 93.2\std{0.3} & 95.5\std{0.0} & 80.3\std{0.1} & 89.5\std{0.2} & 94.4\std{0.4} & 89.2\std{0.9} & 92.1\std{0.6} & 90.2\std{0.1} \\
200M & OSSCAR & 97.8\std{0.2} & 86.6\std{0.2} & 92.1\std{0.4} & 71.0\std{0.7} & 74.2\std{3.9} & 82.0\std{1.0} & 92.4\std{0.8} & 75.0\std{0.5} & 89.0\std{0.4} & 95.3\std{0.5} & 86.6\std{0.1} & 92.2\std{0.2} & 86.2\std{0.5} \\
100M & SliceGPT & 95.8\std{0.2} & 77.5\std{0.1} & 67.1\std{0.8} & 68.2\std{1.0} & 58.1\std{4.8} & 86.4\std{0.5} & 86.4\std{0.2} & 47.0\std{0.4} & 78.7\std{1.3} & 92.1\std{0.4} & 84.1\std{0.7} & 91.3\std{0.7} & 77.7\std{0.3} \\
100M & OSSCAR & 84.9\std{0.5} & 62.0\std{0.1} & 61.3\std{0.8} & 47.2\std{1.3} & 27.7\std{2.8} & 56.5\std{3.8} & 69.9\std{4.7} & 63.4\std{0.5} & 56.2\std{2.4} & 92.4\std{0.1} & 84.4\std{1.4} & 91.5\std{0.5} & 66.4\std{0.5} \\
\bottomrule
\end{tabular}
\end{adjustbox}
\end{table}
\end{landscape}

\FloatBarrier
\section{Carve VFM Implementation Details}
\label{sec:supp-method}

\subsection{Search Space and Budget Projection}

An architecture stores, for every transformer layer, an MLP removal ratio, an
attention removal ratio, and an attention operator. These removal ratios are
$1-s_l^{\mathrm{mlp}}$ and $1-s_l^{\mathrm{att}}$, the complements of the
retention ratios in the main paper. The MLP decision removes
matched intermediate channels from the up- and down-projections. The
attention operator is either whole-head carving (H) or coupled column carving
(C), which narrows every surviving head using paired Q/K coordinates and
matched V/O coordinates. Every proposed architecture is projected onto the
same parameter budget before scoring. This projection applies a common shift
to its continuous removal ratios, clips them to $[0,0.95]$ (retention in
$[0.05,1]$), solves the remaining
budget discrepancy by bisection, and rounds attention structures to a
feasible dense shape. Consequently, fitness cannot improve by silently
retaining more parameters.

The evolutionary population contains 20 candidates, of which eight are
retained as elites. Initial candidates include uniform and depth-biased
allocations, MLP-only boundary cases, and systematic and random H/C patterns.
Mutation perturbs continuous ratios with standard deviation 0.08 and flips an
operator with probability 0.18. Crossover swaps complete layer decisions.
Runs use at most 12 generations, stop after four generations without
improvement, and cap the total number of evaluated architectures at 120 or
200 depending on the campaign.

\subsection{Local Recoverability and Adaptive Global Audit}

Let $Z$ be the input to a linear projection, $W$ the downstream weight, and
$G=Z^\top Z/n+\lambda I$ its regularized Gram matrix. For a removable
coordinate group $\mathcal I$, the local damage score is
\begin{equation}
 d(\mathcal I)=\operatorname{tr}\!\left(
 W_{:\mathcal I}[G^{-1}_{\mathcal I\mathcal I}]^{-1}W_{:\mathcal I}^{\top}\right).
 \label{eq:supp-local-damage}
\end{equation}
Groups are removed greedily. After each removal, the working weights are
refitted and $G^{-1}$ is updated by a Schur complement. This yields a nested
removal order and cumulative damage curve for every layer and operator.
Candidate fitness is then a sum of curve lookups, so a complete generation can
be ranked without constructing or forwarding its models. Q/K rotary pairs,
whose effect is bilinear, are ordered by their perturbation of calibration
attention logits. These local quantities are used only for ranking; held-out
evaluations of repaired candidate subnetworks then audit those rankings.

A local sum cannot represent all cross-layer interactions. We therefore audit
a rank-stratified sample of 12 candidates with a global score: repeated
stratified probe accuracy on held-out partitions of the calibration set. The
search changes to this global fitness if the Spearman agreement with the
local ranking is below 0.35 or if the locally preferred candidate incurs more
than 0.03 normalized held-out regret. Otherwise, the inexpensive local score
continues to guide exploration. In both cases, global accuracy re-ranks the 20
finalists over five repeated splits. This policy switched to global search in
26\% of mild and 51\% of tight DINOv3 ViT-S/B runs, 79\% of MetaCLIP-2 runs,
and 89\% of DINOv3 ViT-L runs.

\subsection{Closed-Form Repair}

For a linear path whose activation coordinates are divided into retained
indices $\mathcal{S}$ and removed indices $\mathcal{I}$, the deleted output is
$Y_{\mathcal{I}}=Z_{\mathcal{I}}W_{\mathcal{I}}^\top$. Carve VFM predicts
this output from the \mbox{retained activation}:
\begin{equation}
 \Delta W^\star =
 Y_{\mathcal{I}}^\top Z_{\mathcal{S}}
 \left(Z_{\mathcal{S}}^\top Z_{\mathcal{S}}+\lambda I\right)^{-1},
 \qquad
 W_{\mathrm{merged}}=W_{\mathcal{S}}+
 \operatorname{SVD}_{r}(\Delta W^\star).
 \label{eq:supp-repair}
\end{equation}
Here $\operatorname{SVD}_r(A)=U_r\Sigma_rV_r^\top$ is the reconstructed matrix
from the leading $r$ singular triplets of $A$, with the same dimensions as $A$.
We use $\lambda=10^{-4}$ and include a mean-residual bias correction. The
low-rank factors exist only while estimating and truncating the correction;
after addition, the deployed projection is a single smaller dense matrix.
Equation~\eqref{eq:supp-local-damage} therefore supplies a search score, whereas
Equation~\eqref{eq:supp-repair} specifies the correction used when a candidate
is explicitly repaired.
The solve repairs MLP down-projections and linearly recoverable attention
paths. Q/K column carving uses alternating ridge updates against the original
pre-softmax logits. This update is retained only when it lowers end-to-end
calibration feature error; otherwise it is reverted, and up to the next two
finalists are tried. No masks, adapters, low-rank branches, or sparse kernels
remain at inference.

\begin{table}[t]
\caption{Search, repair, and frozen-backbone linear-probing settings.}
\label{tab:supp-hyperparameters}
\centering
\normalsize
\setlength{\tabcolsep}{3.5pt}
\begin{tabular}{@{}lr@{\qquad}lr@{}}
\toprule
Search setting & Value & Repair/linear-probing setting & Value \\
\midrule
Population & 20 & Calibration examples & 512 \\
Elites & 8 & Ridge $\lambda$ & $10^{-4}$ \\
Generations & $\leq12$ & DINOv3 S/B rank & 128 \\
Candidate cap & 120--200 & MetaCLIP/ViT-L rank & 256 \\
Mutation $\sigma$ & 0.08 & Linear-head epochs & 10 \\
Operator flip & 0.18 & Linear-head learning rate & $10^{-3}$ \\
Patience & 4 & Linear-head weight decay & $10^{-4}$ \\
Finalists & 20 & Checkpoints & 1, 3, 5, 10 \\
Audit candidates & 12 & Input resolution & $224^2$ \\
Audit $\rho$ threshold & 0.35 & Seeds & 17, 42, 91 \\
Audit regret threshold & 0.03 & & \\
\bottomrule
\end{tabular}

\end{table}

\subsection{Software and Hardware}

Runs use Python 3.10.19, PyTorch 2.9.1 with CUDA 13.0, torchvision 0.24.1,
and the Transformers 5.0 development branch. Each compression run uses one
NVIDIA RTX PRO 6000 Blackwell Server Edition GPU with 96 GB of memory;
independent task--seed runs are distributed across eight such devices. The
host contains two AMD EPYC 9754 128-core processors and 2 TB of system memory.
Unless a deployment benchmark explicitly states otherwise, reported training and search times
exclude queueing between independent jobs but include model loading and data
preparation performed inside the run.

\section{Wall-Clock Cost}
\label{sec:supp-runtime}

Each JSON summary records elapsed wall time from model loading through
compression and ten-epoch linear-head training. Table~\ref{tab:supp-runtime-table}
aggregates those values over completed tasks, budgets, and seeds on one RTX
PRO 6000 Blackwell GPU. The numbers therefore compare complete per-task runs,
not only the score calculation inside each method.

\begin{table}[!h]
\caption{Wall-clock minutes per complete task run, reported as mean$\pm$sample
standard deviation. A dash marks the incomplete archived DINOv3 S/B CORP
timing grid.}
\label{tab:supp-runtime-table}
\centering
\normalsize
\setlength{\tabcolsep}{3.2pt}
\begin{tabular}{@{}lrrrrr@{}}
\toprule
Source & Carve VFM & BlockDrop & FLAP & SVD-LLM & CORP \\
\midrule
DINOv3 ViT-S/B & 5.3\std{2.1} & 1.5\std{1.1} & 0.9\std{0.5}
 & 1.4\std{0.7} & -- \\
MetaCLIP-2 ViT-B & 7.6\std{1.8} & 2.0\std{1.3} & 1.2\std{0.6}
 & 2.1\std{0.5} & 2.5\std{1.6} \\
DINOv3 ViT-L & 16.4\std{2.7} & 7.2\std{2.9} & 2.2\std{0.8}
 & 4.7\std{0.8} & 27.1\std{7.4} \\
\bottomrule
\end{tabular}

\end{table}

Carve VFM costs more than most fixed one-shot rules because it evaluates an
architecture population, but remains a minutes-scale procedure: 5.3 minutes
for ViT-S/B, 7.6 for MetaCLIP-2 ViT-B, and 16.4 for ViT-L on average. The
adaptive audit is important to this cost. When the local ordering is
trustworthy, most candidates are ranked by cached table lookup rather than
model construction. At ViT-L scale, Carve VFM is also faster than the
iterative CORP implementation under the same end-to-end protocol.

\section{Full Fine-Tuning}
\label{sec:supp-full-finetuning}

The selected 14M and 10M DINOv3 ViT-S specialists are fine-tuned end to end
without repeating search. AdamW uses learning rate $10^{-5}$ and weight decay
$10^{-2}$ for the backbone, and learning rate $10^{-3}$ and weight decay
$10^{-4}$ for the classifier. The batch size is 32 and the schedule lasts ten
epochs. Every native compact model starts from its released pretrained
weights and follows the identical data, augmentation, optimization, and
checkpoint protocol.

\begin{table}[!t]
\caption{Full-fine-tuning accuracy (\%) under a 10-epoch schedule. Native models use
their published pretrained weights; our rows fine-tune the already selected
14M and 10M compressed ViT-S children without rerunning structure search.
All rows use batch size 32 and the same checkpoint-selection rule over epochs
1, 3, 5, and 10. Parameter counts vary slightly with the dataset-specific
classifier; representative counts are shown. Bold denotes the best fully
fine-tuned result in each column (ties at the reported precision are all
highlighted). Accuracies are displayed to one decimal.}
\label{tab:full-ft-native}
\centering
\small
\renewcommand{\arraystretch}{1.0}
\setlength{\tabcolsep}{2.4pt}
\resizebox{\textwidth}{!}{%
\begin{tabular}{@{}lc*{13}{r}@{}}
\toprule
Model & Params (M)
  & \multicolumn{3}{c}{General}
  & \multicolumn{4}{c}{Fine-grained}
  & \multicolumn{2}{c}{Texture/scene}
  & \multicolumn{3}{c}{Medical} & Mean \\
\cmidrule(lr){3-5}\cmidrule(lr){6-9}\cmidrule(lr){10-11}\cmidrule(lr){12-14}
  & & C10 & C100 & Cal. & Air. & CUB & Cars & Pets & DTD & MIT & Blood & Pneu. & OrganA & \\
\midrule
ResNet-18
  & 11.2 & 93.7 & 77.0 & 82.0 & 61.1 & 69.6 & 63.1 & 90.4 & 64.3 & 68.7 & 98.7 & 90.5 & 95.4 & 79.6 \\
ResNet-50
  & 23.5 & 95.9 & 81.1 & 87.8 & 61.7 & 74.2 & 66.8 & 92.3 & 66.3 & 73.1 & 98.7 & 91.5 & 94.2 & 82.0 \\
EfficientNet-B4
  & 17.6 & 94.8 & 79.9 & 86.9 & 49.3 & 65.6 & 46.4 & 89.3 & 63.6 & 64.0 & 97.6 & 89.1 & 93.2 & 76.6 \\
DeiT-Tiny
  & 5.5 & 95.8 & 80.6 & 84.1 & 53.6 & 73.3 & 58.2 & 91.3 & 65.5 & 71.6 & 98.7 & 92.3 & 95.9 & 80.1 \\
ConvNeXt-Pico
  & 8.5 & 96.7 & 83.5 & 88.8 & 71.7 & 79.5 & 77.4 & \textbf{93.5} & 70.4 & 79.1 & 98.7 & 94.4 & 95.3 & 85.8 \\
RegNetY-1.6GF
  & 10.3 & 96.5 & 82.6 & \textbf{89.0} & 57.2 & 74.5 & 65.4 & 92.3 & 63.3 & 70.1 & 98.7 & 89.6 & 95.2 & 81.2 \\
PoolFormer-S12
  & 11.4 & 95.3 & 79.7 & 87.7 & 65.1 & 76.1 & 71.7 & 93.0 & 70.4 & 76.4 & 98.4 & \textbf{95.2} & 95.7 & 83.7 \\
DenseNet-121
  & 7.0 & 95.7 & 79.9 & 86.5 & 69.7 & 77.8 & 72.9 & 92.3 & 68.5 & 72.8 & \textbf{98.8} & 92.3 & 94.8 & 83.5 \\
\midrule
Carve VFM
  & 14 & \textbf{97.3} & \textbf{86.7} & 88.8 & \textbf{74.2} & \textbf{80.8} & \textbf{89.7} & 91.5 & \textbf{75.5} & \textbf{82.2} & \textbf{98.8} & 94.2 & \textbf{97.2} & \textbf{88.1} \\
Carve VFM
  & 10 & 95.7 & 82.2 & 75.3 & 66.2 & 72.6 & 83.8 & 87.7 & 69.4 & 75.9 & 98.4 & 90.7 & 96.9 & 82.9 \\
\bottomrule
\end{tabular}%
}

\end{table}

Against the best native result in each task, the 14M carved specialist wins on
eight tasks and ties on one at the reported precision. Because
the table selects each model's best checkpoint over epochs 1, 3, 5, and 10,
it should not be interpreted as a fixed-epoch comparison.

The advantage depends on model capacity: the 10M specialist reaches 82.9\%,
below ConvNeXt-Pico (85.8\%), PoolFormer-S12 (83.7\%), and DenseNet-121
(83.5\%). Their representative parameter counts are 8.5M, 11.4M, and 7.0M,
respectively. Thus, the 14M result does not imply that every carved budget
outperforms native compact models; these comparisons use different model sizes.

\FloatBarrier
\section{Deployment Measurements}
\label{sec:supp-deployment}

The main paper visualizes the operating points; Table~\ref{tab:vitl-deployment}
reports the underlying measurements. CPU experiments use FP32, batch 16, and
32 threads on an AMD EPYC 9754. GPU experiments use BF16 and batch 64 on an
RTX PRO 6000 Blackwell. Batch-1 latency is averaged over 50 timed iterations
and throughput over 30 batches after warm-up. Preprocessing and data loading
are excluded. CPU memory is peak process resident set size and GPU memory is
peak allocated \mbox{device memory}.

\providecommand{\hstd}[1]{\,{\tiny$\pm$#1}}
\begin{table}[!h]
\caption{Deployment efficiency of ViT-L/16 compression at $224^2$ resolution.
We report compute, batch-1 latency, throughput, and peak memory usage on CPU and GPU.
Throughput uses batch 16 on CPU and batch 64 on GPU; memory is peak process RSS
on CPU and peak allocated device memory on GPU. Timings are mean$\pm$standard deviation.}
\label{tab:vitl-deployment}
\centering
\scriptsize
\setlength{\tabcolsep}{1.4pt}
\renewcommand{\arraystretch}{0.6}
\resizebox{\columnwidth}{!}{%
\begin{tabular}{@{}llcrrrr@{}}
\toprule
Device & Model & Budget (M) & GFLOPs$\downarrow$ &
Lat. (ms)$\downarrow$ & img/s$\uparrow$ & Mem. (MiB)$\downarrow$ \\
\midrule
\multirow{7}{*}{CPU}
 & ViT-L/16 & Full & 125.68 & 214.2\hstd{1.8} & 7.4\hstd{2.7} & 2246 \\
 & SVD-LLM & 200 & 84.17 & 190.3\hstd{1.6} & 9.3\hstd{0.5} & 1837 \\
 & Carve VFM & 200 & \textbf{83.49} & \textbf{117.4\hstd{3.9}} & \textbf{14.2\hstd{0.6}} & \textbf{1769\hstd{18}} \\
 & SVD-LLM & 150 & 64.11 & 182.7\hstd{1.3} & 9.8\hstd{0.6} & 1641 \\
 & Carve VFM & 150 & \textbf{63.30} & \textbf{105.6\hstd{6.6}} & \textbf{18.4\hstd{0.7}} & \textbf{1541\hstd{13}} \\
 & SVD-LLM & 100 & 43.96 & 161.1\hstd{1.0} & 10.4\hstd{1.0} & 1453 \\
 & Carve VFM & 100 & \textbf{42.87} & \textbf{99.0\hstd{14.1}} & \textbf{23.4\hstd{3.0}} & \textbf{1331\hstd{12}} \\
\midrule
\multirow{7}{*}{GPU}
 & ViT-L/16 & Full & 125.68 & 10.06\hstd{0.04} & 1371\hstd{1} & 934 \\
 & SVD-LLM & 200 & 84.17 & 13.24\hstd{0.04} & 1026\hstd{1} & 738 \\
 & Carve VFM & 200 & \textbf{83.49} & \textbf{10.82\hstd{0.20}} & \textbf{1722\hstd{38}} & \textbf{689\hstd{9}} \\
 & SVD-LLM & 150 & 64.11 & 12.96\hstd{0.83} & 1542\hstd{2} & 642 \\
 & Carve VFM & 150 & \textbf{63.30} & \textbf{10.71\hstd{0.27}} & \textbf{2090\hstd{65}} & \textbf{557\hstd{12}} \\
 & SVD-LLM & 100 & 43.96 & 13.43\hstd{0.82} & 1533\hstd{1} & 546 \\
 & Carve VFM & 100 & \textbf{42.87} & \textbf{10.85\hstd{0.24}} & \textbf{2708\hstd{131}} & \textbf{454\hstd{2}} \\
\bottomrule
\end{tabular}%
}

\end{table}

Higher batched throughput does not imply lower batch-1 latency. The full
ViT-L takes 10.06\,ms on GPU, compared with 10.82, 10.71, and 10.85\,ms for
the 200M, 150M, and 100M carved models: increases of 7.6\%, 6.5\%, and
7.9\%, respectively. These measurements distinguish batched deployment gains
from single-image response time; they do not identify the cause of the
latency difference.

\FloatBarrier
\section{Limitations and Future Work}
\label{sec:supp-limitations}

Our evaluation demonstrates Carve VFM across image classification, four source
backbones, and 512 labeled calibration examples per task. Natural extensions
include dense prediction, multimodal tasks, and unlabeled calibration. The
medical benchmarks broaden the evaluated domain coverage; extending this
evaluation to clinical settings will require dedicated validation.

Performance remains task- and budget-dependent. At 6M parameters, the
ViT-S frozen-backbone results are 18.5\% on CUB and 10.2\% on Cars
(Table~\ref{tab:supp-dinov3-vits}), highlighting room to improve frozen
representations under extreme compression. Some tight-budget estimates also
vary substantially across seeds, including 17.3$\pm$7.5\% on CUB for the 20M
ViT-B (Table~\ref{tab:supp-dinov3-vitb}). Full fine-tuning currently covers the
14M and 10M ViT-S specialists; evaluating the 6M specialist is a natural next
step for the most aggressive budget. Native compact models and factorized
models remain competitive in some settings, as detailed in
Appendices~\ref{sec:supp-full-finetuning} and~\ref{sec:supp-deployment}.

Deployment measurements cover the stated hardware, precision, and batch sizes.
The observed throughput and memory gains coexist with higher batch-1 GPU
latency than the full model. Complete task runs, including model loading,
compression, and ten-epoch linear-head training, also take longer than most fixed
one-shot baselines (Table~\ref{tab:supp-runtime-table}). These costs are paid
during compression and should be considered against the recurring savings over
the subsequent inference workload; measuring the break-even workload is an
important direction for deployment analysis.

\end{document}